\documentclass[a4paper,onecolumn, superscriptaddress,10pt,accepted=2022-09-16,issue=2, volume=5, shorttitle=papers/compositionality-5-2]{compositionalityarticle}
 \pdfoutput=1
\usepackage[utf8]{inputenc}
\usepackage[english]{babel}
\usepackage[T1]{fontenc}
\usepackage{amsmath}
\usepackage{hyperref}
\usepackage[numbers,sort&compress]{natbib} %jg use natbib style

\usepackage{tikz}
\usepackage{lipsum}

\newtheorem{definition}{Definition}
\newtheorem{proposition}{Proposition}
\newtheorem{proof}{Proof}

\usepackage{tikz}
\usepackage{pgfplots}
\usetikzlibrary{trees}
\usetikzlibrary{topaths}
\usetikzlibrary{decorations.pathmorphing}
\usetikzlibrary{decorations.markings}
\usetikzlibrary{matrix,backgrounds,folding}
\usetikzlibrary{chains,scopes,positioning,fit}
\usetikzlibrary{arrows,shadows}
\usetikzlibrary{calc} 
\usetikzlibrary{chains}
\usetikzlibrary{shapes,shapes.geometric,shapes.misc}

\tikzset{->-/.style={decoration={
  markings,
  mark=at position .5 with {\arrow{>}}},postaction={decorate}}}

\pgfdeclarelayer{edgelayer}
\pgfdeclarelayer{nodelayer}
\pgfsetlayers{background,edgelayer,nodelayer,main}

\tikzstyle{none}=[fill=none, font=\tiny]
\tikzstyle{title}=[fill=none]
\tikzstyle{lab}=[fill=white, font=\tiny]
\tikzstyle{new style 0}=[fill=white, draw=black, shape=rectangle]
\tikzstyle{new style 1}=[fill=white, draw=black, shape=circle]
\tikzstyle{new style 2}=[fill=white, draw=green, shape=circle]
\tikzstyle{new style 3}=[fill=white, draw=red, shape=circle]
\tikzstyle{new style 4}=[fill=white, draw=black, thick, shape=regular polygon, regular polygon sides=3]
\tikzstyle{counit}=[fill=black, shape=circle]
\tikzstyle{bigClasp}=[fill=white, draw=black, shape=circle, minimum width=1.7cm]

\tikzstyle{downArrow}=[->-]
\tikzstyle{thickArr}=[->-,double]
\tikzstyle{thick}=[line width = 2pt]
\tikzstyle{thickerArr}=[->-,double]
\tikzstyle{thicker}=[double]\tikzstyle{dashed}=[-, dashed]
\tikzstyle{every loop}=[]

\tikzstyle{(null)}=[]
\tikzstyle{plain}=[]

\def\VDOTS{\begin{picture}(0,0)\put(0,2){\circle*{1}}\put(0,-2){\circle*{1}}\put(0,-6){\circle*{1}}\end{picture}}

\usepackage{prftree, proof, stmaryrd}
\usepackage{mathtools, float, cancel, amssymb}
\usepackage{hyperref, geometry, enumerate, array}
\usepackage[all,cmtip]{xy}
\usepackage{lscape}

\newcolumntype{L}{>{$}l<{$}}

\renewcommand{\abstractname}{} 

\newcommand{\blstar}{\mathbf{!L^*}}

\newcommand{\id}{\mathrm{id}}
\newcommand{\ev}{\mathrm{ev}}
\newcommand{\Typ}{\mathrm{Typ}}

\newcommand{\C}{\mathcal{C}}
\newcommand{\D}{\mathcal{D}}
\newcommand{\F}{\mathcal{F}}

\newcommand{\R}{\mathbb{R}}
\newcommand{\fdVect}{\mathbf{FdVect}}
\newcommand{\Vect}{\mathbf{Vect}}

\newcommand{\Alg}{\mathbf{Alg}}

\newcommand{\bs}{\backslash}
\renewcommand{\L}{\mathbf{L}}

\newcommand{\ov}[1]{\overrightarrow{#1}}

\renewcommand{\epsilon}{\varepsilon}

\newcommand{\semantics}[1]{\llbracket #1 \rrbracket } 
\newcommand{\csemantics}[1]{(\!| #1 |\!)} 

\begin{document}

\title{Categorical Vector Space Semantics for\hfill\break Lambek Calculus with a Relevant Modality} %jg right-justified title looks terrible

\author{Lachlan McPheat}
\email{l.mcpheat@ucl.ac.uk}
\homepage{https://lmcpheat.github.io/CV/}
%\orcid{0000-0003-0290-4698}
%\thanks{You can use the \texttt{\textbackslash{}email}, \texttt{\textbackslash{}homepage}, and \texttt{\textbackslash{}thanks} commands to add additional information for the preceding \texttt{\textbackslash{}author}. If applicable, this can also be used to indicate that a work has previously been published in conference proceedings.}
\affiliation{Department of Computer Science, University College London, 66-72 Gower St, WC1E 6EA, London, UK} %jg4
	\author{Mehrnoosh Sadrzadeh}
\email{m.sadrzadeh@ucl.ac.uk}
\homepage{https://msadrzadeh.com}
%\orcid{0000-0003-0290-4698}
%\thanks{You can use the \texttt{\textbackslash{}email}, \texttt{\textbackslash{}homepage}, and \texttt{\textbackslash{}thanks} commands to add additional information for the preceding \texttt{\textbackslash{}author}. If applicable, this can also be used to indicate that a work has previously been published in conference proceedings.}
\affiliation{Department of Computer Science, University College London, 66-72 Gower St, WC1E 6EA, London, UK} %jg4
	\author{Hadi Wazni}
\email{hadi.wazni.20@ucl.ac.uk}
%\homepage{http://compositionality-journal.org}
%\orcid{0000-0003-0290-4698}
%\thanks{You can use the \texttt{\textbackslash{}email}, \texttt{\textbackslash{}homepage}, and \texttt{\textbackslash{}thanks} commands to add additional information for the preceding \texttt{\textbackslash{}author}. If applicable, this can also be used to indicate that a work has previously been published in conference proceedings.}
\affiliation{Department of Computer Science, University College London, 66-72 Gower St, WC1E 6EA, London, UK} %jg4
	\author{Gijs Wijnholds}
\email{g.wijnholds@uu.nl}
\homepage{https://gijswijnholds.github.io/organisation/}
%\orcid{0000-0003-0290-4698}
%\thanks{You can use the \texttt{\textbackslash{}email}, \texttt{\textbackslash{}homepage}, and \texttt{\textbackslash{}thanks} commands to add additional information for the preceding \texttt{\textbackslash{}author}. If applicable, this can also be used to indicate that a work has previously been published in conference proceedings.}
\affiliation{Institute for Language Sciences, Utrecht University, Trans 10, 3512 JK, Utrecht, NL} %jg4
\maketitle 
\begin{abstract}
We develop a categorical compositional distributional semantics for Lambek Calculus with a Relevant Modality $\blstar$, a modality that allows for the use of limited editions of contraction and permutation in the logic. Lambek Calculus has been introduced to analyse syntax of natural language and the linguistic motivation behind this modality is to extend the domain of the applicability of the calculus to fragments which witness the \emph{discontinuity} phenomena. The categorical part of the semantics is a monoidal biclosed category with a $!$-functor, very similar to the structure of a Differential Category. We instantiate this category to finite dimensional vector spaces and linear maps via ``quantisation'' functors and work with three concrete interpretations of the $!$-functor. We apply the model to construct categorical and concrete semantic interpretations for the motivating example of $\blstar$: the derivation of a phrase with a parasitic gap. The efficacy of the concrete interpretations are evaluated via a disambiguation task, on an extension of a sentence disambiguation dataset to parasitic gap phrase one, using BERT, Word2Vec, and FastText vectors and relational tensors. 
 %150-200 words
%\keywords{Vector Semantics \and Differential Category \and Relevant Modality \and Parasitic Gaps \and Linguistic Data \and Disambiguation.}
\end{abstract}
% end: P R E A M B L E ------------------------ P R E A M B L E ------------------------ P R E A M B L E ------------------------
\section{Introduction}
Distributional Semantics of natural language are semantics which model the \textit{Distributional Hypothesis} due to Firth \cite{Firth1957} and Harris \cite{Harris1954} which assumes \textit{a word is characterized by the company it keeps}. Research in Natural Language Processing (NLP) has turned to Vector Space Models (VSMs) of natural language to accurately model the distributional hypothesis. % see \cite{NLP overview} %TODO ADD REFERENCE for an overview.
Such models date as far back as to Rubinstein and Goodenough's co-occurence matrices \cite{RubinsteinGoodenough} in 1965, until today's neural machine learning methods, leading to embeddings, such as Word2Vec \cite{Word2Vec}, GloVe \cite{pennington2014glove}, FastText \cite{FastText} or BERT \cite{BERT} to name a few. VSMs were used even earlier by Salton \cite{Salton64} for information retrieval.
These models have plenty of applications, for instance thesaurus extraction tasks \cite{Curran2003,Grefenstette1994}, automated essay marking \cite{Landauer1997} and semantically guided information retrieval \cite{Manning2008}.
However, they lack grammatical compositionality, thus making it difficult to sensibly reason about the semantics of portions of language larger than words, such as phrases and sentences. 

Somewhat orthogonally, Type Logical Grammars (TLGs) form highly compositional models of language by accurately modelling grammar, however they lack distributionality, in that such models do not accurately describe the distributional semantics of a word, only its grammatical role.
Applications of type-logics with limited contraction and permutation to phenomena that witness discontinuities such as unbounded dependencies is a line of research initiated in \cite{Morrilletal1990,Barryetal1995}, with a later boost in \cite{morrill2016logic,Morrill2017,Morrill2018}, and also more recently in \cite{WijnSadr2019}. 
Distributional Compositional Categorical Semantics (DisCoCat)\cite{Coeckeetal2010} and the models of Preller \cite{preller2011,preller2014}, combine these two approaches using category theoretic methods, originally developed to model Quantum protocols. DisCoCat has proven its efficacy empirically \cite{GrefenSadrEMNLP,Grefenstette2015,Sadrzadeh2018,wijnholds-sadrzadeh-2019-evaluating,kartsaklis-sadrzadeh-2013-prior,milajevs-etal-2014-evaluating} and has the added utility of being a modular framework which is open to additions and extensions, such as modelling relative pronouns \cite{Sadretal2013Frob,Sadrzadeh2016}.

DisCoCat is a categorical semantics of a formal system which models natural language syntax, known as Lambek Calculus\footnote{There is a parallel pregroup syntax which gives you the same semantics, as discussed in \cite{Coeckeetal2013}}, denoted by $\L$. The work in \cite{Kanovich2016} extends Lambek calculus with a \textit{relevant modality}, and denotes the resulting logic by $\blstar$. 
Relevant (or relevance) logics were originally developed as a critique of the semantics of material implication \cite{curry_craig_1953}, where a false premise followed by a true conclusion should be regarded as true even if the statements have nothing to do with one another. 
The resulting logics, viewed proof-theoretically, form a family substructural logics which admit the structural rules of contraction, the rule %jg2 redundant "where"
allowing reuse of formulas, and permutation, the rule allowing for permuting formulas.
In \cite{Kanovich2016} the authors extend $\L$ with a modality, $!$, which allows for the use of \textit{controlled} versions of the structural rules of permutation and contraction (see Table \ref{Lsrules}), originally missing from the Lambek Calculus. As a motivating example for the introduction of this modality, the authors of \cite{Kanovich2016} apply the new logic to formalise the grammatical structure of a discontinuity phenomena referred to as \emph{parasitic gaps}. 

The first formal linguistic studies of parasitic gaps goes back to the work of Norwegian and Swedish linguists Taraldsen \cite{Taraldsen1979} and Engdahl \cite{Engdahl1983}. %jg2 other people don't get initials; anyway, Taraldsen is K
Phrases that witness this phenomenon %jg2 was "phenomena"
have multiple dependencies in them. In order to distinguish between single and multiple dependencies, consider the following relative phrase: %jg4 colon not full stop
\begin{quote}
articles$_{[\text{-}]}$ which I will file $[$-$]$
\end{quote}
Here, there is only one dependency, between the head of the phrase (\textit{articles}) and the object of the verb of the relative clause (\textit{file}). This phrase can almost immediately be extended to one that had two dependencies in it, e.g. by adding a prepositional phrase to it as follows: %jg4 colon
\begin{quote}
articles$_{[\text{-}]}$ which I will file $[$-$]$ without reading $[$-$]$
\end{quote}
Here, we have a dependency between the head of the phrase and the object of the main verb of the relative clause, and also a dependency between the object of the main verb and the object of the verb of the prepositional phrase within the relative clause. When analysing dependencies, linguists often denote the dependency with a blank mark $[$-$]$, it is this second of the blanks that is referred to as a gap and as a parasitic gap in phrases with multiple dependencies. The reason for this naming is as follows. In phrases with parasitic gaps, it is assumed that the second, third, and $\ldots$ blanks are parasitic, i.e. their existence relies on the existence of something that is itself a blank, and thus is parasitic: it lives on a dependency that is itself dependent on another part of the phrase. Phrases with parasitic gaps in them are often exemplified with two dependencies, i.e. only one parasitic gap, as in our above example. They can however contain more such gaps by adding prepositional or adverbial phrases, e.g. as in the following example which is an extension of our above example with an adverbial phrase:
\begin{quote}
 articles$_{[\text{-}]}$ which I will file $[$-$]$ without reading $[$-$]$ after just browsing $[$-$]$.
 \end{quote}
 Parasitic gaps can also be introduced directly into the argument of the verb, for example as done in the following phrase: %jg4 colon
 \begin{quote}
 articles$_{[\text{-}]}$ that I will persuade every author of ${[\text{-}]}$ to withdraw
 \end{quote}
 The work of Engdahl showed that these examples can become increasingly grammatically unacceptable. For instance, the following is deemed as ungrammatical by multiple traditional linguists, but tolerated by Engdahl:
 \begin{quote}
 authors$_{[\text{-}]}$ whom I will send a picture of ${[\text{-}]}$ to ${[\text{-}]}$
 \end{quote}
In this paper we only deal with the first kind, but as other type logical approaches to parasitic gaps, e.g. \cite{Steedman1987} and \cite{Moortgatetal2019} have shown other kinds of such gaps can be be analysed in similar ways once a type logic is used.

This paper is by no means the only venue where a type logic is used to analyse parasitic gaps. Firstly, our work uses the modal analysis of Kanovich et al.\ \cite{Kanovich2016}, which uses Lambek Calculus. Other work that uses extensions of Lambek Calculus with modalities to derive parasitic gaps in the logic is the work of Morrill and Valentin \cite{morrill2016logic,morrill2015computational}. Their analysis differs from that of Kanovich et al.\ in that their base logic is \emph{Displacement Calculus}, an extension of Lambek Calculus with a set of binary operations. These approaches all rely on syntactic copying. What our work has added to these treatments is the semantic part, in particular equipping the logic of Kanovich et al.\ with a vector space semantics and thus relating traditional work on type logical grammars with a more modern approach that use embeddings and machine learning. Syntactic copying is by no means the only type logical approach used for analysing parasitic gaps; soon after the work of Engdahl, the phenomena was analysed by Steedman in Combinatory Categorial Grammar (CCG) \cite{Steedman1987} using a new operation called \emph{substitution}, shorthanded as ${\bf S}$. More recently, Moortgat et al.\ \cite{Moortgatetal2019} showed that lexical polymorphism can be used as an alternative to syntactic copying to treat parasitic gaps. All of these approaches are amenable to over generation. The $\blstar$ and other syntactic copying approaches overcome this by restricting the lexicon, i.e. by assigning copyable and permutable types only to words that require them. The approach offered by CCG overcomes it by subjecting the ${\bf S}$ operator to rule features such as directional consistency and directional inheritance. The polymorphic type approach, both uses syntactic modalities in the lexicon and creates auxiliary semantic types that need to be learnt later via modern techniques such as machine learning. 

In view of further work, Engdahl observed parallels between acceptability of phrases with parasitic gaps in them and coreference of pronouns and bounded anaphors. In another work \cite{mcpheat2021}, we have shown that indeed the logic of this paper and its categorical vector space semantics can be applied to anaphora and other coreference phenomena known as VP-ellipsis. The Achilles heel of using $\blstar$ as a base logic for analysing complex linguistic phenomena such as the discontinuities discussed %jg2 was "discusses"
in this paper and co-referencing discussed in another paper \cite{mcpheat2021} is that the base logic is undecidable. This will not pose a problem for the examples discussed in the papers, but in the context of large scale wide coverage applications one needs to automate the parsing algorithm and would face infinite loops and nontermination. This criticism is not new and is the reason for the existence of bounded modality versions of linear logic such as \cite{Girard1992,Lafont2004}. In current work, we are exploring how $\blstar$ variants of these logics can be used for similar applications. 

In this paper, we first form a sound categorical semantics of $\blstar$, which we call $\C(\blstar)$. This boils down to interpreting the logical contraction of $\blstar$ using endofunctors which we call \textit{$!$-functors}, inspired largely by coalgebra modalities of differential categories defined in \cite{St2006}. 
In order to facilitate the categorical computations, we use the clasp-string calculus of \cite{BaezStay2011}, developed for depicting the computations of a monoidal biclosed category. To this monoidal diagrammatic calculus, we add the necessary new constructions for the coalgebra modality and its operations. Although not proven complete, this graphical language is sound, and makes visualising our work orders of magnitude %jg3 suggest "orders of magnitude" [Agreed]
easier.
Next, we define three candidate $!$-functors on the category of finite dimensional real vector spaces in order to form a sound VSM of $\blstar$ in terms of structure-preserving functors $\C(\blstar) \to \fdVect$. 
We conclude this paper with an experiment to test the accuracy of the different $!$-functors on $\fdVect$. The experiment is performed using different neural word embeddings and on a disambiguation task over an extended version of dataset of \cite{Grefenstette2015} from transitive sentences to phrases with parasitic gaps.

\subsection{Acknowledgements} %jg4 at end of introduction

This article is the full edition of the extended abstract \cite{mcpheatACT} which was accepted as a keynote presentation in the Applied Category Theory Conference (ACT) 2020, organised online at MIT, 6-10 July 2020.
% \url{https://act2020.mit.edu}. %jg4 don't need the URL
% The article is available in the arXiv via the code \verb#2005.03074#.
Part of the motivation behind this work came from the Dialogue and Discourse Challenge project of the Applied Category Theory adjoint school during the week 22–26 July 2019. We would like to thank the organisers of the school. We would also like to thank Adriana Correia, Alexis Toumi, and Dan Shiebler, for discussions. McPheat acknowledges support from the UKRI EPSRC Doctoral Training Programme scholarship, Sadrzadeh from the Royal Academy of Engineering Industrial Fellowship IF-192058. 

\section{\texorpdfstring{$\blstar$}{!L*}: Lambek Calculus with a Relevant Modality} %jg PDF alternative for maths
Following \cite{Kanovich2016}, we assume that the formulae of Lambek calculus $\L$ are generated by a set of atomic types $\mathrm{At}$, and three connectives, $\bs$, $/$ and $,$ via the following grammar. %jg3 suggest "the following grammar"; BNF isn't important [Fair enough, happy to use grammar]
	\[
		 \phi ::= \phi \in \mathrm{At} \mid (\phi,\phi) \mid (\phi \bs \phi) \mid (\phi/\phi).
	\]
We refer to the formulae of $\L$ as \textbf{types}, $\Typ_\L$, an element of which is thus either atomic, or is made up of %jg2 redundant "a"
two types joined by a comma or a slash. We will use uppercase Roman letters to denote arbitrary types of $\L$, and uppercase Greek letters to denote a list of types, for example, $\Gamma = \{A_1, A_2, \ldots, A_n\} = A_1, A_2, \ldots, A_n$. It is assumed that the comma %jg2 was $,$ 
is associative, allowing us to omit brackets in expressions like $A_1, A_2, \ldots, A_n$.
	
	A \textbf{sequent} of $\L$ is a pair of an ordered set of types and a type, denoted by $\Gamma \to A$. 
	The derivations of $\L$ are generated by the set of axioms and rules presented in Table \ref{Lrules}. 

	\begin{table}[H]
	\centering
	\begin{tabular}{rl}
    
    \prftree{A \to A} 
    
    & 
    
    \\
    \\
    
    \prftree[r]{$\scriptstyle{(/ L)}$}
    {\Gamma \to A}
    {\Delta_1, B, \Delta_2 \to C}
    {\Delta_1, B / A, \Gamma, \Delta_2 \to C}
    
    &
    
    \prftree[r]{$\scriptstyle{(/ R)}$}
    {\Gamma, A \to B}
    {\Gamma \to B/A}
    
    \\
    \\
    
    \prftree[r]{$\scriptstyle{(\bs L)}$}
    {\Gamma \to A}
    {\Delta_1, B, \Delta_2 \to C}
    {\Delta_1, \Gamma, A\bs B, \Delta_2 \to C}
    
    &
    
    \prftree[r]{$\scriptstyle{(\bs R)}$}
    {A, \Gamma \to B}
    {\Gamma \to A \bs B}
   \\
   &
	\end{tabular}
	\caption{Rules of $\L$.}\label{Lrules}
	\end{table}

The logic $\blstar$ extends $\L$ by endowing it with a modality denoted by $!$, inspired by the exponential modality, $!$, of Linear Logic, to enable the structural rule of contraction in a controlled way, albeit here it is introduced on a non-commutative base and has an extra property that allows the $!$-types to be permuted. This logic was introduced in \cite{Kanovich2016}. The types of $\blstar$ are generated via the following grammar: %jg3 suggest "grammar" again
	\[
		 \phi ::= \phi \in At \mid \emptyset \mid (\phi, \phi) \mid (\phi /\phi) \mid (\phi\bs \phi) \mid {!\phi} . %jg4 full stop
	\]
We refer to the types of $\blstar$ by $\Typ_{\blstar}$	; here, $\emptyset$ denotes the empty type. The set of rules of $\blstar$ consists of the rules of $\L$ and the five rules of Table \ref{Lsrules}. 
	\begin{table}[H]
   \centering
   \begin{tabular}{ll}
    \prftree[r]{$\scriptstyle{(!L)}$}
    {\Gamma_1, A, \Gamma_2 \to C}
    {\Gamma_1, !A, \Gamma_2\to C}
    
    &
    \prftree[r]{$\scriptstyle{(!R)}$}
    {!A_1,\ldots, !A_n \to B}
    {!A_1,\ldots, !A_n \to !B}
    
    \\
    \\
    
    \prftree[r]{$\scriptstyle{(\mathrm{perm}_1)}$}{\Delta_1, !A,\Gamma, \Delta_2\to C}{\Delta_1, \Gamma, !A, \Delta_2 \to C}
    
    &
    \prftree[r]{$\scriptstyle{(\mathrm{perm}_2)}$}{\Delta_1, \Gamma, !A, \Delta_2 \to C}{\Delta_1, !A,\Gamma, \Delta_2\to C}
    
    \\ 
    \\
    
    &
    
    \prftree[r]{$\scriptstyle{(\mathrm{contr})}$}
    {\Delta_1, !A, !A, \Delta_2 \to C}
    {\Delta_1, !A, \Delta_2 \to C}
    \\
    &
	\end{tabular}
	\caption{Additional rules in $\blstar$.}\label{Lsrules}
\end{table}
	
	It is common to distinguish the comma %jg was "$,$" - is this what you meant? [Yes, thanks for clarifying!]
and another connective $\bullet$ along with $(\bullet L), (\bullet R)$ rules, however we present this calculus in the same style as in \cite{Kanovich2016}, where no such distinction is made. In our applications there is no practical difference between including a $\bullet$ connective and the presentation in Table \ref{Lrules}, since we interpret $,$ as a monoidal product (see Section \ref{sec:catsem}). In the same lines, $\emptyset$ is the unit of the monoidal product, hence the sequents $\{\emptyset, A\}$ and $\{A,\emptyset\}$ are abbreviations for the formula $\{A\}$. Also note that $\blstar$ does admit the cut rule, see \cite{Kanovich2016}.

\section{Categorical Semantics for \texorpdfstring{$\blstar$}{!L*}} %jg PDF alternative for maths
\label{sec:catsem}

	Following the formalisms set out in \cite{mellies2009} we provide a passage from logic to category theory by interpreting $\blstar$ as a category $\C(\blstar)$. It is conventional to denote such an interpretation using semantic brackets, allowing us the shorthand $\semantics{\,}:\blstar \to \C(\blstar)$ meaning a categorical semantics of $\blstar$. 
	We will go on to show that $\C(\blstar)$ has a particular categorical structure, and we show how to define sound semantics of $\blstar$ in any suitable category $\D$ as a functor $\C(\blstar)\to \D$.
	This is essentially a standard construction of \cite{Johnstone}, but we have included far greater detail in the style of \cite{mellies2009} as we will make practical use of the semantics, and are not concerned with the model theory of $\blstar$ in this paper. We will however need a notion of a class of models, which we will call $\blstar$-categories, defined in the following.

	\begin{definition}\label{def:blstarCategories}
	A $\blstar$-\textbf{category} is a monoidal biclosed category $\C = (\C, \otimes, I, \Rightarrow, \Leftarrow)$ equipped with a lax monoidal endofunctor $F = (F,m)$, equipped with, for lack of a better name, \textit{precomonadic}\footnote
	{One could simply say that $F$ is a monoidal comonad $(F,\delta,\epsilon)$ which is not required to satisfy the comonad equations (and $\delta$ is not necessarily monoidal).} structure $ (F, \delta, \epsilon)$, where $\delta:F\to F^2$ and $\epsilon :F \to 1_\C$ are natural transformations and $\epsilon$ is monoidal. We finally require $F$ to have a $\C$-indexed family of \textbf{copying maps} $(\Delta_A : FA \to FA \otimes FA)_{A\in \C}$ (not necessarily natural). We also require that $F$ commutes with $1_{\C}$, that is, we have a natural isomorphism $\sigma: F \otimes 1_{\C} \stackrel{\sim}{\longrightarrow} 1_{\C} \otimes F$.
\end{definition}

We note that the definition of the $!$-functor $F$ in definition \ref{def:blstarCategories} is reminiscent of \textit{coalgebra modalities} of differential categories \cite{St2006}. Although a differential category is symmetric monoidal, that structure is recovered by the fact that $F$ commutes with $1_{\C}$. This suggests that one may be able to find further categorical models of $\blstar$ in terms of differential categories, although this has not yet been studied.

An important example of a $\blstar$-category is the category of finite dimensional real vector spaces, $\fdVect$. The monoidal structure of $\fdVect$ comes from the tensor product, and the monoidal biclosure $V\Rightarrow W$ is defined to be the set of linear maps from $V$ to $W$ (which in turn is isomorphic to $V^*\otimes W$). Note that $V\Rightarrow W \cong W\Leftarrow V$ in $\fdVect$, since the tensor product is symmetric. There are several choices of $!$-functor for $\fdVect$ which we discuss at length in Section \ref{sec:VSS}. Next, we define an important $\blstar$-category, namely the syntactic category of $\blstar$.

\begin{definition}\label{def:cblstar}
	We define the category $\C(\blstar)$ as a \textbf{syntactic category of} $\blstar$, roughly speaking, this means that $\C(\blstar)$ has formulas of $\blstar$ as objects, and derivations of $\blstar$ as morphisms. 
	We know that $\C(\blstar)$ is a monoidal biclosed category, as $\blstar$ contains $\L$ whose categorical semantics is exactly a monoidal biclosed category, as shown e.g. in \cite{Coeckeetal2013}. The operations of this category are $\otimes$ for the monoidal tensor with $I$ its unit, $\Rightarrow$ and $\Leftarrow$ for its two closures. 
	More precisely, we define a translation $\semantics{\,}$ from $\blstar$ to a category $\C(\blstar)$ inductively on formulas and derivations as follows:
	
	\[\begin{array}{rccl}
\mbox{Empty Sequent}& \quad \semantics{\emptyset} &:=& C_{\emptyset} = I\\
\mbox{Atomic Formulae}& \quad \semantics{\phi} &:=& C_{\phi}\\
\mbox{Pairs of Formulae}& \quad \semantics{(\phi, \phi)} &:=& (C_{\phi} \otimes C_{\otimes})\\
\mbox{Backward Slash}& \quad \semantics{(\phi \bs \phi)} &:=&(C_{\phi} \Rightarrow C_{\phi})\\
\mbox{Forward Slash}& \quad \semantics{(\phi / \phi)} &:=& (C_{\phi} \Leftarrow C_{\phi})\\
\mbox{Relevant Modality}& \quad \semantics{!\phi} &:=& F (C_\phi)
\end{array}\]

	We make use of the symbols $C_\phi$ as opposed to just $\phi$ for pedagogical reasons, to distinguish between $\blstar$-formulas as objects of a category or as just formulas of the logic. Formally, there is no distinction.
	
	Next, we translate derivations of $\blstar$ to morphisms of $\C(\blstar)$ inductively, in the style of \cite{mellies2009}. 
	Given a derivation $\pi$ of a sequent $\Gamma \to A$, we translate this to a morphism $\semantics{\pi}: \semantics{\Gamma} \to \semantics{A}$. The structure of $\semantics{\pi}$ comes from the following inductive definition.
	\begin{enumerate}
		\item The axiom of $\blstar$ 
			\[\infer[ax]{A\to A}{} \]
		is interpreted as the identity arrow:
		\[ \id_{\semantics{A}} : \semantics{A} \to \semantics{A}.\]
		
		\item The $(\bs L)$ and $(/L)$-rules of $\blstar$ are interpreted using the evaluation maps internal to $\C(\blstar)$. Consider the $(\bs L)$-rule where the two sequents $\Gamma \to A$ and $\Delta_1, B, \Delta_2 \to C$ have derivations $\pi$ and $\pi'$ respectively, which in turn are interpreted as morphisms $f: \semantics{\Gamma} \to \semantics{A}$ and $g: \semantics{\Delta_1}\otimes \semantics{B}\otimes \semantics{\Delta_2} \to \semantics{C}$ of $\C(\blstar)$. 
		 We interpret the derivation:
			\[\infer[(\bs L)]{\Delta_1, \Gamma, A\bs B, \Delta_2 \to C}{\infer[]{\Gamma \to A}{\deduce{\vdots}{\pi}} & \infer[]{\Delta_1, B, \Delta_2 \to C}{\deduce{\vdots}{\pi'}}}
			\]
		as the morphism $^fg : \semantics{\Delta_1}\otimes \semantics{\Gamma} \otimes \semantics{A}\Rightarrow \semantics{B} \otimes \semantics{\Delta_2} \to \semantics{C}$ in the following way:
		\[^fg := g \circ (\id_{\semantics{\Delta_1}} \otimes \ev^l_{\semantics{A},\semantics{B}}\otimes \id_{\semantics{\Delta_2}}) \circ (\id_{\semantics{\Delta_1}} \otimes f \otimes \id_{\semantics{A}\Rightarrow \semantics{B} \otimes \semantics{\Delta_2}} ).\]
		This is of course a morphism of the form $\semantics{\Delta_1}\otimes\semantics{\Gamma}\otimes\semantics{A}\Rightarrow\semantics{B}\otimes\semantics{\Delta_2}\to \semantics{C}$.
		The interpretation of $(/L)$ is very similar.

		\item The interpretations of the $(\bs R)$ and $(/R)$-rules are given by the tensor-hom adjunction. That is, given a derivation $\pi$ of the antecedent $A, \Gamma \to B$, with categorical semantics $f : \semantics{A}\otimes\semantics{\Gamma} \to \semantics{B}$ we have 
			\[\infer[(\bs R) .]{\Gamma \to A\bs B} %jg4 full stop
			{\infer[]{A,\Gamma \to B}{\deduce{\vdots}{\pi}}} 
			\] 
			Thus we interpret the preceding derivation as $\Lambda^l(f) : \semantics{\Gamma} \to \semantics{A} \Rightarrow \semantics{B}$ where $\Lambda^l(f) := (\semantics{\Gamma} \Rightarrow f) \circ \eta_{\semantics{\Gamma}}$ is the transpose of $f$ under the tensor-left hom adjunction (i.e. $\eta_{\semantics{\Gamma}}$ is the unit of the tensor-left hom adjunction). 
			The interpretation of the $(/R)$-rule is very similar.
		
		\item We interpret the $(!L)$-rule using the counit of $F$, letting the antecedent sequent $\Gamma_1, A, \Gamma_2 \to B$ have derivation $\pi$, with categorical semantics $f:\semantics{\Gamma_1}\otimes \semantics{A}\otimes\semantics{\Gamma_2} \to \semantics{B}$, the derivation
			\[
				\infer[(!L)]{\Gamma_1, !A, \Gamma_2 \to B}
					{\infer[]{\Gamma_1, A, \Gamma_2 \to B}{\deduce{\vdots}{\pi}}}
			\]
		is given the categorical semantics $B_l(f) := f \circ (\id_{\Gamma_1}\otimes \epsilon_A \otimes \id_{\Gamma_2}) : \semantics{\Gamma_1}\otimes F\semantics{A}\otimes\semantics{\Gamma_2} \to \semantics{B}$. 
		
		\item The $(!R)$-rule is interpreted using the lax monoidality of $F$ (we call it $m$ here) and the comonadic comultiplication ($\delta$) assuming that the sequent $!A_1,\ldots, !A_n \to B$ has derivation $\pi$, and categorical semantics $f: F\semantics{A_1}\otimes F\semantics{A_2} \otimes \cdots \otimes F\semantics{A_n} \to \semantics{B}$, the derivation
			\[
			\infer[(!R)]{!A_1,\ldots, !A_n \to !B}
				{\infer[]{!A_1,\ldots, !A_n \to B}{\deduce{\vdots}{\pi}}}
			\]
			is given the categorical semantics $B_r(f) :F\semantics{A_1}\otimes F\semantics{A_2} \otimes \cdots \otimes F\semantics{A_n} \to F\semantics{B}$, where 
			\[B_r(f) := F(f) \circ m_{F(\semantics{A_1})\otimes F(\semantics{A_2}) \otimes \cdots \otimes F(\semantics{A_n})}\circ (\delta_{\semantics{A_1}} \otimes \delta_{\semantics{A_2}} \otimes \cdots \otimes \delta_{\semantics{A_n}}).\]
		
		\item The $(perm_1)$ and $(perm_2)$-rules are interpreted using the natural isomorphism $\sigma: F \otimes 1_{\C} \stackrel{\sim}{\longrightarrow} 1_{\C} \otimes F$. Assuming that the antecedent sequent of the $(perm_1)$-rule, $\Delta_1,!A,\Gamma, \Delta_2 \to C$, has derivation $\pi$ and categorical semantics $f : \semantics{\Delta_1}\otimes F\semantics{A} \otimes \semantics{\Gamma} \otimes \semantics{\Delta_2} \to \semantics{C}$, the derivation
			\[
			\infer[(perm_1)]{\Delta_1,\Gamma, !A, \Delta_2 \to C}
				{\infer[]{\Delta_1,!A,\Gamma, \Delta_2 \to C}{\deduce{\vdots}{\pi}}}
			\]
		has semantics $\sigma_1(f) : \semantics{\Delta_1}\otimes \semantics{\Gamma} \otimes F\semantics{A} \otimes \semantics{\Delta_2} \to \semantics{C}$ defined to be $\sigma_1(f) := f \circ (\id_{\Delta_1} \otimes \sigma_{\semantics{\Gamma},\semantics{A}} \otimes \id_{\Delta_2})$. The interpretation of $(perm_2)$ is very similar, and we leave it to the reader to write it out for themselves.
		
		\item The $(contr)$-rule is interpreted using the $\Delta$-maps of definition \ref{def:blstarCategories}. Assuming the antecedent sequent $\Delta_1, !A, !A, \Delta_2 \to B$ has derivation $\pi$ and categorical semantics $f: \semantics{\Delta_1}\otimes F\semantics{A} \otimes F\semantics{A} \otimes \semantics{\Delta_2} \to \semantics{B}$, the derivation
			\[
				\infer[(contr)]{\Delta_1, !A, \Delta_2 \to B}
				{\infer{\Delta_1, !A, !A, \Delta_2 \to B}
				{\deduce{\vdots}{\pi}}}
			\]
			has categorical semantics $C(f) : \semantics{\Delta_1}\otimes F\semantics{A} \otimes \semantics{\Delta_2} \to \semantics{B}$ given by $C(f) := f\circ (\id_{\Delta_1}\otimes \Delta_A \otimes \id_{\Delta_2})$.
		
	\end{enumerate}
\end{definition}

	We can abstract the structures of $\C(\blstar)$ as defined above to obtain the type of categories defined in defined in definition \ref{def:blstarCategories}. Using this fact, we can define a functorial categorical semantics for $\blstar$.

%\noindent
%We now define a categorical semantics for $\blstar$ as the map $\semantics{\ } \colon \blstar \to \C(\blstar)$ and prove that it is sound. 
%
%\begin{definition}
%\label{def:sem}
%The \emph{semantics} of formulae and sequents of \ $\blstar$ is the image of the interpretation map $\semantics{\ } \colon \blstar \to \C(\blstar)$. To elements $\phi$ in $\Typ_\L$, this map assigns objects $C_\phi$ of $\C(\blstar)$, as defined below:
%\[\begin{array}{rccl}
%\mbox{Empty Sequent}& \quad \semantics{\emptyset} &:=& C_{\emptyset} = I\\
%\mbox{Atomic Formulae}& \quad \semantics{\phi} &:=& C_{\phi}\\
%\mbox{Pairs of Formulae}& \quad \semantics{(\phi, \phi)} &:=& C_{\phi} \otimes C_{\otimes}\\
%\mbox{Backward Slash}& \quad \semantics{(\phi \bs \phi)} &:=&(C_{\phi} \Rightarrow C_{\phi})\\
%\mbox{Forward Slash}& \quad \semantics{(\phi / \phi)} &:=& (C_{\phi} \Leftarrow C_{\phi})\\
%\mbox{Relevant Modality}& \quad \semantics{!\phi} &:=& ! C_\phi
%\end{array}\]
%To the sequents $\Gamma \to A$ of $\blstar$, for $\Gamma = \{A_1, A_2, \cdots A_n\}$ where $A_i, A \in \Typ_\L$, it assigns morphism of $\C(\blstar)$ as follows:
%\[
%\semantics{\Gamma \to A} := C_\Gamma \longrightarrow C_A \]
%for $C_\Gamma = \semantics{A_1} \otimes \semantics{A_2} \otimes \cdots \otimes \semantics{A_n}$. 
%\end{definition}
%
%\noindent
%Since sequents are not labelled, we have no obvious name for the linear map $\semantics{\Gamma \to A}$, so we will label such morphisms by lower case roman letters as needed.
%

\begin{definition}
\label{def:model}
A \textbf{functorial model} for $\blstar$ %or a \textbf{$\blstar$-model}, 
	is a structure preserving functor $M:\C(\blstar) \to \D$ where $\D$ is a $\blstar$-category.
%pair $(\C, \semantics{ \ }_\C)$, where $\C$ is a monoidal biclosed category with a lax monoidal coalgebraic modality and restricted symmetry, and $\semantics{ \ }_\C $ is a mapping $\Typ_{\blstar} \to \C$ factoring through $\semantics{ \ } : \Typ_\L \to \C(\blstar)$. 

%A \emph{categorical model} for $\blstar$ is the tuple $(\C(\blstar), \semantics{\ })$, for $\C(\blstar)$ as in Definition \ref{def:cblstar} and $\semantics{\ }$ as in Definition \ref{def:sem}. 
\end{definition}

\begin{definition}\label{def:interpretation}
	A \textbf{categorical model} of $\blstar$ in a $\blstar$-category $\D$ is a mapping of formulas and derivations of $\blstar$ to objects and morphisms of $\D$. We denote an interpretation as an arrow $\csemantics{\,}: \blstar \to \D$, and require the interpretations of complex types of $\blstar$ to satisfy the following equations:
	\begin{eqnarray*}
		\csemantics{A,B} & =& \csemantics{A} \otimes_\D \csemantics{B}\\
		\csemantics{A\bs B} & =& \csemantics{A} \Rightarrow_\D \csemantics{B}\\
		\csemantics{A/B} & =& \csemantics{A} \Leftarrow_\D \csemantics{B}\\
		\csemantics{!A} &=& F\csemantics{A}.
	\end{eqnarray*}
	We also require the $\csemantics{\,}$-interpretations of derivations be defined as for that of $\semantics{\,}$ in definition \ref{def:cblstar}.
	In other words, we require what we call a categorical model to be a sound model of $\blstar$.
\end{definition}

	A more condensed, but less practical way of describing $\C(\blstar)$ is to define it as the free category over a $\blstar$-signature. This definition follows that in Chapter D of \cite{Johnstone}, and lets us classify categorical models in $\D$ of $\blstar$ canonically as functors $\C(\blstar) \to \D$. Although this is a known result, again found in Chapter D of \cite{Johnstone}, we briefly sketch the proof in this particular case, namely that $\C(\blstar)$ is (isomorphic to) the free category over $\blstar$ in the following sense. 
	
	\begin{proposition}
	Functorial models of $\blstar$ are in bijection with categorical models of $\blstar$. 
	That is, given a $\blstar$-category $\D$ and an interpretation of $\csemantics{\,}: \blstar \to \D$ there is a unique functor $\C(\blstar)\to \D$ factoring through $\semantics{\,}$. 
	\begin{equation}\label{eqn:freetriangle}
		\xymatrix{
		\blstar \ar[r]^{\semantics{\,}} \ar[rd]_{\csemantics{\,}}& \C(\blstar) \ar@{-->}[d]^{T_{\csemantics{\,}}}\\ 
		& \D}
	\end{equation}
	\end{proposition}
	\begin{proof}[Sketch]
		Given the interpretation $\csemantics{\,}: \blstar \to \D$ we define a functor $T_{\csemantics{\,}}: \C(\blstar) \to \D$ as:
		\begin{eqnarray*}
			T_{\csemantics{\,}}(\semantics{A}) := \csemantics{A} . %jg4 full stop
		\end{eqnarray*}
		To show that $T_{\csemantics{\,}}$ is unique we simply suppose there is some other functor $T:\C(\blstar) \to \D$ which makes \eqref{eqn:freetriangle} commute. Then $T(\semantics{A}) = \csemantics{A} = T_{\csemantics{\,}}(\semantics{A})$, so $T$ and $T_{\csemantics{\,}}$ are identical on objects. 
			We also see that $T$ and $T_{\csemantics{\,}}$ agree on morphisms since we require $\csemantics{\,}$ to be sound in definition \ref{def:interpretation}. We recommend the motivated reader to confirm that $T$ and $T_{\csemantics{\,}}$ agree on morphisms, and omit this part of the proof here. %\qedhere
	\end{proof}

	The very particular nature of the structures surrounding $!$-functors in Definition \ref{def:blstarCategories} comes from first mistakenly identifying such functors as coalgebra modalities. Pacaud Lemay 
showed in \cite{JS2020} that one of the intended models of $!$-functors in Section \ref{sec:concrete} was in fact not a comonad, but our model still worked as intended, suggesting that requiring $!$-functors to be comonads was too strong a requirement. After carefully translating the sequents corresponding to the structure diagrams of a monoidal comonad, we realised that the structure of $!$-functors is not necessarily monoidal comonadic, but weaker as laid out in Definition \ref{def:blstarCategories}. Clearly one may consider monoidal comonads/coalgebra modalities as models of $!$-functors, but these will necessarily be more expressive than $\blstar$, making those models incomplete. Of course, we are not too concerned with incompleteness in terms of vector space semantics, but when defining syntactic categories this is a great concern.

	To summarise, we now have a simple way to define semantics of $\blstar$ in terms of structure-preserving functors $\C(\blstar) \to \D$. The inductive interpretation in Definition \ref{def:cblstar} will serve as a practical guide to defining vector space semantics in the following section.

\section{Vector Space Semantics for \texorpdfstring{${\cal C}(\blstar)$}{C(!L*)}}\label{sec:VSS} %jg PDF alternative for maths

	Following \cite{Coeckeetal2013}, we develop vector space semantics for $\blstar$, by defining a functorial $\blstar$-model in $\fdVect$. This is known to some as a \emph{quantisation} functor, first introduced by Atiyah in Topological Quantum Field Theory, as a functor from the category of manifolds and cobordisms to the category of vector spaces and linear maps. Since the category of cobordisms is monoidal, quantisation was later generalised to refer to a functor that `quantises' any category in $\fdVect$.
		
	To define a vector space semantics we first need to provide the $\blstar$-category structures on $\fdVect$, most importantly we have to define a $!$-functor.
	As we noted earlier $\fdVect$ already is a monoidal biclosed category, and it is in fact symmetric, so for any choice of $!$-functor $F$ on $\fdVect$ we already know that $F\otimes 1_\fdVect \cong 1_\fdVect \otimes F$. Hence defining vector space semantics really boils down to simply specifying what our $!$-functors should be, and in doing so, determining what their diagonal maps should look like.
	Although quantisations are simply instances of functorial models, we still explicitly show what a quantisation looks like below, to allow us to use it practically in examples later.
	
\begin{definition}
\label{def:quantF}
A \textbf{quantisation} is a closed monoidal functor $Q : \C(\blstar) \to (\fdVect,F)$, defined on the objects of $\C(\blstar)$ using the structure of the formulae of $\blstar$, as follows: 
	\begin{align*}
		Q(C_{\emptyset}) &:= \R \\
		Q(C_\phi) &:= V_\phi\\
   	Q(C_{\phi, \phi}) & := V_\phi \otimes V_\phi \\
   	Q(C_{\phi \bs \phi}) &:= (V_\phi \Rightarrow V_\phi) \\
   	Q(C_{\phi / \phi}) &:=(V_\phi \Leftarrow V_\phi) \\
		Q (C_{!\phi}) &:=F(V_\phi) . %jg4 full stop
	\end{align*}
Here, $V_\phi$ is the vector space in which vectors of words with an atomic type live and the other vector spaces are obtained from it by induction on the structure of the formulae they correspond to. Morphisms of $\C(\blstar)$ are of the form $C_\Gamma \longrightarrow C_A$, associated with proofs of sequents $\Gamma \to A$ of $\blstar$, for $\Gamma = \{A_1, A_2, \cdots, A_n\}$. The quantisation functor is defined on these morphisms as follows: 
\[
Q(C_{\Gamma} \longrightarrow C_A) := Q(C_\Gamma) \longrightarrow Q(C_A) = V_{A_1} \otimes V_{A_2} \otimes \cdots \otimes V_{A_n} \longrightarrow V_A
\]

\end{definition}	
	
	Recall that given two vector spaces $V,W\in \fdVect$, the set $V\Rightarrow W$ is the space of linear maps from $V$ to $W$, which in turn is isomorphic to $V^*\otimes W$. Since the monoidal product in $\fdVect$ is symmetric, there is formally no need to distinguish between $(\semantics{A} \Rightarrow \semantics{B})$ and $(\semantics{B} \Leftarrow \semantics{A})$. However it may be practical to do so when doing calculations by hand, for example when retracing derivations in the semantics. 
	We should also make clear that the freeness of $\C(\blstar)$ makes $F$ a \emph{strict} monoidal closed functor, meaning that $F(C_A \otimes C_B) = FC_A \otimes FC_B$, or rather, $V_{(A\otimes B)} = (V_A \otimes V_B)$, and similarly, $V_{(A\Rightarrow B)} = (V_A \Rightarrow V_B)$ etc.
		Further, since we are working with finite dimensional vector spaces we know that $V_\phi^\bot \cong V_\phi$, thus our internal homs have an even simpler structure, which we exploit when computing, which is $V_\phi \Rightarrow V_\phi \cong V_\phi \otimes V_\phi$.

	Next, we define three different $!$-functors on $\fdVect$, providing three functorial models of $\blstar$ in $\fdVect$.

\section{Concrete Constructions}
\label{sec:concrete}

	In this section we introduce the three candidate $!$-functors on $\fdVect$ that we have identified. The first construction follows a more classical approach whose goal is to find models of full linear logic \cite{Blute1994}. The latter constructions use the identity functor as the underlying endofunctor, on top of which we define diagonal maps.
	
	It is also worth noting that at a high level, the goal of the diagonal maps of a $!$-functor, $\Delta:F\to F\otimes F$ is to \textit{copy}. Although copying is clearly a nonlinear operation, we try to approximate nonlinear copying with these diagonal maps.

\subsection{\texorpdfstring{$!$}{!} as the Dual of an Algebra} %jg PDF alternative for maths
\label{sec:FSpace}
Following \cite{Blute1994} we interpret $!$ using the Fermionic Fock space functor $\F : \fdVect \to \Alg_\R$. In order to define $\F$ we first introduce the simpler free algebra construction, typically studied in the theory of representations of Lie algebras \cite{Humphreys}. This construction is applicable to all vector spaces, which are not necessarily finite dimensional. 
	The choice of the symbol $\F$ comes from ``Fermionic Fock space'' (as opposed to ``Bosonic''), but is also known as the exterior algebra functor, or the Grassmannian algebra functor \cite{Humphreys}. In the following by ``\textit{algebra}'' we mean associative algebra. That is, a vector space with a ring structure and appropriate relations between scalars and multiplication. These are objects of the category $\Alg_\R$. The morphisms of this category are linear maps that moreover preserve the ring-multiplication.

\begin{definition}
\label{def:T}
The {\bf free algebra functor} $T : \Vect_\R \to \Alg_\R$ is defined on objects as:
	\[ V \longmapsto \bigoplus_{n\geq 0} V^{\otimes n} = \R \oplus V \oplus (V \otimes V) \oplus (V\otimes V \otimes V) \otimes \cdots \]
and for morphisms $f : V \to W$, we get the algebra homomorphism $T(f) : T(V) \to T(W)$ defined layer-wise as
	\[ T(f) (v_1 \otimes v_2 \otimes \cdots \otimes v_n) := f(v_1) \otimes f(v_2) \otimes \cdots \otimes f(v_n).\]
\end{definition}

The algebra structure on $T(V)$ is given by concatenation. That is, given elements $v_1\otimes v_2 \otimes \cdots \otimes v_n$ and $v_{n+1}\otimes \cdots v_m$ their product is $v_1 \otimes \cdots \otimes v_n \otimes v_{n+1}\otimes \cdots \otimes v_m$. The multiplicative unit is simply $1\in \R$.

$T$ is free in the sense that it is left adjoint to the forgetful functor $U : \Alg_\R \to \Vect_\R$, thus giving us a monad $UT$ on $\Vect_\R$ with a monoidal algebra modality structure, i.e. the dual of what we are looking for.

However note that even when restricting $T$ to finite dimensional vector spaces $V \in \fdVect$ the resulting $UT(V)$ and $UT(V^\bot)^\bot$ are infinite-dimensional. The necessity of working in $\fdVect$ motivates us to use $\F$, defined below, rather than $T$.

\begin{definition}
\label{def:F}
The {\bf Fermionic Fock space functor} $\F : \Vect_\R \to \Alg_\R$ is defined on objects as 
	\[V \mapsto \bigoplus_{n \geq 0} V^{\wedge n} = \R \oplus V \oplus (V \wedge V) \oplus (V\wedge V \wedge V) \oplus \cdots\]
where $V^{\wedge n}$ is the coequaliser of the family of maps $(-\tau_\sigma)_{\sigma \in S_n}$, defined as $-\tau_\sigma : V^{\otimes n} \to V^{\otimes n}$ and given as follows:
	\[(-\tau_{\sigma}) ( v_1 \otimes \cdots \otimes v_n) := \mathrm{sgn}(\sigma) (v_{\sigma(1)} \otimes v_{\sigma(2)} \otimes \cdots \otimes v_{\sigma(n)}).\]	
	$\F$ applied to linear maps gives an analogous algebra homomorphism as in \ref{def:T}. 	
\end{definition}

One may equivalently define $V^{\wedge n }$ as the $n$-fold tensor product of $V$ where we quotient by the equivalence relation $v_1 \otimes v_2 \otimes \cdots \otimes v_n \sim \mathrm{sgn}(\sigma)(v_{\sigma(1)} \otimes v_{\sigma(2)} \otimes \cdots \otimes v_{\sigma(n)})$ and denoting the equivalence class of a vector $v_1 \otimes v_2 \otimes \cdots \otimes v_n$ by $v_1 \wedge v_2 \wedge \cdots \wedge v_n$. 

In fact, we have a very similar multiplicative structure on $\F(V)$ to $T(V)$, namely given again by concatenation. As mentioned in our above reference on representation theory of Lie algebras \cite{Humphreys}, the difference is that $\F(V)$ is a free graded alternating algebra over $V$. We explain what this means by means of examples below; note that knowing that $\F(V)$ is graded alternating is not too important for the purposes of this paper. 

For some concrete examples of what elements of $\F(V)$ look like, consider a 3-dimensional real vector space $V$ spanned by the basis $\{u,v,w\}$. In $\F(V)$ we have vectors like $v, u\wedge v, 4v+u, w\wedge v + u$ etc. We also have that $v\wedge w = - w\wedge v$ and $u\wedge v \wedge w = w \wedge u \wedge v = - v \wedge u \wedge w$. 

Vectors like $v\wedge v$ or $u\wedge w\wedge w$ containing repeated factors, are always equal to $0$, since $v\wedge v = -v \wedge v$. More generally, in a space with basis $\{v_1,\ldots, v_n\}$, if $v_i = v_j$ for some $1 \leq i,j\leq n$ and $i\neq j$ in the above, the permutation $(ij) \in S_n$ has odd sign, and so $v_1 \wedge v_2 \wedge \cdots \wedge v_n = 0$, since $v_1 \wedge v_2 \wedge \cdots \wedge v_n = \mathrm{sgn}(ij) (v_1 \wedge v_2 \wedge \cdots \wedge v_n) = - (v_1 \wedge v_2 \wedge \cdots \wedge v_n)$. 

From the preceding remark about vectors in $\F(V)$ with repeated factors being zero, it follows that $\F(V)$ is in fact finite dimensional whenever $V$ is finite dimensional. This can be seen by a simple application of the pigeonhole principle. Suppose that $V$ has basis $\{v_1,\ldots, v_n\}$. Then, if we consider any basis vector in $\F(V)$ with $>n$ factors it must repeat at least one of the basis vectors. By the preceding remark, the whole vector is zero. Thus we have that $V^{\wedge m} = 0$ for all $m>n$, meaning that $\F(V) = \R \oplus V \oplus (V\wedge V) \oplus \cdots \oplus (V^{\wedge n})$.

Now that we know $\F$, we consider two ways to obtain a diagonal structure $\Delta_V$ on $U\F(V)$, thus defining two $!$-functors $(U\F, \delta, \epsilon,\Delta_V)$ on $\fdVect$, as desired. For both of these we use the same $\delta$ and $\epsilon$ maps, which are inclusion and projection respectively. That is, $\delta_V : \F(V) \to \F^2(V)$ includes vectors of $\F(V)$ into the first layer of $\F^2(v)$, and $\epsilon_V:\F(V) \to V$ projects the first layer of $\F(V)$ to $V$, that is
\[1, e_{i_1}, e_{i_2} \wedge e_{i_3}, e_{i_4} \wedge e_{i_5} \wedge e_{i_6}, \cdots \longmapsto e_{i_1}.\]

Next, we define the two diagonal maps on $\F$. One, referred to as a \emph{Cogebra} construction is given below, for a basis $\{e_i\}_i$ of $V$, and thus a basis $\{1, e_{i_1}, e_{i_2} \wedge e_{i_3}, e_{i_4} \wedge e_{i_5} \wedge e_{i_6}, \cdots\}_{i_j}$ of $U\F(V)$ as:
\[\begin{array}{l}
\Delta(1, e_{i_1}, e_{i_2} \wedge e_{i_3}, e_{i_4} \wedge e_{i_5} \wedge e_{i_6}, \cdots) 
\\ \qquad = %jg linebreak
(1, e_{i_1}, e_{i_2} \wedge e_{i_3}, e_{i_4} \wedge e_{i_5} \wedge e_{i_6}, \cdots)\otimes (1, e_{i_1}, e_{i_2} \wedge e_{i_3}, e_{i_4} \wedge e_{i_5} \wedge e_{i_6}, \cdots) . %jg4 full stop
\end{array}\]

The other diagonal map is a more classical comultiplication, which we may denote as $\mu^{-1}$ for now. This map sends a vector $v$ to the sum of all possible pairs of factors which when concatenated, produce $v$. Consider the 3-dimensional example from earlier. Here 
	\[\mu^{-1}(u\wedge v\wedge w) = u \otimes (v\wedge w) + (u\wedge v)\otimes w.\]

There are other ways to obtain this simpler construction, but one has to work with infinite dimensional vector spaces to take advantage of the isomorphism between polynomial rings $K[X]$ and symmetric algebras over vector spaces with basis $X$. In either case, the $\Delta$ structure is the dual of the algebra multiplication and its counit $\epsilon$ is obtained via the inclusion of constant polynomials. Another option is to change category and move to category of suplattices, where the dual of the additive algebra functor provides an example \cite{HylandSchalk}. In this setting, the comultiplication is the map that sends a monomial to the join of all pairs of monomials whose multiplication give the input monomial.

However, comultiplications on $\F(V)$ quickly increase in complexity and become difficult to instantiate in applications, since the dimensions of the domain and codomain are $2^n$ and $(2^n)^2$ we would have matrices representing the comultiplications of size $2^n \times (2^n)^2$, where $n$ is taken to be at the very least 100, and usually 300.

\subsection{\texorpdfstring{$!$}{!} as the Identity Functor} %jg PDF alternative for maths

The Cogebra construction can be simplified when one works with free vector spaces. Given a set $S$, consider the set functions $S \to \R$ mapping all but finitely many elements $s \in S$ to 0. It is easy to see that a scalar multiplication and addition can be defined on this set %jg2 redundant "using"
by defining them pointwise using the structure of $\R$. The set of such functions forms a vector space, known as the \emph{free vector space over $S$}. An element $s \in S$ can be identified with function $f_s : S \to \R$ defined by $f_s(s) = 1$ and $f_s(s') = 0$, for all $s' \in S, s \neq s'$, thus defining a vector space structure, denoted by $\R^S$. On such vector spaces, one can define a coalgebra structure by setting $\Delta(s) = s \otimes s$ and $\epsilon(s) = 1$. Because the functions $f_s$ are defined on $S$, the basis of $\R^S$, and send all but finitely many elements of $S$ to 0, the above comonoidal structure can be extended to all of $\R^S$ linearly. As one %jg3 recommend "as one can see" [Agreed!]
can see, this construction is not limited to finite dimensional spaces, but when working with the finite case, the condition that all but finitely many elements need to be sent to 0 can be dropped and the construction becomes simpler. This comonoid structure defines a $!$-functor (indeed even a coalgebra modality) over the identity comonad on $\fdVect$.

This version of the construction clearly resembles half of a bialgebra over $\fdVect$, known as \emph{Special Frobenius bialgebras}, which were used in \cite{Sadretal2013Frob,MoortgatWijn2017,Moortgatetal2019} to model relative pronouns in English and Dutch. As argued in \cite{WijnSadr2017}, however, the copying map resulting from this comonoid structure only copies the basis vectors and does not seem adequate for a full copying operation. A quick computation shows that this $\Delta$ copies only half of the input vector. In order to see this, consider a vector $\ov{v} = \sum_i C_i s_i$, for $s_i \in S$. Extending the comultiplication $\Delta$ linearly provides us with 
\[
\Delta(\ov{v}) = \sum_i C_i \Delta(s_i) = \sum_i C_i s_i \otimes s_i = (\sum_i C_i s_i ) \otimes (\sum_i s_i) = (\sum_i C_i s_i ) \otimes (\sum_i 1 s_i) . %jg4 full stop
\]
The result is one copy of $\ov{v}$, i.e. the $(\sum_i C_i s_i )$ part of the above, and a sum of its bases, i.e. the $\sum_i s_i$ part of the above. In the second term, we have lost the $C_i$ weights, in other words, we have replaced the second copy with a vector of 1's. 
 
The above problem can be partially overcome by observing that there are infinitely many linear copying structures on $\fdVect$, one for each element $k \in \R$. Formally speaking, over the vector space $V_\phi$ with a basis $(v_i)_i$, a \emph{Cofree-inspired} comonoid $(V_\phi, \Delta, e)$ is defined as follows: 
		\[
		\Delta: V_\phi \to V_\phi \otimes V_\phi :: v \mapsto (v \otimes \vec{k}) + (\vec{k} \otimes v), \quad 
		e: V_\phi \to \R :: \sum_i C_i v_i \mapsto \sum_i C_i . %jg4 full stop
		\]
Here, $\ov{v}$ is as before and $\vec{k}$ stands for an element of $V$ padded with number $k$. In the simplest case, when $k=1$, we obtain two copies of the weights $\ov{v}$ and also of its basis vectors, as the following calculation demonstrates. Consider a two dimensional vector space and the vector $a e_1 + b e_2$ in it. The 1 vector $\ov{1}$ is the 2-dimensional vector $e_1 + e_2$ in $V$. Suppose $\ov{v}$ and $\vec{1}$ are column vectors, then applying $\Delta$ results in the matrix $2a\, e_1\otimes e_1 + ab\, e_1 \otimes e_2 + ab\, e_2 \otimes e_1 + 2b\, e_2 \otimes e_2$, where we have two copies of the weights in the diagonal and also the basis vectors have obviously multiplied. 
%\[
%(\left ( \begin{array}{c} a\\b\end{array} \right) \otimes \left ( \begin{array}{c} 1\\1\end{array} \right) ) +
%(\left ( \begin{array}{c} 1\\1\end{array} \right) \otimes \left ( \begin{array}{c} a\\b\end{array} \right) ) = 
%\left ( \begin{array}{cc} a & a\\ b & b\end{array} \right) + \left ( \begin{array}{cc} a & b\\a & b\end{array} \right) 
%\]

This construction is inspired by the graded algebra construction on vector spaces, whose dual construction is referred to as a \emph{Cofree coalgebra}. The Cofree-inspired coalgebra over a vector space defines a second $!$-functor (and again also a coalgebra modality) structure over the identity comonad on $\fdVect$, which provides another $\blstar$-model, or rather, another quantisation $\C(\blstar) \to \fdVect$.

\section{Clasp Diagrams}
\label{sec:claspDiagrams}
	\usetikzlibrary{arrows,decorations,backgrounds,positioning,fit}
	\tikzset{func/.style={shape=rectangle,rounded corners=8,minimum width=2cm,minimum height=.5cm,draw}}
	\tikzset{claspnode/.style={shape=circle,minimum width=0.25cm,fill=white,draw}}

In order to display the semantic computations for the parasitic gap, we introduce a diagrammatic notation. The derivation of the parasitic gap is involved and its categorical and vector space interpretations become far more legible using a diagrammatic language. In what follows we first introduce notation for the Clasp diagrams, then extend them with extras necessary to model the $!$ coalgebra modality. 

We adopt the usual notation of objects as labelled strings %jg3 should that be "labelled strings"? [Agreed!]
and morphisms as boxes.
\[
{%
\beginpgfgraphicnamed{obsHoms}
\begin{tikzpicture}
	\begin{pgfonlayer}{nodelayer}
		\node [style=none] (0) at (-1, 1) {};
		\node [style=none] (1) at (-1, -1) {};
		\node [style=title] (2) at (-1, 1.5) {$id_A$};
		\node [style=title] (3) at (-1.5, 0) {$A$};
		\node [style=none] (4) at (1, 1) {};
		\node [style=none] (5) at (1, -1) {};
		\node [style=title] (6) at (1, 1.5) {$f:A \to B$};
		\node [style=title] (7) at (1.5, 0.75) {$A$};
		\node [func, minimum width=1.3cm] (8) at (1, 0) {$f$};
		\node [style=title] (9) at (1.5, -0.75) {$B$};
	\end{pgfonlayer}
	\begin{pgfonlayer}{edgelayer}
		\draw [style=downArrow] (4.center) to (8);
		\draw [style=downArrow] (8) to (5.center);
		\draw [style=downArrow] (0.center) to (1.center);
	\end{pgfonlayer}
\end{tikzpicture}}
\endpgfgraphicnamed}
\]
Composition of morphisms is simply vertical juxtaposition. The monoidal product and its internal left and right homs are depicted below. This is where clasps of \cite{BaezStay2011} are introduced, to depict the left and right internal hom-sets of the category.

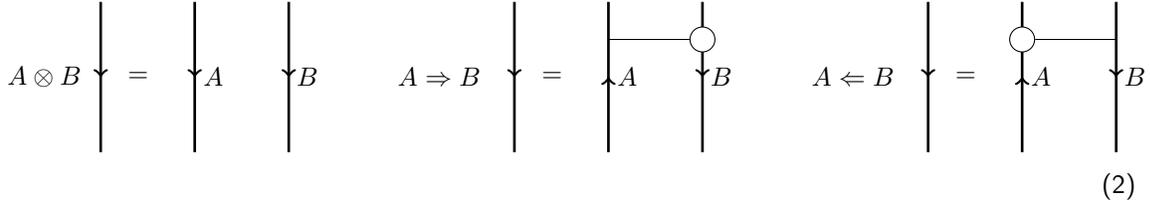
\[
{%
\beginpgfgraphicnamed{tensorsClasps-narrow}
\begin{tikzpicture}
	\begin{pgfonlayer}{nodelayer}
		\node [style=none] (0) at (-5.25, 0.75) {};
		\node [style=none] (1) at (-5.25, -0.75) {};
		\node [style=title] (2) at (-4.75, 0) {$=$};
		\node [style=none] (3) at (-4, 0.75) {};
		\node [style=none] (4) at (-4, -0.75) {};
		\node [style=none] (5) at (-2.75, 0.75) {};
		\node [style=none] (6) at (-2.75, -0.75) {};
		\node [style=title] (7) at (-6, 0) {$A\otimes B$};
		\node [style=title] (8) at (-3.75, 0) {$A$};
		\node [style=title] (9) at (-2.5, 0) {$B$};
		\node [style=none] (10) at (0, 0.75) {};
		\node [style=none] (11) at (0, -0.5) {};
		\node [style=title] (12) at (0.5, 0) {$=$};
		\node [style=none] (13) at (1.25, 0.75) {};
		\node [style=none] (14) at (1.25, -0.75) {};
		\node [style=none] (15) at (2.5, 0.75) {};
		\node [style=none] (16) at (2.5, -0.75) {};
		\node [style=title] (17) at (-1, 0) {$A\Rightarrow B$};
		\node [style=title] (18) at (1.5, 0) {$A$};
		\node [style=title] (19) at (2.75, 0) {$B$};
		\node [style=none] (20) at (1.25, 0.25) {};
		\node [style=new style 1] (21) at (2.5, 0.25) {};
		\node [style=none] (22) at (5.25, 0.75) {};
		\node [style=none] (23) at (5.25, -0.75) {};
		\node [style=title] (24) at (5.75, 0) {$=$};
		\node [style=none] (25) at (6.5, -0.75) {};
		\node [style=none] (26) at (6.5, 0.75) {};
		\node [style=none] (27) at (7.75, -0.75) {};
		\node [style=none] (28) at (7.75, 0.75) {};
		\node [style=title] (29) at (4.25, 0) {$A \Leftarrow B$};
		\node [style=title] (30) at (6.75, 0) {$A$};
		\node [style=title] (31) at (8, 0) {$B$};
		\node [style=none] (32) at (7.75, 0.25) {};
		\node [style=new style 1] (33) at (6.5, 0.25) {};
	\end{pgfonlayer}
	\begin{pgfonlayer}{edgelayer}
		\draw [style=downArrow] (0.center) to (1.center);
		\draw [style=downArrow] (3.center) to (4.center);
		\draw [style=downArrow] (5.center) to (6.center);
		\draw [style=downArrow] (10.center) to (11.center);
		\draw [style=downArrow] (14.center) to (13.center);
		\draw (20.center) to (21);
		\draw [style=downArrow] (22.center) to (23.center);
		\draw [style=downArrow] (27.center) to (28.center);
		\draw (32.center) to (33);
		\draw (26.center) to (33);
		\draw (15.center) to (21);
		\draw [style=downArrow] (33) to (25.center);
		\draw [style=downArrow] (21) to (16.center);
	\end{pgfonlayer}
\end{tikzpicture}}
\endpgfgraphicnamed} %jg tighter packing
\]

There are special diagrams for the biclosed structure of the category, i.e. its left and right tensor-hom adjunctions, or Currying, which are depicted below. 

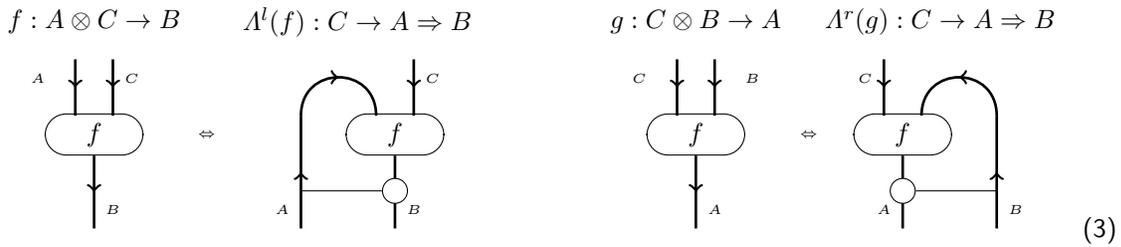
\[
{%
\beginpgfgraphicnamed{caps}
\begin{tikzpicture}
	\begin{pgfonlayer}{nodelayer}
		\node [style=none] (0) at (-6.25, 1) {};
		\node [style=none] (1) at (-5.75, 1) {};
		\node [func, minimum width=1.3cm] (2) at (-6, 0) {$f$};
		\node [style=none] (3) at (-6, -1.25) {};
		\node [style=title] (4) at (-6, 1.5) {$f: A \otimes C \to B$};
		\node [style=none] (5) at (-4.5, 0) {$\Leftrightarrow$};
		\node [func, minimum width=1.3cm] (7) at (-2, 0) {$f$};
		\node [style=none] (8) at (-2, -1.25) {};
		\node [style=none] (9) at (-3.25, -1.25) {};
		\node [style=none] (10) at (-3.25, 0.25) {};
		\node [style=none] (11) at (-3.25, -0.75) {};
		\node [style=new style 1] (12) at (-2, -0.75) {};
		\node [style=title] (13) at (-2.5, 1.5) {$\Lambda^l(f) : C \to A \Rightarrow B$};
		\node [style=none] (14) at (-6.75, 0.75) {$A$};
		\node [style=none] (15) at (-5.5, 0.75) {$C$};
		\node [style=none] (16) at (-5.75, -1) {$B$};
		\node [style=none] (17) at (-1.5, 0.75) {$C$};
		\node [style=none] (18) at (-3.5, -1) {$A$};
		\node [style=none] (19) at (-1.75, -1) {$B$};
		\node [style=none] (20) at (1.75, 1) {};
		\node [style=none] (21) at (2.25, 1) {};
		\node [func, minimum width=1.3cm] (22) at (2, 0) {$f$};
		\node [style=none] (23) at (2, -1.25) {};
		\node [style=title] (24) at (2, 1.5) {$g: C \otimes B \to A$};
		\node [style=none] (25) at (3.5, 0) {$\Leftrightarrow$};
		\node [func, minimum width=1.3cm] (27) at (4.75, 0) {$f$};
		\node [style=none] (28) at (4.75, -1.25) {};
		\node [style=none] (29) at (6, -1.25) {};
		\node [style=none] (30) at (6, 0.25) {};
		\node [style=none] (31) at (6, -0.75) {};
		\node [style=new style 1] (32) at (4.75, -0.75) {};
		\node [style=title] (33) at (5.25, 1.5) {$\Lambda^r(g) : C \to A \Rightarrow B$};
		\node [style=none] (34) at (1.25, 0.75) {$C$};
		\node [style=none] (35) at (2.75, 0.75) {$B$};
		\node [style=none] (36) at (2.25, -1) {$A$};
		\node [style=none] (37) at (4.25, 0.75) {$C$};
		\node [style=none] (38) at (4.5, -1) {$A$};
		\node [style=none] (39) at (6.25, -1) {$B$};
		\node [style=none] (40) at (-6.25, 0.25) {};
		\node [style=none] (41) at (-5.75, 0.25) {};
		\node [style=none] (42) at (1.75, 0.25) {};
		\node [style=none] (43) at (2.25, 0.25) {};
		\node [style=none] (45) at (-1.75, 1) {};
		\node [style=none] (46) at (-2.25, 0.25) {};
		\node [style=none] (47) at (-1.75, 0.25) {};
		\node [style=none] (48) at (4.5, 1) {};
		\node [style=none] (50) at (4.5, 0.25) {};
		\node [style=none] (51) at (5, 0.25) {};
	\end{pgfonlayer}
	\begin{pgfonlayer}{edgelayer}
		\draw [style=downArrow] (2) to (3.center);
		\draw [style=downArrow] (9.center) to (10.center);
		\draw [style=downArrow] (7) to (8.center);
		\draw (11.center) to (12);
		\draw [style=downArrow] (22) to (23.center);
		\draw [style=downArrow] (29.center) to (30.center);
		\draw [style=downArrow] (27) to (28.center);
		\draw (31.center) to (32);
		\draw [style=downArrow] (0.center) to (40.center);
		\draw [style=downArrow] (1.center) to (41.center);
		\draw [style=downArrow] (20.center) to (42.center);
		\draw [style=downArrow] (21.center) to (43.center);
		\draw [style=downArrow] (45.center) to (47.center);
		\draw [style=downArrow, bend left=90, looseness=1.75] (10.center) to (46.center);
		\draw [style=downArrow] (48.center) to (50.center);
		\draw [style=downArrow, bend left=270, looseness=1.75] (30.center) to (51.center);
	\end{pgfonlayer}
\end{tikzpicture}}
\endpgfgraphicnamed}
\]

Finally, the left and right evaluation morphisms of the category, also coming from the biclosed structure, depicted graphically as directed cups, are as follows:

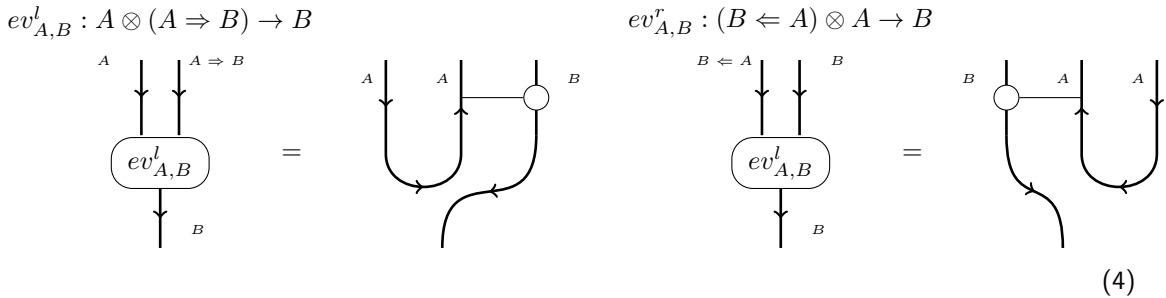
\[
{%
\beginpgfgraphicnamed{cups-narrow}
\begin{tikzpicture}
	\begin{pgfonlayer}{nodelayer}
		\node [style=title] (0) at (-5.75, 1.75) {$ev_{A,B}^l : A \otimes (A\Rightarrow B) \to B$};
		\node [func, minimum width=1.3cm] (3) at (-5.75, -0.12) {$ev_{A,B}^l$};
		\node [style=none] (4) at (-5.75, -1.25) {};
		\node [style=none] (5) at (-3.25, 1.25) {};
		\node [style=none] (6) at (-2.25, 1.25) {};
		\node [style=none] (7) at (-1.25, 0.25) {};
		\node [style=none] (8) at (-3.25, 0) {};
		\node [style=none] (9) at (-2.25, 0) {};
		\node [style=none] (11) at (-2.5, -1.25) {};
		\node [style=new style 1] (12) at (-1.25, 0.75) {};
		\node [style=none] (13) at (-2.25, 0.75) {};
		\node [style=none] (14) at (-1.25, 1.25) {};
		\node [style=none] (15) at (-3.5, 1) {$A$};
		\node [style=none] (16) at (-2.5, 1) {$A$};
		\node [style=none] (17) at (-0.75, 1) {$B$};
		\node [style=title] (18) at (-4.25, 0) {$=$};
		\node [style=none] (19) at (-6.5, 1.25) {$A$};
		\node [style=none] (20) at (-5, 1.25) {$A\Rightarrow B$};
		\node [style=title] (21) at (2, 1.75) {$ev_{A,B}^r : (B \Leftarrow A) \otimes A \to B$};
		\node [func, minimum width=1.3cm] (24) at (2, -0.12) {$ev_{A,B}^l$};
		\node [style=none] (25) at (2, -1.25) {};
		\node [style=none] (26) at (6.5, 1.25) {};
		\node [style=none] (27) at (5.5, 1.25) {};
		\node [style=none] (28) at (4.5, 0.25) {};
		\node [style=none] (29) at (6.5, 0) {};
		\node [style=none] (30) at (5.5, 0) {};
		\node [style=none] (31) at (5.25, -1.25) {};
		\node [style=new style 1] (32) at (4.5, 0.75) {};
		\node [style=none] (33) at (5.5, 0.75) {};
		\node [style=none] (34) at (4.5, 1.25) {};
		\node [style=none] (35) at (6.25, 1) {$A$};
		\node [style=none] (36) at (5.25, 1) {$A$};
		\node [style=none] (37) at (4, 1) {$B$};
		\node [style=title] (38) at (3.5, 0) {$=$};
		\node [style=none] (39) at (2.75, 1.25) {$B$};
		\node [style=none] (40) at (1.25, 1.25) {$B \Leftarrow A$};
		\node [style=none] (41) at (1.75, 1.25) {};
		\node [style=none] (42) at (2.25, 1.25) {};
		\node [style=none] (43) at (1.75, 0.25) {};
		\node [style=none] (44) at (2.25, 0.25) {};
		\node [style=none] (45) at (-6, 1.25) {};
		\node [style=none] (46) at (-5.5, 1.25) {};
		\node [style=none] (47) at (-6, 0.25) {};
		\node [style=none] (48) at (-5.5, 0.25) {};
		\node [style=none] (49) at (-5.25, -1) {$B$};
		\node [style=none] (50) at (2.5, -1) {$B$};
	\end{pgfonlayer}
	\begin{pgfonlayer}{edgelayer}
		\draw [style=downArrow] (3) to (4.center);
		\draw [style=downArrow] (5.center) to (8.center);
		\draw [style=downArrow] (9.center) to (6.center);
		\draw (13.center) to (12);
		\draw [style=downArrow, bend right=90, looseness=1.50] (8.center) to (9.center);
		\draw [style=downArrow, in=90, out=-90, looseness=1.75] (7.center) to (11.center);
		\draw [style=downArrow] (14.center) to (7.center);
		\draw [style=downArrow] (24) to (25.center);
		\draw [style=downArrow] (26.center) to (29.center);
		\draw [style=downArrow] (30.center) to (27.center);
		\draw (33.center) to (32);
		\draw [style=downArrow, bend left=90, looseness=1.50] (29.center) to (30.center);
		\draw [style=downArrow, in=90, out=-90, looseness=1.75] (28.center) to (31.center);
		\draw [style=downArrow] (34.center) to (28.center);
		\draw [style=downArrow] (41.center) to (43.center);
		\draw [style=downArrow] (42.center) to (44.center);
		\draw [style=downArrow] (45.center) to (47.center);
		\draw [style=downArrow] (46.center) to (48.center);
	\end{pgfonlayer}
\end{tikzpicture}}
\endpgfgraphicnamed} %jg tighter packing
\]

To these, we first add a diagram for $!$-ed objects, as follows: 

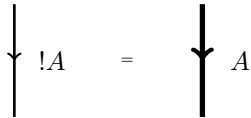
\[
{%
\beginpgfgraphicnamed{bangObs}
\begin{tikzpicture}
	\begin{pgfonlayer}{nodelayer}
		\node [style=none] (0) at (1, -0.75) {};
		\node [style=none] (1) at (1, 0.75) {};
		\node [style=title] (2) at (1.5, 0) {$A$};
		\node [style=none] (3) at (-1.5, -0.75) {};
		\node [style=none] (4) at (-1.5, 0.75) {};
		\node [style=title] (5) at (-1, 0) {$!A$};
		\node [style=none] (6) at (0, 0) {$=$};
	\end{pgfonlayer}
	\begin{pgfonlayer}{edgelayer}
		\draw [style=downArrow] (4.center) to (3.center);
		\draw [style=thickArr] (1.center) to (0.center);
	\end{pgfonlayer}
\end{tikzpicture}}
\endpgfgraphicnamed}
\]

We then equip the notation with necessary diagrams for the $!$-functor, that is the diagonal maps $\Delta$, the counit $\epsilon $ and the comultiplication $\delta$. These are respectively depicted by a triangle node, a filled black circle, and a white circle labelled with $\delta$, as follows.

\[
{%
\beginpgfgraphicnamed{bangDiagrams1}
\begin{tikzpicture}
	\begin{pgfonlayer}{nodelayer}
		\node [style=title] (8) at (-5, 1.5) {$!$-Functor Diagonal Map};
		\node [style=none] (9) at (-5, 1) {};
		\node [fill=white, draw=black, thick, shape=regular polygon, regular polygon sides=3] (10) at (-5, -0.25) {};
		\node [style=none] (11) at (-6, -1) {};
		\node [style=none] (12) at (-4, -1) {};
		\node [style=title] (13) at (0, 1.5) {$!$-Functor Counit};
		\node [style=none] (14) at (0, 1) {};
		\node [style=none] (15) at (0, -1) {};
		\node [style=counit] (16) at (0, 0) {};
		\node [style=title] (17) at (4, 1.5) {$!$-Functor Comultiplication};
		\node [style=none] (18) at (4, 1) {};
		\node [style=new style 1, font={\small}] (20) at (4, 0) {$\delta_A$};
		\node [style=none] (21) at (-4.5, 0.5) {$A$};
		\node [style=none] (22) at (-6.25, -0.75) {$A$};
		\node [style=none] (23) at (-3.75, -0.75) {$A$};
		\node [style=none] (24) at (0.5, 0.5) {$A$};
		\node [style=none] (25) at (0.5, -0.75) {$A$};
		\node [style=none] (26) at (4.75, 0.5) {$A$};
		\node [style=none] (27) at (4.75, -0.75) {$A$};
		\node [style=none] (29) at (4, 0) {};
		\node [style=none] (30) at (4, -0.5) {};
		\node [style=none] (31) at (3.975, 0) {};
		\node [style=none] (32) at (3.975, -1) {};
		\node [style=none] (33) at (4.025, 0) {};
		\node [style=none] (34) at (4.025, -1) {};
		\node [style=none] (35) at (4, -1) {};
	\end{pgfonlayer}
	\begin{pgfonlayer}{edgelayer}
		\draw [style=thickArr] (9.center) to (10);
		\draw [style=thickArr, in=90, out=-150] (10) to (11.center);
		\draw [style=thickArr, in=90, out=-30] (10) to (12.center);
		\draw [style=thickArr] (14.center) to (16);
		\draw [style=downArrow] (16) to (15.center);
		\draw [style=thickArr] (18.center) to (20);
		\draw [style=downArrow] (29.center) to (30.center);
		\draw (33.center) to (34.center);
		\draw (31.center) to (32.center);
		\draw [style=downArrow] (30.center) to (35.center);
	\end{pgfonlayer}
\end{tikzpicture}}
\endpgfgraphicnamed}
\]

\section{Linguistic Examples}
\label{sec:ling}

We now provide the interpretations of the motivating example of \cite{Kanovich2016} which was the parasitic gap ``the paper that John signed without reading''. We first briefly illustrate how to do this with two simpler related examples, and then take the reader through how to perform the same calculations on the full, considerably more complex example. 

\subsection{``John signed the papers''}
First off is the declarative sentence ``John signed the papers'', which uses the following lexicon:

\[\{
	(\text{John}, NP),	
	(\text{signed}, (NP\bs S)/NP),
	(\text{the}, NP/N),	
	(\text{papers}, N)
\}.\]

Showing that ``John signed the papers'' is a declarative sentence is now a question of whether the corresponding sequent, 
	\[ NP, (NP\bs S)/NP, NP/N, N \to S\]
is derivable in $\blstar$, or in this case just $\L$ is enough, since none of the types are modal. We derive the sequent below:

\[
		\infer[(/L)]{NP, (NP\bs S)/NP, NP/N, N \to S}
			{N \to N
			&
			\infer[(/L)]{NP, (NP\bs S)/NP, NP \to S}
				{NP \to NP
				&
				\infer[(\bs L)]{NP, NP\bs S \to S}
					{NP \to NP
					&
					S \to S}}}
\]

Next, we introduce the diagrammatic interpretation of the derivation. This diagram is drawn by first drawing the types as (loose) strings, forming the top of the diagram, and then tracing up the derivation and connecting the strings according to the definition in Section \ref{sec:claspDiagrams}.

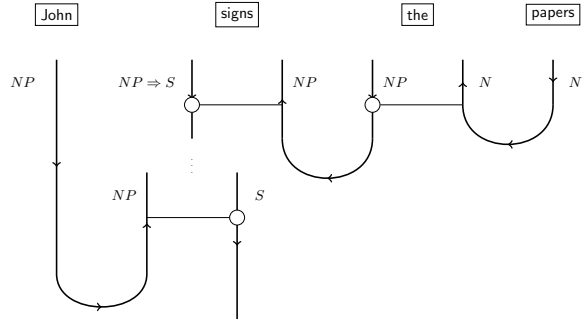
\[
	\scalebox{0.6}{{%
\beginpgfgraphicnamed{JohnSignsThePapers}
\begin{tikzpicture}
	\begin{pgfonlayer}{nodelayer}
		\node [style=none] (0) at (-4, 3) {};
		\node [style=none] (1) at (-1, 3) {};
		\node [style=none] (2) at (1, 3) {};
		\node [style=none] (3) at (3, 3) {};
		\node [style=none] (4) at (5, 3) {};
		\node [style=none] (5) at (7, 3) {};
		\node [style=new style 0] (6) at (-4, 4) {\strut John};
		\node [style=new style 0] (7) at (0, 4) {\strut signed};
		\node [style=new style 0] (8) at (4, 4) {\strut the};
		\node [style=new style 0] (9) at (7, 4) {\strut papers};
		\node [style=none] (10) at (-4, -1.75) {};
		\node [style=none] (11) at (-1, 1.25) {};
		\node [style=none] (12) at (1, 1.25) {};
		\node [style=none] (13) at (3, 1.25) {};
		\node [style=none] (14) at (5, 2) {};
		\node [style=none] (15) at (7, 2) {};
		\node [style=new style 1] (16) at (-1, 2) {};
		\node [style=new style 1] (17) at (3, 2) {};
		\node [style=none] (18) at (1, 2) {};
		\node [style=none] (19) at (5, 2) {};
		\node [style=none] (20) at (-1, 0.75) {$\VDOTS$};
		\node [style=none] (21) at (-2, 0.5) {};
		\node [style=none] (22) at (0, 0.5) {};
		\node [style=none] (23) at (-4.75, 2.5) {\small$NP$};
		\node [style=none] (24) at (-2, 2.5) {\small$NP \Rightarrow S$};
		\node [style=none] (25) at (1.5, 2.5) {\small$NP$};
		\node [style=none] (26) at (3.5, 2.5) {\small$NP$};
		\node [style=none] (27) at (5.5, 2.5) {\small$N$};
		\node [style=none] (28) at (7.5, 2.5) {\small$N$};
		\node [style=none] (30) at (-2, -1.75) {};
		\node [style=none] (31) at (0, -2.75) {};
		\node [style=new style 1] (32) at (0, -0.5) {};
		\node [style=none] (33) at (-2, -0.5) {};
		\node [style=none] (34) at (-2.5, 0) {\small$NP$};
		\node [style=none] (35) at (0.5, 0) {};
		\node [style=none] (36) at (0.5, 0) {\small$S$};
	\end{pgfonlayer}
	\begin{pgfonlayer}{edgelayer}
		\draw [style=downArrow] (5.center) to (15.center);
		\draw [style=downArrow] (14.center) to (4.center);
		\draw [style=downArrow] (3.center) to (13.center);
		\draw [style=downArrow] (12.center) to (2.center);
		\draw [style=downArrow] (1.center) to (11.center);
		\draw [style=downArrow] (0.center) to (10.center);
		\draw (19.center) to (17);
		\draw (18.center) to (16);
		\draw [style=downArrow, bend left=90, looseness=1.50] (15.center) to (14.center);
		\draw [style=downArrow, bend left=90, looseness=1.50] (13.center) to (12.center);
		\draw (33.center) to (32);
		\draw [style=downArrow, bend right=90, looseness=1.25] (10.center) to (30.center);
		\draw [style=downArrow] (30.center) to (21.center);
		\draw [style=downArrow] (22.center) to (31.center);
	\end{pgfonlayer}
\end{tikzpicture}}
\endpgfgraphicnamed}} %jg struts in boxes
\]

Note here that the use of vertical dots %jg3 those dots are very faint! [I know, have tried to thicken them]
is to denote the equality of a string labelled with a complex type, and a clasp with simpler types i.e. the second equality in Section \ref{sec:claspDiagrams}. 

We end this example by computing the interpretation of the sequent for ``John signed the papers''. We do this in a general setting without choosing a specific VSS. This interpretation is of course a linear map
	\[i: 
	\semantics{NP} \otimes \semantics{(NP\bs S)/NP} \otimes \semantics{NP/N} \otimes \semantics{N} 
	\longrightarrow \semantics{S}
	\]
or equivalently, upon expanding the domain, a linear map of type
	\[i:
	\semantics{NP} \otimes ((\semantics{NP} \Rightarrow \semantics{S})\Leftarrow \semantics{NP}) \otimes (\semantics{NP}\Leftarrow \semantics{N}) \otimes \semantics{N} 
	\longrightarrow 
	\semantics{S}.
	\]
By inspecting the diagram we see that the $i$ will evaluate its arguments as:
	\[ %jg \textit not \mathit
        i: 
	(\overrightarrow{\textit{John}} \otimes \textit{signed}(-,-) \otimes \textit{the}(-) \otimes \overrightarrow{\textit{papers}})
	\longmapsto 
	\textit{signed}(\overrightarrow{\textit{John}}, \textit{the}(\overrightarrow{\textit{papers}})).
	\]
We often use the names of the types i.e. the \emph{words} as variable names to remind us what type the variables have. The parentheses and hyphens denote functional types. One could demonstrate the effect of this interpretation morphism using more short-hand variable names, but for longer examples this becomes difficult to keep track of. However, later when using more complex types we will have to use more explicit variable names which involve the various bases of the vector spaces they live in.

\subsection{``The papers that John signed''}
The second example shows that ``The papers that John signed'' is an English noun phrase, which is proved using the following lexicon:
\[\{
	(\text{the}, NP/N),	
	(\text{papers}, N),	
	(\text{that}, (N\bs N)/(S/!NP)),	
	(\text{John}, NP),	
	(\text{signed}, (NP\bs S)/NP)
\}.\]

This is, again, equivalent to deriving the corresponding sequent:
	\[NP/N, N, (N\bs N)/(S/!NP), NP, (NP\bs S)/NP \to NP,\]

which may be done as follows:
\[
	\infer[(/L)]{NP/N, N, (N\bs N)/(S/!NP), NP, (NP\bs S)/NP \to NP}
		{\infer[(\bs L)]{NP/N, N, N\bs N \to NP}
			{N \to N
			&
			\infer[(/L)]{NP/N, N \to NP}
				{N \to N
				&
				NP \to NP}}
		&
		\infer[(/ R)]{NP, (NP\bs S) / NP \to S/!NP}
			{\infer[(!L)]{NP, (NP\bs S) / NP, !NP \to S}
				{\infer[(/L)]{NP, (NP\bs S) / NP, NP \to S}
					{NP \to NP
					&
					\infer[(\bs L)]{NP, NP\bs S \to S}
						{NP \to NP 
						&
						S \to S}}}}}
\]

This proof gives rise to the following diagram:
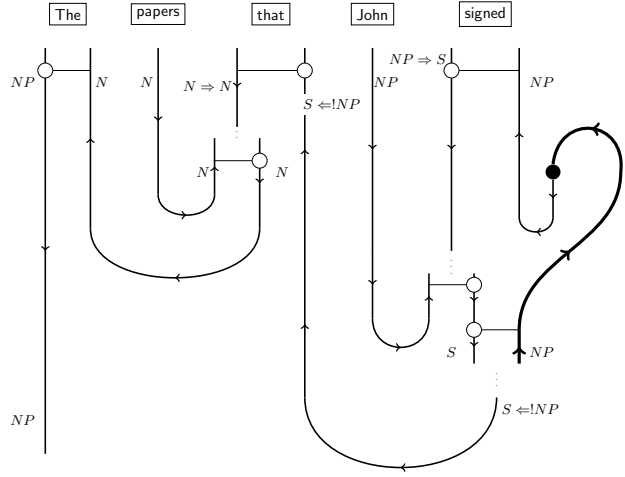
\[\scalebox{0.6}{{%
\beginpgfgraphicnamed{papersThatJohnSigned}
\begin{tikzpicture}
	\begin{pgfonlayer}{nodelayer}
		\node [style=none] (0) at (1.5, 3.25) {};
		\node [style=none] (1) at (3.25, 3.25) {};
		\node [style=none] (2) at (4.75, 3.25) {};
		\node [style=none] (5) at (1.5, -1.25) {};
		\node [style=none] (6) at (3.25, -1.25) {};
		\node [style=none] (7) at (4.75, -0.5) {};
		\node [style=new style 1] (11) at (3.25, 2.75) {};
		\node [style=none] (13) at (4.75, 2.75) {};
		\node [style=none] (14) at (1.75, 2.5) {\, \small $NP$};
		\node [style=none] (15) at (5.25, 2.5) {\small$NP$};
		\node [style=none] (18) at (2.5, 3) {\small $NP\Rightarrow S$};
		\node [style=counit] (19) at (5.5, 0.5) {};
		\node [style=none] (20) at (5.5, -0.5) {};
		\node [style=none] (43) at (-5.75, 3.25) {};
		\node [style=none] (44) at (-4.75, 3.25) {};
		\node [style=none] (45) at (-5.75, -5.75) {};
		\node [style=none] (46) at (-4.75, -0.75) {};
		\node [style=none] (47) at (-4.75, 2.75) {};
		\node [style=new style 1] (48) at (-5.75, 2.75) {};
		\node [style=none] (49) at (-1, 1.25) {};
		\node [style=none] (50) at (-2, 1.25) {};
		\node [style=none] (51) at (-1, -0.75) {};
		\node [style=none] (52) at (-2, 0) {};
		\node [style=none] (53) at (-2, 0.75) {};
		\node [style=new style 1] (54) at (-1, 0.75) {};
		\node [style=none] (55) at (-6.25, 2.5) {\small $NP$};
		\node [style=none] (56) at (-4.5, 2.5) {\small $N$};
		\node [style=none] (57) at (-2.25, 0.5) {\small $N$};
		\node [style=none] (58) at (-0.5, 0.5) {\small $N$};
		\node [style=none] (59) at (-3.25, 3.25) {};
		\node [style=none] (60) at (-3.25, 0) {};
		\node [style=none] (61) at (-3.5, 2.5) {\small $N$};
		\node [style=none] (62) at (-1.5, 1.75) {};
		\node [style=none] (63) at (-1.5, 1.5) {$\VDOTS$};
		\node [style=none] (64) at (-1.5, 3.25) {};
		\node [style=none] (65) at (-2.15, 2.4) {\small $N \Rightarrow N$};
		\node [style=none] (70) at (0, -1.25) {};
		\node [style=none] (71) at (0, 3.25) {};
		\node [style=new style 1] (72) at (0, 2.75) {};
		\node [style=none] (73) at (-1.5, 2.75) {};
		\node [style=lab] (74) at (0.6, 2) {\small $S \Leftarrow !NP$};
		\node [style=none] (75) at (7, 0.5) {};
		\node [style=none] (76) at (3.75, -1.75) {};
		\node [style=none] (77) at (2.75, -1.75) {};
		\node [style=none] (79) at (3.25, -1.5) {$\VDOTS$};
		\node [style=none] (80) at (3.75, -3) {};
		\node [style=none] (81) at (2.75, -2.75) {};
		\node [style=none] (82) at (2.75, -2) {};
		\node [style=new style 1] (83) at (3.75, -2) {};
		\node [style=none] (84) at (3.75, -3.75) {};
		\node [style=none] (85) at (4.75, -3.75) {};
		\node [style=none] (86) at (4.25, -4) {$\VDOTS$};
		\node [style=none] (87) at (4.25, -4.5) {};
		\node [style=none] (88) at (4.75, -3) {};
		\node [style=none] (89) at (1.5, -2.75) {};
		\node [style=none] (90) at (4.75, -3) {};
		\node [style=new style 1] (91) at (3.75, -3) {};
		\node [style=none] (92) at (0, -4.5) {};
		\node [style=none] (93) at (5.25, -3.5) {\small $NP$};
		\node [style=none] (94) at (3.25, -3.5) {\small $S$};
		\node [style=none] (95) at (5, -4.75) {\small $S \Leftarrow !NP$};
		\node [style=new style 0] (96) at (-5.25, 4) {\strut The};
		\node [style=new style 0] (97) at (-3.25, 4) {\strut papers};
		\node [style=new style 0] (98) at (-0.75, 4) {\strut that};
		\node [style=new style 0] (99) at (1.5, 4) {\strut John};
		\node [style=new style 0] (100) at (4, 4) {\strut signed};
		\node [style=none] (101) at (-6.25, -5) {\small $NP$};
	\end{pgfonlayer}
	\begin{pgfonlayer}{edgelayer}
		\draw [style=downArrow] (7.center) to (2.center);
		\draw [style=downArrow] (1.center) to (6.center);
		\draw [style=downArrow] (0.center) to (5.center);
		\draw (13.center) to (11);
		\draw [style=downArrow] (19) to (20.center);
		\draw [style=downArrow, bend left=90, looseness=1.50] (20.center) to (7.center);
		\draw (47.center) to (48);
		\draw [style=downArrow] (46.center) to (44.center);
		\draw [style=downArrow] (43.center) to (45.center);
		\draw (53.center) to (54);
		\draw [style=downArrow] (52.center) to (50.center);
		\draw [style=downArrow] (49.center) to (51.center);
		\draw [style=downArrow] (59.center) to (60.center);
		\draw [style=downArrow, bend right=90, looseness=1.25] (60.center) to (52.center);
		\draw [style=downArrow, bend left=90] (51.center) to (46.center);
		\draw [style=downArrow] (64.center) to (63.center);
		\draw [style=downArrow] (70.center) to (71.center);
		\draw (73.center) to (72);
		\draw (82.center) to (83);
		\draw [style=downArrow] (76.center) to (80.center);
		\draw [style=downArrow] (80.center) to (84.center);
		\draw [style=thickArr] (85.center) to (88.center);
		\draw [style=thickArr, in=-90, out=90] (88.center) to (75.center);
		\draw [style=thickArr, bend left=270, looseness=2.00] (75.center) to (19);
		\draw [style=downArrow] (5.center) to (89.center);
		\draw [style=downArrow, bend right=90, looseness=1.75] (89.center) to (81.center);
		\draw [style=downArrow] (81.center) to (77.center);
		\draw (90.center) to (91);
		\draw [style=downArrow, bend left=90, looseness=1.25] (87.center) to (92.center);
		\draw [style=downArrow] (92.center) to (70.center);
	\end{pgfonlayer}
\end{tikzpicture}}
\endpgfgraphicnamed}}\] %jg struts in boxes

To conclude the example, we compute the interpretation of this derivation/diagram. This is a linear map of type 
\begin{align*}
	i :
	(\semantics{NP} {\Leftarrow} \semantics{N}) \otimes
	\semantics{N} \otimes 
	((\semantics{N}{\Rightarrow} \semantics{N})\Leftarrow(\semantics{S} {\Leftarrow} \semantics{!NP})) \otimes 
	\semantics{NP} \otimes
	((\semantics{NP} {\Rightarrow} \semantics{S}) \Leftarrow \semantics{NP}) 
\\ %jg linebreak, and reduce spacing around some arrows
	\longrightarrow
	\semantics{NP} . %jg4 full stop
\end{align*}
We again form the interpretation linear map by reading the diagram from top to bottom, and in this case, from left to right. We see that ``the'' is being applied to the result of the application of ``that'' to ``papers'' and the rest of the right side of the diagram. The right side of the diagram is more confusing. Here we see that ``signed'' is being applied to ``John'' on the left and curried on the right. Putting this together we have the following linear map:
	\[ %jg \textit, not \mathit
	i :
	\textit{The} (-) \otimes \overrightarrow{\textit{papers}} \otimes \textit{that}(-,-) \otimes \overrightarrow{\textit{John}} \otimes \textit{signed}(-,-) 
	\longmapsto 
	\textit{The}(\textit{that}(\overrightarrow{\textit{papers}},\textit{signed}(\overrightarrow{\textit{John}},-))).
	\]

Note that we have not made use of contraction or permutation at any point in this proof. This sequent is derivable in $\mathbf{L}$ if you re-type ``that'' to $(N\bs N) /(S/NP)$, however we do not do so to illustrate where contraction will be applied when we deal with parasitic gaps in the following example.

\subsection{``The papers that John signed without reading''}
We now prove that ``The papers that John signed without reading'' is indeed a noun phrase. We extend the previous example's lexicon to:
\begin{eqnarray*}
&&	\left \{ (\mbox{The}, NP \bs N), (\mbox{paper}, N), (\mbox{that}, (N\bs N)/(S/!NP)), (\mbox{John}, NP), (\mbox{signed}, (NP\bs S)/NP), \right. \\
 && \left.(\mbox{without}, ((NP\bs S)\bs(NP \bs S))/NP), (\mbox{reading}, NP / NP)\right \}.
\end{eqnarray*}

The $\blstar$ derivation of ``the paper that John signed without reading'' is as follows: %jg4 colon
\[
\infer[(/ L)]
{NP/N, N, (N\bs N)/(S/!NP), NP, (NP \bs S)/NP, ((NP \bs S) \bs (NP \bs S))/NP, NP / NP \to NP}
{\infer[(/R)]{NP, (NP \bs S)/NP, ((NP \bs S) \bs (NP \bs S))/NP, NP / NP \to S/!NP}{\mathbf{D_l}} & \mathbf{D_r} }
\]
where $\mathbf{D_l}, \mathbf{D_r}$ are the following derivations (formatted to fit on page).
\[
\mathbf{D_l} =
\infer[(contr)]{NP, (NP \bs S)/NP, ((NP \bs S) \bs (NP \bs S))/NP, NP / NP, !NP \to S}
	{\infer[(perm_1)]{NP, (NP \bs S)/NP, ((NP \bs S) \bs (NP \bs S))/NP, NP / NP, !NP, !NP \to S}
		{\infer[(perm_1)]{NP, (NP \bs S)/NP, ((NP \bs S) \bs (NP \bs S))/NP, !NP, NP / NP, !NP \to S}
			{\infer[(!L)]{NP, (NP \bs S)/NP, !NP, ((NP \bs S) \bs (NP \bs S))/NP, NP / NP, !NP \to S}
				{\infer[(!L)]{NP, (NP \bs S)/NP, NP, ((NP \bs S) \bs (NP \bs S))/NP, NP / NP, !NP \to S}
					{\infer[(/L)]{NP, (NP \bs S)/NP, NP, ((NP \bs S) \bs (NP \bs S))/NP, NP / NP, NP \to S}
						{\infer[]{NP \to NP}{}
						&
						\infer[(/ L)]{NP, (NP \bs S)/NP, NP, ((NP \bs S) \bs (NP \bs S))/NP, NP \to S}
						{\infer[]{NP \to NP}{}
						&
						\infer[(/L)]{NP, (NP \bs S)/NP, NP, (NP \bs S) \bs (NP \bs S) \to S}
							{\infer[]{NP \to NP}{}
							&
							\infer[(\bs L)]{NP, NP \bs S, (NP \bs S) \bs (NP \bs S) \to S}
								{\infer[]{NP \bs S \to NP \bs S}{}
								&
								\infer[(\bs L)]{NP, NP \bs S \to S}
									{\infer[]{NP \to NP}{}
									&
									\infer[]{S \to S}{}}}}}}}}}}}
\]
\[
\mathbf{D_r} =
\infer[(\bs L)]{NP/N, N, N \bs N \to NP}
	{\infer[]{N \to N}{} & \infer[(\bs L)]{NP \bs N, N \to NP}{\infer[]{N \to N}{} & \infer[]{NP \to NP}{}}}
%		{\infer[]{N \to N}{} &
%		\infer[(/L)]{NP/N, N \to NP}
%			{\infer[]{N \to N}{} & \infer[]{NP \to NP}{}}}}
\]
	Next, we provide a step-by-step construction of the diagrammatic interpretation of the preceding derivation. Note that we do not formally distinguish between working with the types/derivable sequents of $\blstar$ and objects/morphisms of $\C(\blstar)$. Formally the diagrams we draw are morphisms in $\C(\blstar)$, but there is no need to consider the object $C_{NP}, C_{(N \Leftarrow N)}$ etc. as it is clear that these correspond exactly to the types $NP, N / N$ etc.
	
	\begin{enumerate}[(1)]
		\item
			
			We start by drawing the domain of the sequent which $\mathbf{D_l}$ is a derivation of, namely 
			\[NP, (NP \bs S)/NP, ((NP \bs S) \bs (NP \bs S))/NP, NP / NP, !NP \to S .\]
			The domain is given as the following diagram, where we suggestively write ``(papers)'' to show where, roughly speaking, a copy of ``papers'' will be sent later in the full morphism:
			\[{%
\beginpgfgraphicnamed{rightDiagram1}
\begin{tikzpicture}
	\begin{pgfonlayer}{nodelayer}
		\node [style=none] (7) at (5, 2) {};
		\node [style=none] (15) at (5, 0) {};
		\node [style=none] (22) at (5.5, 1) {$NP$};
		\node [style=none] (30) at (-5.5, 2) {};
		\node [style=none] (31) at (-3.75, 2) {};
		\node [style=none] (32) at (-2.25, 2) {};
		\node [style=none] (33) at (-0.5, 2) {};
		\node [style=none] (34) at (0.75, 2) {};
		\node [style=none] (35) at (2.5, 2) {};
		\node [style=none] (36) at (3.75, 2) {};
		\node [style=none] (38) at (-5.5, 0) {};
		\node [style=none] (39) at (-3.75, 0) {};
		\node [style=none] (40) at (-2.25, 0) {};
		\node [style=none] (41) at (-0.5, 0) {};
		\node [style=none] (42) at (0.75, 0) {};
		\node [style=none] (43) at (2.5, 0) {};
		\node [style=none] (44) at (3.75, 0) {};
		\node [style=new style 1] (45) at (2.5, 1.5) {};
		\node [style=new style 1] (46) at (-0.5, 1.5) {};
		\node [style=new style 1] (47) at (-3.75, 1.5) {};
		\node [style=none] (48) at (3.75, 1.5) {};
		\node [style=none] (49) at (0.75, 1.5) {};
		\node [style=none] (50) at (-2.25, 1.5) {};
		\node [style=none] (52) at (4.25, 1) {$NP$};
		\node [style=none] (53) at (-6, 1) {$NP$};
		\node [style=none] (54) at (-1.75, 1.25) {$NP$};
		\node [style=none] (55) at (1.75, 0.75) {$NP$};
		\node [style=none] (56) at (1.25, 1.25) {$NP$};
		\node [style=lab] (57) at (-1, 0.75) {\tiny $(NP\Rightarrow S)\Rightarrow (NP\Rightarrow S)$};
		\node [style=none] (58) at (-4.5, 1.25) {$NP\Rightarrow S$};
		\node [style=new style 0] (59) at (-5.5, 2.5) {\strut John};
		\node [style=new style 0] (60) at (-3, 2.5) {\strut signed};
		\node [style=new style 0] (61) at (0.125, 2.5) {\strut without};
		\node [style=new style 0] (62) at (3.125, 2.5) {\strut reading};
		\node [style=new style 0] (63) at (5, 2.5) {\strut (papers)};
	\end{pgfonlayer}
	\begin{pgfonlayer}{edgelayer}
		\draw [style=thickArr] (7.center) to (15.center);
		\draw [style=downArrow] (44.center) to (36.center);
		\draw [style=downArrow] (35.center) to (43.center);
		\draw [style=downArrow] (42.center) to (34.center);
		\draw [style=downArrow] (33.center) to (41.center);
		\draw [style=downArrow] (40.center) to (32.center);
		\draw [style=downArrow] (31.center) to (39.center);
		\draw [style=downArrow] (30.center) to (38.center);
		\draw (48.center) to (45);
		\draw (49.center) to (46);
		\draw (50.center) to (47);
	\end{pgfonlayer}
\end{tikzpicture}}
\endpgfgraphicnamed}\] %jg struts in boxes
			
			Next, we apply the comonoid comultiplication to $\semantics{!NP}$, on the right, giving us:
			\[{%
\beginpgfgraphicnamed{rightDiagram2}
\begin{tikzpicture}
	\begin{pgfonlayer}{nodelayer}
		\node [style=none] (0) at (-5, 0.75) {};
		\node [style=none] (1) at (-3.25, 0.75) {};
		\node [style=none] (2) at (-1.75, 0.75) {};
		\node [style=none] (3) at (0, 0.75) {};
		\node [style=none] (4) at (1.25, 0.75) {};
		\node [style=none] (5) at (3, 0.75) {};
		\node [style=none] (6) at (4.25, 0.75) {};
		\node [style=none] (7) at (5.5, 0.75) {};
		\node [style=none] (8) at (-5, -0.75) {};
		\node [style=none] (9) at (-3.25, -0.75) {};
		\node [style=none] (10) at (-1.75, -0.75) {};
		\node [style=none] (11) at (0, -0.75) {};
		\node [style=none] (12) at (1.25, -0.75) {};
		\node [style=none] (13) at (3, -0.75) {};
		\node [style=none] (14) at (4.25, -0.75) {};
		\node [style=new style 1] (15) at (3, 0.5) {};
		\node [style=new style 1] (16) at (0, 0.5) {};
		\node [style=new style 1] (17) at (-3.25, 0.5) {};
		\node [style=none] (18) at (4.25, 0.5) {};
		\node [style=none] (19) at (1.25, 0.5) {};
		\node [style=none] (20) at (-1.75, 0.5) {};
		\node [style=none] (21) at (6, 0) {$NP$};
		\node [style=none] (22) at (4.75, 0.25) {$NP$};
		\node [style=none] (23) at (-5.5, 0.25) {$NP$};
		\node [style=none] (24) at (-1.25, 0.5) {$NP$};
		\node [style=none] (25) at (2.25, 0) {$NP$};
		\node [style=none] (26) at (1.75, 0.5) {$NP$};
		\node [style=lab] (27) at (-0.625, 0) {$(NP\Rightarrow S)\Rightarrow (NP\Rightarrow S)$};
		\node [style=none] (28) at (-4, 0.5) {$NP\Rightarrow S$};
		\node [style=new style 4] (29) at (5.5, 0) {};
		\node [style=none] (30) at (5, -0.75) {};
		\node [style=none] (31) at (6, -0.75) {};
		\node [style=none] (32) at (6.5, -0.75) {$NP$};
		\node [style=none] (33) at (5.5, -0.75) {$NP$};
		\node [style=new style 0] (34) at (-5, 1.25) {\strut John};
		\node [style=new style 0] (35) at (-2.5, 1.25) {\strut signed};
		\node [style=new style 0] (37) at (0.625, 1.25) {\strut without};
		\node [style=new style 0] (38) at (3.625, 1.25) {\strut reading};
		\node [style=new style 0] (39) at (5.5, 1.25) {\strut (papers)};
	\end{pgfonlayer}
	\begin{pgfonlayer}{edgelayer}
		\draw [style=downArrow] (14.center) to (6.center);
		\draw [style=downArrow] (5.center) to (13.center);
		\draw [style=downArrow] (12.center) to (4.center);
		\draw [style=downArrow] (3.center) to (11.center);
		\draw [style=downArrow] (10.center) to (2.center);
		\draw [style=downArrow] (1.center) to (9.center);
		\draw [style=downArrow] (0.center) to (8.center);
		\draw (18.center) to (15);
		\draw (19.center) to (16);
		\draw (20.center) to (17);
		\draw [style=thickArr] (7.center) to (29);
		\draw [style=thickArr, in=90, out=-120] (29) to (30.center);
		\draw [style=thickArr, in=90, out=-60] (29) to (31.center);
	\end{pgfonlayer}
\end{tikzpicture}}
\endpgfgraphicnamed}\] %jg struts in boxes
			
			We continue by applying the $\mathrm{perm}_1$ rule, twice, to the first/leftmost $\semantics{!NP}$:
			\[{%
\beginpgfgraphicnamed{rightDiagram3}
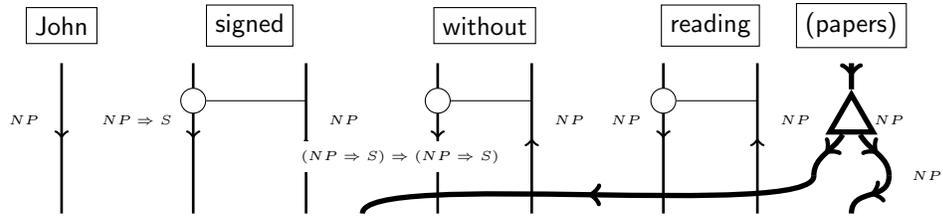
\begin{tikzpicture}
	\begin{pgfonlayer}{nodelayer}
		\node [style=none] (0) at (-5, 0.75) {};
		\node [style=none] (1) at (-3.25, 0.75) {};
		\node [style=none] (2) at (-1.75, 0.75) {};
		\node [style=none] (3) at (0, 0.75) {};
		\node [style=none] (4) at (1.25, 0.75) {};
		\node [style=none] (5) at (3, 0.75) {};
		\node [style=none] (6) at (4.25, 0.75) {};
		\node [style=none] (7) at (5.5, 0.75) {};
		\node [style=none] (8) at (-5, -1.25) {};
		\node [style=none] (9) at (-3.25, -1.25) {};
		\node [style=none] (10) at (-1.75, -1.25) {};
		\node [style=none] (11) at (0, -1.25) {};
		\node [style=none] (12) at (1.25, -1.25) {};
		\node [style=none] (13) at (3, -1.25) {};
		\node [style=none] (14) at (4.25, -1.25) {};
		\node [style=new style 1] (15) at (3, 0.25) {};
		\node [style=new style 1] (16) at (0, 0.25) {};
		\node [style=new style 1] (17) at (-3.25, 0.25) {};
		\node [style=none] (18) at (4.25, 0.25) {};
		\node [style=none] (19) at (1.25, 0.25) {};
		\node [style=none] (20) at (-1.75, 0.25) {};
		\node [style=none] (21) at (6, 0) {$NP$};
		\node [style=none] (22) at (4.75, 0) {$NP$};
		\node [style=none] (23) at (-5.5, 0) {$NP$};
		\node [style=none] (24) at (-1.25, 0) {$NP$};
		\node [style=none] (25) at (2.5, 0) {$NP$};
		\node [style=none] (26) at (1.75, 0) {$NP$};
		\node [style=lab] (27) at (-0.5, -0.5) {$(NP\Rightarrow S)\Rightarrow (NP\Rightarrow S)$};
		\node [style=none] (28) at (-4, 0) {$NP\Rightarrow S$};
		\node [style=new style 4] (29) at (5.5, 0) {};
		\node [style=none] (30) at (5, -0.75) {};
		\node [style=none] (31) at (6, -0.75) {};
		\node [style=none] (32) at (6.5, -0.75) {$NP$};
		\node [style=none] (33) at (-1, -1.25) {};
		\node [style=none] (34) at (5.5, -1.25) {};
		\node [style=new style 0] (35) at (-5, 1.25) {\strut John};
		\node [style=new style 0] (36) at (-2.5, 1.25) {\strut signed};
		\node [style=new style 0] (37) at (0.625, 1.25) {\strut without};
		\node [style=new style 0] (38) at (3.625, 1.25) {\strut reading};
		\node [style=new style 0] (39) at (5.5, 1.25) {\strut (papers)};
	\end{pgfonlayer}
	\begin{pgfonlayer}{edgelayer}
		\draw [style=downArrow] (14.center) to (6.center);
		\draw [style=downArrow] (5.center) to (13.center);
		\draw [style=downArrow] (12.center) to (4.center);
		\draw [style=downArrow] (3.center) to (11.center);
		\draw [style=downArrow] (10.center) to (2.center);
		\draw [style=downArrow] (1.center) to (9.center);
		\draw [style=downArrow] (0.center) to (8.center);
		\draw (18.center) to (15);
		\draw (19.center) to (16);
		\draw (20.center) to (17);
		\draw [style=thickArr] (7.center) to (29);
		\draw [style=thickArr, in=90, out=-120] (29) to (30.center);
		\draw [style=thickArr, in=90, out=-60] (29) to (31.center);
		\draw [style=thickArr, in=90, out=-90, looseness=0.25] (30.center) to (33.center);
		\draw [style=thickArr, in=90, out=-90, looseness=1.25] (31.center) to (34.center);
	\end{pgfonlayer}
\end{tikzpicture}}
\endpgfgraphicnamed}\] %jg struts in boxes
			
			Now we apply the counit to the $\semantics{!NP}$ strings. This corresponds to dropping the bar in the previous calculation.
			\[{%
\beginpgfgraphicnamed{rightDiagram4}
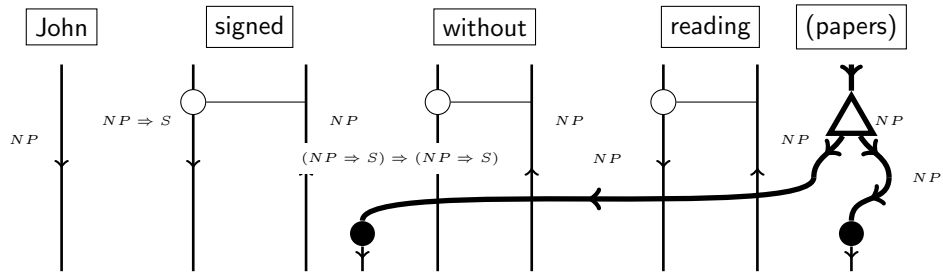
\begin{tikzpicture}
	\begin{pgfonlayer}{nodelayer}
		\node [style=none] (0) at (-5, 0.75) {};
		\node [style=none] (1) at (-3.25, 0.75) {};
		\node [style=none] (2) at (-1.75, 0.75) {};
		\node [style=none] (3) at (0, 0.75) {};
		\node [style=none] (4) at (1.25, 0.75) {};
		\node [style=none] (5) at (3, 0.75) {};
		\node [style=none] (6) at (4.25, 0.75) {};
		\node [style=none] (7) at (5.5, 0.75) {};
		\node [style=none] (8) at (-5, -2) {};
		\node [style=none] (9) at (-3.25, -2) {};
		\node [style=none] (10) at (-1.75, -2) {};
		\node [style=none] (11) at (0, -2) {};
		\node [style=none] (12) at (1.25, -2) {};
		\node [style=none] (13) at (3, -2) {};
		\node [style=none] (14) at (4.25, -2) {};
		\node [style=new style 1] (15) at (3, 0.25) {};
		\node [style=new style 1] (16) at (0, 0.25) {};
		\node [style=new style 1] (17) at (-3.25, 0.25) {};
		\node [style=none] (18) at (4.25, 0.25) {};
		\node [style=none] (19) at (1.25, 0.25) {};
		\node [style=none] (20) at (-1.75, 0.25) {};
		\node [style=none] (21) at (6, 0) {$NP$};
		\node [style=none] (22) at (4.75, -0.25) {$NP$};
		\node [style=none] (23) at (-5.5, -0.25) {$NP$};
		\node [style=none] (24) at (-1.25, 0) {$NP$};
		\node [style=none] (25) at (2.25, -0.5) {$NP$};
		\node [style=none] (26) at (1.75, 0) {$NP$};
		\node [style=lab] (27) at (-0.5, -0.5) {$(NP\Rightarrow S)\Rightarrow (NP\Rightarrow S)$};
		\node [style=none] (28) at (-4, 0) {$NP\Rightarrow S$};
		\node [style=new style 4] (29) at (5.5, 0) {};
		\node [style=none] (30) at (5, -0.75) {};
		\node [style=none] (31) at (6, -0.75) {};
		\node [style=none] (32) at (6.5, -0.75) {$NP$};
		\node [style=counit] (35) at (-1, -1.5) {};
		\node [style=counit] (36) at (5.5, -1.5) {};
		\node [style=none] (37) at (5.5, -2) {};
		\node [style=none] (38) at (-1, -2) {};
		\node [style=new style 0] (39) at (-5, 1.25) {\strut John};
		\node [style=new style 0] (40) at (-2.5, 1.25) {\strut signed};
		\node [style=new style 0] (41) at (0.625, 1.25) {\strut without};
		\node [style=new style 0] (42) at (3.625, 1.25) {\strut reading};
		\node [style=new style 0] (43) at (5.5, 1.25) {\strut (papers)};
	\end{pgfonlayer}
	\begin{pgfonlayer}{edgelayer}
		\draw [style=downArrow] (14.center) to (6.center);
		\draw [style=downArrow] (5.center) to (13.center);
		\draw [style=downArrow] (12.center) to (4.center);
		\draw [style=downArrow] (3.center) to (11.center);
		\draw [style=downArrow] (10.center) to (2.center);
		\draw [style=downArrow] (1.center) to (9.center);
		\draw [style=downArrow] (0.center) to (8.center);
		\draw (18.center) to (15);
		\draw (19.center) to (16);
		\draw (20.center) to (17);
		\draw [style=thickArr] (7.center) to (29);
		\draw [style=thickArr, in=90, out=-120] (29) to (30.center);
		\draw [style=thickArr, in=90, out=-60] (29) to (31.center);
		\draw [style=thickArr, in=90, out=-90, looseness=1.25] (31.center) to (36);
		\draw [style=thickArr, in=90, out=-90, looseness=0.25] (30.center) to (35);
		\draw [style=downArrow] (35) to (38.center);
		\draw [style=downArrow] (36) to (37.center);
	\end{pgfonlayer}
\end{tikzpicture}}
\endpgfgraphicnamed}\] %jg struts in boxes
			
			Next, we evaluate, starting on the right. Evaluations diagrammatically look like cups, and the direction of the arrow on the cup tells us whether it is a left or right evaluation. The first three evaluations are right evaluations. The order of application is recovered by the vertical arrangement.
%			\[\tikzfig{rightDiagram5}\]
%			\[\tikzfig{rightDiagram6}\]
			\[{%
\beginpgfgraphicnamed{rightDiagram7}
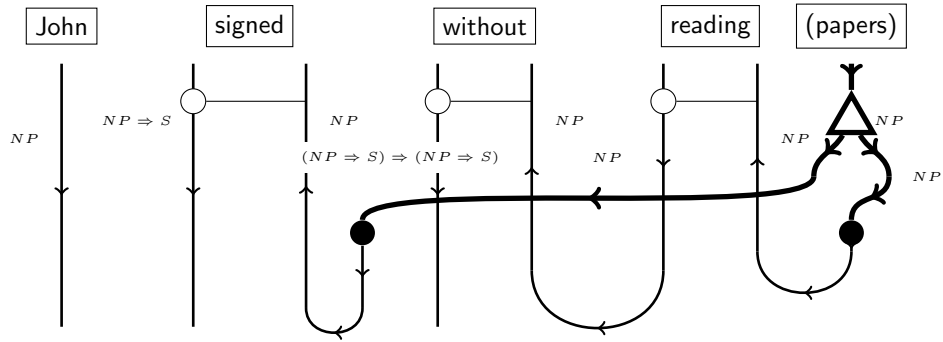
\begin{tikzpicture}
	\begin{pgfonlayer}{nodelayer}
		\node [style=none] (0) at (-5, 1) {};
		\node [style=none] (1) at (-3.25, 1) {};
		\node [style=none] (2) at (-1.75, 1) {};
		\node [style=none] (3) at (0, 1) {};
		\node [style=none] (4) at (1.25, 1) {};
		\node [style=none] (5) at (3, 1) {};
		\node [style=none] (6) at (4.25, 1) {};
		\node [style=none] (7) at (5.5, 1) {};
		\node [style=none] (8) at (-5, -2.5) {};
		\node [style=none] (9) at (-3.25, -2.5) {};
		\node [style=none] (10) at (-1.75, -2.25) {};
		\node [style=none] (11) at (0, -2.5) {};
		\node [style=none] (12) at (1.25, -1.75) {};
		\node [style=none] (13) at (3, -1.75) {};
		\node [style=none] (14) at (4.25, -1.5) {};
		\node [style=new style 1] (15) at (3, 0.5) {};
		\node [style=new style 1] (16) at (0, 0.5) {};
		\node [style=new style 1] (17) at (-3.25, 0.5) {};
		\node [style=none] (18) at (4.25, 0.5) {};
		\node [style=none] (19) at (1.25, 0.5) {};
		\node [style=none] (20) at (-1.75, 0.5) {};
		\node [style=none] (21) at (6, 0.25) {$NP$};
		\node [style=none] (22) at (4.75, 0) {$NP$};
		\node [style=none] (23) at (-5.5, 0) {$NP$};
		\node [style=none] (24) at (-1.25, 0.25) {$NP$};
		\node [style=none] (25) at (2.25, -0.25) {$NP$};
		\node [style=none] (26) at (1.75, 0.25) {$NP$};
		\node [style=lab] (27) at (-0.5, -0.25) {$(NP\Rightarrow S)\Rightarrow (NP\Rightarrow S)$};
		\node [style=none] (28) at (-4, 0.25) {$NP\Rightarrow S$};
		\node [style=new style 4] (29) at (5.5, 0.25) {};
		\node [style=none] (30) at (5, -0.5) {};
		\node [style=none] (31) at (6, -0.5) {};
		\node [style=none] (32) at (6.5, -0.5) {$NP$};
		\node [style=counit] (33) at (-1, -1.25) {};
		\node [style=counit] (34) at (5.5, -1.25) {};
		\node [style=none] (35) at (5.5, -1.5) {};
		\node [style=none] (36) at (-1, -2.25) {};
		\node [style=new style 0] (37) at (-5, 1.5) {\strut John};
		\node [style=new style 0] (38) at (-2.5, 1.5) {\strut signed};
		\node [style=new style 0] (39) at (0.625, 1.5) {\strut without};
		\node [style=new style 0] (40) at (3.625, 1.5) {\strut reading};
		\node [style=new style 0] (41) at (5.5, 1.5) {\strut (papers)};
	\end{pgfonlayer}
	\begin{pgfonlayer}{edgelayer}
		\draw [style=downArrow] (14.center) to (6.center);
		\draw [style=downArrow] (5.center) to (13.center);
		\draw [style=downArrow] (12.center) to (4.center);
		\draw [style=downArrow] (3.center) to (11.center);
		\draw [style=downArrow] (10.center) to (2.center);
		\draw [style=downArrow] (1.center) to (9.center);
		\draw [style=downArrow] (0.center) to (8.center);
		\draw (18.center) to (15);
		\draw (19.center) to (16);
		\draw (20.center) to (17);
		\draw [style=thickArr] (7.center) to (29);
		\draw [style=thickArr, in=90, out=-120] (29) to (30.center);
		\draw [style=thickArr, in=90, out=-60] (29) to (31.center);
		\draw [style=thickArr, in=90, out=-90, looseness=1.25] (31.center) to (34);
		\draw [style=thickArr, in=90, out=-90, looseness=0.25] (30.center) to (33);
		\draw [style=downArrow] (33) to (36.center);
		\draw [style=downArrow] (34) to (35.center);
		\draw [style=downArrow, bend left=90, looseness=1.50] (35.center) to (14.center);
		\draw [style=downArrow, bend left=90, looseness=1.50] (13.center) to (12.center);
		\draw [style=downArrow, bend left=90, looseness=1.50] (36.center) to (10.center);
	\end{pgfonlayer}
\end{tikzpicture}}
\endpgfgraphicnamed}\] %jg struts in boxes
			
			In order to perform the next evaluation, we ``open up'' the $(\semantics{NP}\Rightarrow \semantics{S})\Rightarrow (\semantics{NP} \Rightarrow\semantics{S})$ string. We depict this using ellipses. The result is the following diagram, which is the exact same morphism of $\C(\blstar)$ as the preceding one.
			\[{%
\beginpgfgraphicnamed{rightDiagram8}
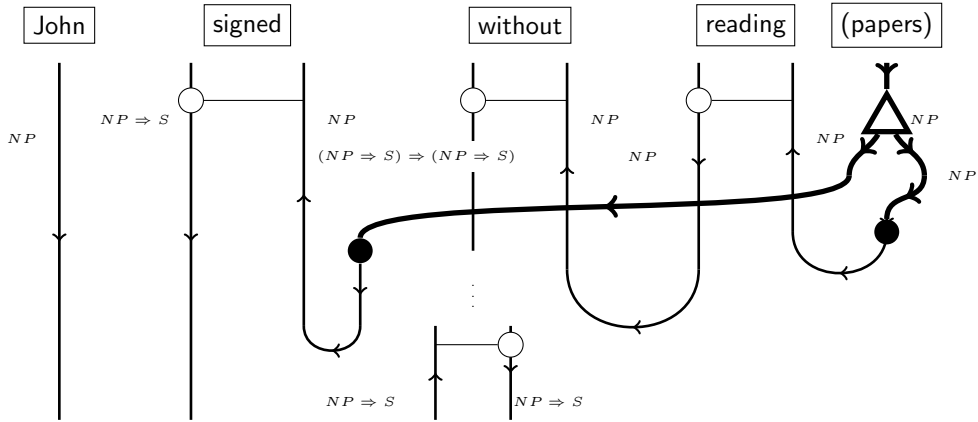
\begin{tikzpicture}
	\begin{pgfonlayer}{nodelayer}
		\node [style=none] (0) at (-5.5, 1) {};
		\node [style=none] (1) at (-3.75, 1) {};
		\node [style=none] (2) at (-2.25, 1) {};
		\node [style=none] (3) at (0, 1) {};
		\node [style=none] (4) at (1.25, 1) {};
		\node [style=none] (5) at (3, 1) {};
		\node [style=none] (6) at (4.25, 1) {};
		\node [style=none] (7) at (5.5, 1) {};
		\node [style=none] (8) at (-5.5, -3.75) {};
		\node [style=none] (9) at (-3.75, -3.75) {};
		\node [style=none] (10) at (-2.25, -2.5) {};
		\node [style=none] (11) at (0, -1.5) {};
		\node [style=none] (12) at (1.25, -1.75) {};
		\node [style=none] (13) at (3, -1.75) {};
		\node [style=none] (14) at (4.25, -1.25) {};
		\node [style=new style 1] (15) at (3, 0.5) {};
		\node [style=new style 1] (16) at (0, 0.5) {};
		\node [style=new style 1] (17) at (-3.75, 0.5) {};
		\node [style=none] (18) at (4.25, 0.5) {};
		\node [style=none] (19) at (1.25, 0.5) {};
		\node [style=none] (20) at (-2.25, 0.5) {};
		\node [style=none] (21) at (6, 0.25) {$NP$};
		\node [style=none] (22) at (4.75, 0) {$NP$};
		\node [style=none] (23) at (-6, 0) {$NP$};
		\node [style=none] (24) at (-1.75, 0.25) {$NP$};
		\node [style=none] (25) at (2.25, -0.25) {$NP$};
		\node [style=none] (26) at (1.75, 0.25) {$NP$};
		\node [style=lab] (27) at (-0.75, -0.25) {$(NP\Rightarrow S)\Rightarrow (NP\Rightarrow S)$};
		\node [style=none] (28) at (-4.5, 0.25) {$NP\Rightarrow S$};
		\node [style=new style 4] (29) at (5.5, 0.25) {};
		\node [style=none] (30) at (5, -0.5) {};
		\node [style=none] (31) at (6, -0.5) {};
		\node [style=none] (32) at (6.5, -0.5) {$NP$};
		\node [style=counit] (33) at (-1.5, -1.5) {};
		\node [style=counit] (34) at (5.5, -1.25) {};
		\node [style=none] (35) at (5.5, -1.25) {};
		\node [style=none] (36) at (-1.5, -2.5) {};
		\node [style=none] (74) at (0, -2) {$\VDOTS$};
		\node [style=none] (75) at (-0.5, -2.5) {};
		\node [style=none] (76) at (0.5, -2.5) {};
		\node [style=new style 1] (77) at (0.5, -2.75) {};
		\node [style=none] (78) at (-0.5, -2.75) {};
		\node [style=none] (79) at (-0.5, -3.75) {};
		\node [style=none] (80) at (0.5, -3.75) {};
		\node [style=none] (81) at (1, -3.5) {\tiny $NP \Rightarrow S$};
		\node [style=none] (82) at (-1.5, -3.5) {\tiny $NP \Rightarrow S$};
		\node [style=new style 0] (83) at (-5.5, 1.5) {\strut John};
		\node [style=new style 0] (84) at (-3, 1.5) {\strut signed};
		\node [style=new style 0] (39) at (0.625, 1.5) {\strut without};
		\node [style=new style 0] (40) at (3.625, 1.5) {\strut reading};
		\node [style=new style 0] (41) at (5.5, 1.5) {\strut (papers)};	\end{pgfonlayer}
	\begin{pgfonlayer}{edgelayer}
		\draw [style=downArrow] (14.center) to (6.center);
		\draw [style=downArrow] (5.center) to (13.center);
		\draw [style=downArrow] (12.center) to (4.center);
		\draw [style=downArrow] (3.center) to (11.center);
		\draw [style=downArrow] (10.center) to (2.center);
		\draw [style=downArrow] (1.center) to (9.center);
		\draw [style=downArrow] (0.center) to (8.center);
		\draw (18.center) to (15);
		\draw (19.center) to (16);
		\draw (20.center) to (17);
		\draw [style=thickArr] (7.center) to (29);
		\draw [style=thickArr, in=90, out=-120] (29) to (30.center);
		\draw [style=thickArr, in=90, out=-60] (29) to (31.center);
		\draw [style=thickArr, in=90, out=-90, looseness=1.25] (31.center) to (34);
		\draw [style=thickArr, in=90, out=-90, looseness=0.25] (30.center) to (33);
		\draw [style=downArrow] (33) to (36.center);
		\draw [style=downArrow] (34) to (35.center);
		\draw [style=downArrow, bend left=90, looseness=1.50] (35.center) to (14.center);
		\draw [style=downArrow, bend left=90, looseness=1.50] (13.center) to (12.center);
		\draw [style=downArrow, bend left=90, looseness=1.50] (36.center) to (10.center);
		\draw (78.center) to (77);
		\draw [style=downArrow] (79.center) to (75.center);
		\draw [style=downArrow] (76.center) to (80.center);
	\end{pgfonlayer}
\end{tikzpicture}}
\endpgfgraphicnamed}\] %jg struts in boxes
			
			We may now evaluate again, giving us:
			\[{%
\beginpgfgraphicnamed{rightDiagram9}
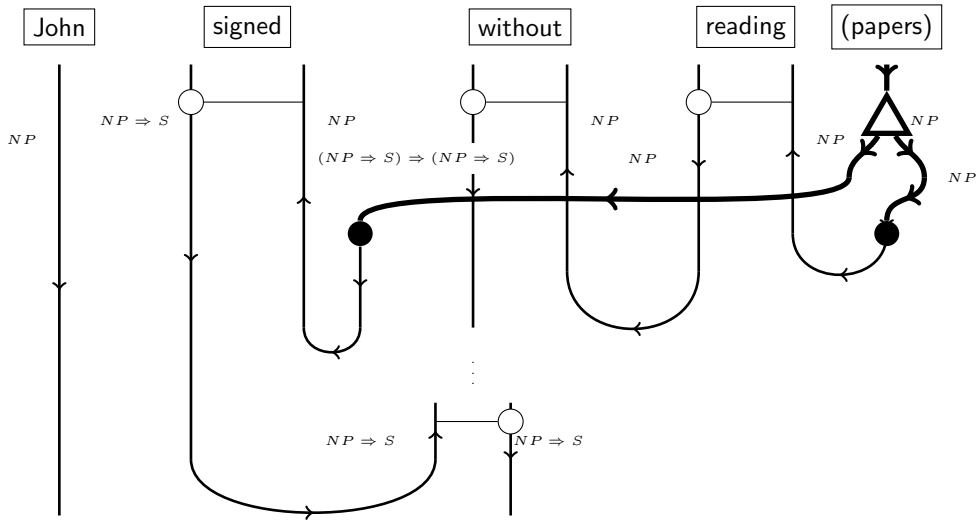
\begin{tikzpicture}
	\begin{pgfonlayer}{nodelayer}
		\node [style=none] (0) at (-5.5, 1) {};
		\node [style=none] (1) at (-3.75, 1) {};
		\node [style=none] (2) at (-2.25, 1) {};
		\node [style=none] (3) at (0, 1) {};
		\node [style=none] (4) at (1.25, 1) {};
		\node [style=none] (5) at (3, 1) {};
		\node [style=none] (6) at (4.25, 1) {};
		\node [style=none] (7) at (5.5, 1) {};
		\node [style=none] (8) at (-5.5, -5) {};
		\node [style=none] (9) at (-3.75, -4.25) {};
		\node [style=none] (10) at (-2.25, -2.5) {};
		\node [style=none] (11) at (0, -2.5) {};
		\node [style=none] (12) at (1.25, -1.75) {};
		\node [style=none] (13) at (3, -1.75) {};
		\node [style=none] (14) at (4.25, -1.25) {};
		\node [style=new style 1] (15) at (3, 0.5) {};
		\node [style=new style 1] (16) at (0, 0.5) {};
		\node [style=new style 1] (17) at (-3.75, 0.5) {};
		\node [style=none] (18) at (4.25, 0.5) {};
		\node [style=none] (19) at (1.25, 0.5) {};
		\node [style=none] (20) at (-2.25, 0.5) {};
		\node [style=none] (21) at (6, 0.25) {$NP$};
		\node [style=none] (22) at (4.75, 0) {$NP$};
		\node [style=none] (23) at (-6, 0) {$NP$};
		\node [style=none] (24) at (-1.75, 0.25) {$NP$};
		\node [style=none] (25) at (2.25, -0.25) {$NP$};
		\node [style=none] (26) at (1.75, 0.25) {$NP$};
		\node [style=lab] (27) at (-0.75, -0.25) {$(NP\Rightarrow S)\Rightarrow (NP\Rightarrow S)$};
		\node [style=none] (28) at (-4.5, 0.25) {$NP\Rightarrow S$};
		\node [style=new style 4] (29) at (5.5, 0.25) {};
		\node [style=none] (30) at (5, -0.5) {};
		\node [style=none] (31) at (6, -0.5) {};
		\node [style=none] (32) at (6.5, -0.5) {$NP$};
		\node [style=counit] (33) at (-1.5, -1.25) {};
		\node [style=counit] (34) at (5.5, -1.25) {};
		\node [style=none] (35) at (5.5, -1.25) {};
		\node [style=none] (36) at (-1.5, -2.5) {};
		\node [style=none] (74) at (0, -3) {$\VDOTS$};
		\node [style=none] (75) at (-0.5, -3.5) {};
		\node [style=none] (76) at (0.5, -3.5) {};
		\node [style=new style 1] (77) at (0.5, -3.75) {};
		\node [style=none] (78) at (-0.5, -3.75) {};
		\node [style=none] (79) at (-0.5, -4.25) {};
		\node [style=none] (80) at (0.5, -5) {};
		\node [style=none] (81) at (1, -4) {\tiny $NP \Rightarrow S$};
		\node [style=none] (82) at (-1.5, -4) {\tiny $NP \Rightarrow S$};
		\node [style=new style 0] (83) at (-5.5, 1.5) {\strut John};
		\node [style=new style 0] (84) at (-3, 1.5) {\strut signed};
		\node [style=new style 0] (94) at (0.625, 1.5) {\strut without};
		\node [style=new style 0] (95) at (3.625, 1.5) {\strut reading};
		\node [style=new style 0] (96) at (5.5, 1.5) {\strut (papers)};
	\end{pgfonlayer}
	\begin{pgfonlayer}{edgelayer}
		\draw [style=downArrow] (14.center) to (6.center);
		\draw [style=downArrow] (5.center) to (13.center);
		\draw [style=downArrow] (12.center) to (4.center);
		\draw [style=downArrow] (3.center) to (11.center);
		\draw [style=downArrow] (10.center) to (2.center);
		\draw [style=downArrow] (1.center) to (9.center);
		\draw [style=downArrow] (0.center) to (8.center);
		\draw (18.center) to (15);
		\draw (19.center) to (16);
		\draw (20.center) to (17);
		\draw [style=thickArr] (7.center) to (29);
		\draw [style=thickArr, in=90, out=-120] (29) to (30.center);
		\draw [style=thickArr, in=90, out=-60] (29) to (31.center);
		\draw [style=thickArr, in=90, out=-90, looseness=1.25] (31.center) to (34);
		\draw [style=thickArr, in=90, out=-90, looseness=0.25] (30.center) to (33);
		\draw [style=downArrow] (33) to (36.center);
		\draw [style=downArrow] (34) to (35.center);
		\draw [style=downArrow, bend left=90, looseness=1.50] (35.center) to (14.center);
		\draw [style=downArrow, bend left=90, looseness=1.50] (13.center) to (12.center);
		\draw [style=downArrow, bend left=90, looseness=1.50] (36.center) to (10.center);
		\draw (78.center) to (77);
		\draw [style=downArrow] (79.center) to (75.center);
		\draw [style=downArrow] (76.center) to (80.center);
		\draw [style=downArrow, bend right=90, looseness=0.75] (9.center) to (79.center);
	\end{pgfonlayer}
\end{tikzpicture}}
\endpgfgraphicnamed}\]  %jg struts in boxes
			
			We repeat the ``opening'' procedure, and evaluate again to get the full diagram of the $\mathbf{D_l}$ sequent:
			\[{%
\beginpgfgraphicnamed{rightDiagram10}
\begin{tikzpicture}
	\begin{pgfonlayer}{nodelayer}
		\node [style=none] (0) at (-5.5, 1) {};
		\node [style=none] (1) at (-3.75, 1) {};
		\node [style=none] (2) at (-2.25, 1) {};
		\node [style=none] (3) at (0, 1) {};
		\node [style=none] (4) at (1.25, 1) {};
		\node [style=none] (5) at (3, 1) {};
		\node [style=none] (6) at (4.25, 1) {};
		\node [style=none] (7) at (5.5, 1) {};
		\node [style=none] (8) at (-5.5, -7.25) {};
		\node [style=none] (9) at (-3.75, -5.25) {};
		\node [style=none] (10) at (-2.25, -3.75) {};
		\node [style=none] (11) at (0, -3) {};
		\node [style=none] (12) at (1.25, -3) {};
		\node [style=none] (13) at (3, -3) {};
		\node [style=none] (14) at (4.25, -2.5) {};
		\node [style=new style 1] (15) at (3, 0.5) {};
		\node [style=new style 1] (16) at (0, 0.5) {};
		\node [style=new style 1] (17) at (-3.75, 0.5) {};
		\node [style=none] (18) at (4.25, 0.5) {};
		\node [style=none] (19) at (1.25, 0.5) {};
		\node [style=none] (20) at (-2.25, 0.5) {};
		\node [style=none] (21) at (6, 0) {$NP$};
		\node [style=none] (22) at (4.75, 0) {$NP$};
		\node [style=none] (23) at (-6, 0) {$NP$};
		\node [style=none] (24) at (-1.75, 0.25) {$NP$};
		\node [style=none] (25) at (2.25, -0.25) {$NP$};
		\node [style=none] (26) at (1.75, 0.25) {$NP$};
		\node [style=lab] (27) at (-0.75, -0.25) {$(NP\Rightarrow S)\Rightarrow (NP\Rightarrow S)$};
		\node [style=none] (28) at (-4.5, 0.25) {$NP\Rightarrow S$};
		\node [style=new style 4] (29) at (5.5, 0) {};
		\node [style=none] (30) at (5, -1) {};
		\node [style=none] (31) at (6, -1) {};
		\node [style=none] (32) at (6.5, -1) {$NP$};
		\node [style=counit] (33) at (-1.5, -1.75) {};
		\node [style=counit] (34) at (5.5, -1.75) {};
		\node [style=none] (35) at (5.5, -2.5) {};
		\node [style=none] (36) at (-1.5, -3.75) {};
		\node [style=none] (74) at (0, -3.5) {$\VDOTS$};
		\node [style=none] (75) at (-0.5, -4) {};
		\node [style=none] (76) at (0.5, -4) {};
		\node [style=new style 1] (77) at (0.5, -4.25) {};
		\node [style=none] (78) at (-0.5, -4.25) {};
		\node [style=none] (79) at (-0.5, -5.25) {};
		\node [style=none] (80) at (0.5, -5.5) {};
		\node [style=none] (81) at (1, -5) {\tiny $NP \Rightarrow S$};
		\node [style=none] (82) at (-1.5, -5) {\tiny $NP \Rightarrow S$};
		\node [style=none] (83) at (0.5, -6) {$\VDOTS$};
		\node [style=none] (84) at (0, -6.5) {};
		\node [style=none] (85) at (1, -6.5) {};
		\node [style=new style 1] (86) at (1, -6.75) {};
		\node [style=none] (87) at (0, -6.75) {};
		\node [style=none] (88) at (0, -7.25) {};
		\node [style=none] (89) at (1, -8.5) {};
		\node [style=none] (90) at (-0.5, -6.75) {\tiny $NP$};
		\node [style=none] (91) at (1.5, -6.75) {\tiny $S$};
		\node [style=new style 0] (92) at (-5.5, 1.5) {\strut John};
		\node [style=new style 0] (93) at (-3, 1.5) {\strut signed};
		\node [style=new style 0] (94) at (0.625, 1.5) {\strut without};
		\node [style=new style 0] (95) at (3.625, 1.5) {\strut reading};
		\node [style=new style 0] (96) at (5.5, 1.5) {\strut (papers)};
	\end{pgfonlayer}
	\begin{pgfonlayer}{edgelayer}
		\draw [style=downArrow] (14.center) to (6.center);
		\draw [style=downArrow] (5.center) to (13.center);
		\draw [style=downArrow] (12.center) to (4.center);
		\draw [style=downArrow] (3.center) to (11.center);
		\draw [style=downArrow] (10.center) to (2.center);
		\draw [style=downArrow] (1.center) to (9.center);
		\draw [style=downArrow] (0.center) to (8.center);
		\draw (18.center) to (15);
		\draw (19.center) to (16);
		\draw (20.center) to (17);
		\draw [style=thickArr] (7.center) to (29);
		\draw [style=thickArr, in=90, out=-120] (29) to (30.center);
		\draw [style=thickArr, in=90, out=-60] (29) to (31.center);
		\draw [style=thickArr, in=90, out=-90, looseness=1.25] (31.center) to (34);
		\draw [style=thickArr, in=90, out=-90, looseness=0.25] (30.center) to (33);
		\draw [style=downArrow] (33) to (36.center);
		\draw [style=downArrow] (34) to (35.center);
		\draw [style=downArrow, bend left=90, looseness=1.50] (35.center) to (14.center);
		\draw [style=downArrow, bend left=90, looseness=1.50] (13.center) to (12.center);
		\draw [style=downArrow, bend left=90, looseness=1.50] (36.center) to (10.center);
		\draw (78.center) to (77);
		\draw [style=downArrow] (79.center) to (75.center);
		\draw [style=downArrow] (76.center) to (80.center);
		\draw [style=downArrow, bend right=90, looseness=0.75] (9.center) to (79.center);
		\draw (87.center) to (86);
		\draw [style=downArrow, bend right=90, looseness=0.75] (8.center) to (88.center);
		\draw [style=downArrow] (88.center) to (84.center);
		\draw [style=downArrow] (85.center) to (89.center);
	\end{pgfonlayer}
\end{tikzpicture}}
\endpgfgraphicnamed}\] %jg struts in boxes
			
			All that remains is currying, which diagrammatically corresponds to bending the rightmost string, and clasping it to the protruding $\semantics{S}$ string. The following is the diagrammatic depiction of the linear map in (1).

			\[{%
\beginpgfgraphicnamed{rightDiagram11}
\begin{tikzpicture}
	\begin{pgfonlayer}{nodelayer}
		\node [style=none] (0) at (-5.5, 1.25) {};
		\node [style=none] (1) at (-3.75, 1.25) {};
		\node [style=none] (2) at (-2.25, 1.25) {};
		\node [style=none] (3) at (0, 1.25) {};
		\node [style=none] (4) at (1.25, 1.25) {};
		\node [style=none] (5) at (3, 1.25) {};
		\node [style=none] (6) at (4.25, 1.25) {};
		\node [style=none] (7) at (6.75, 0.5) {};
		\node [style=none] (8) at (-5.5, -7) {};
		\node [style=none] (9) at (-3.75, -5) {};
		\node [style=none] (10) at (-2.25, -3.5) {};
		\node [style=none] (11) at (0, -2.75) {};
		\node [style=none] (12) at (1.25, -2.75) {};
		\node [style=none] (13) at (3, -2.75) {};
		\node [style=none] (14) at (4.25, -2.25) {};
		\node [style=new style 1] (15) at (3, 0.75) {};
		\node [style=new style 1] (16) at (0, 0.75) {};
		\node [style=new style 1] (17) at (-3.75, 0.75) {};
		\node [style=none] (18) at (4.25, 0.75) {};
		\node [style=none] (19) at (1.25, 0.75) {};
		\node [style=none] (20) at (-2.25, 0.75) {};
		\node [style=none] (21) at (6, 0.25) {$NP$};
		\node [style=none] (22) at (4.75, 0.25) {$NP$};
		\node [style=none] (23) at (-6, 0.25) {$NP$};
		\node [style=none] (24) at (-1.75, 0.5) {$NP$};
		\node [style=none] (25) at (2.25, 0) {$NP$};
		\node [style=none] (26) at (1.75, 0.5) {$NP$};
		\node [style=lab] (27) at (-0.75, 0) {$(NP\Rightarrow S)\Rightarrow (NP\Rightarrow S)$};
		\node [style=none] (28) at (-4.5, 0.5) {$NP\Rightarrow S$};
		\node [style=new style 4] (29) at (5.5, 0.25) {};
		\node [style=none] (30) at (5, -0.75) {};
		\node [style=none] (31) at (6, -0.75) {};
		\node [style=none] (32) at (6.5, -0.75) {$NP$};
		\node [style=counit] (33) at (-1.5, -1.5) {};
		\node [style=counit] (34) at (5.5, -1.5) {};
		\node [style=none] (35) at (5.5, -2.25) {};
		\node [style=none] (36) at (-1.5, -3.5) {};
		\node [style=none] (74) at (0, -3.25) {$\VDOTS$};
		\node [style=none] (75) at (-0.5, -3.75) {};
		\node [style=none] (76) at (0.5, -3.75) {};
		\node [style=new style 1] (77) at (0.5, -4) {};
		\node [style=none] (78) at (-0.5, -4) {};
		\node [style=none] (79) at (-0.5, -5) {};
		\node [style=none] (80) at (0.5, -5.25) {};
		\node [style=none] (81) at (1, -4.75) {\tiny $NP \Rightarrow S$};
		\node [style=none] (82) at (-1.50, -4.75) {\tiny $NP \Rightarrow S$};
		\node [style=none] (83) at (0.5, -5.75) {$\VDOTS$};
		\node [style=none] (84) at (0, -6.25) {};
		\node [style=none] (85) at (1, -6.25) {};
		\node [style=new style 1] (86) at (1, -6.5) {};
		\node [style=none] (87) at (0, -6.5) {};
		\node [style=none] (88) at (0, -7) {};
		\node [style=none] (89) at (1, -8.25) {};
		\node [style=none] (90) at (-0.5, -6.5) {\tiny $NP$};
		\node [style=none] (91) at (1.5, -6.5) {\tiny $S$};
		\node [style=none] (92) at (6.75, -2.75) {};
		\node [style=none] (93) at (2, -7.25) {};
		\node [style=none] (94) at (2, -8.25) {};
		\node [style=none] (95) at (2, -7.5) {};
		\node [style=new style 1] (96) at (1, -7.5) {};
		\node [style=new style 0] (97) at (-5.5, 1.75) {\strut John};
		\node [style=new style 0] (98) at (-3, 1.75) {\strut signed};
		\node [style=new style 0] (99) at (0.625, 1.75) {\strut without};
		\node [style=new style 0] (100) at (3.625, 1.75) {\strut reading};
	\end{pgfonlayer}
	\begin{pgfonlayer}{edgelayer}
		\draw [style=downArrow] (14.center) to (6.center);
		\draw [style=downArrow] (5.center) to (13.center);
		\draw [style=downArrow] (12.center) to (4.center);
		\draw [style=downArrow] (3.center) to (11.center);
		\draw [style=downArrow] (10.center) to (2.center);
		\draw [style=downArrow] (1.center) to (9.center);
		\draw [style=downArrow] (0.center) to (8.center);
		\draw (18.center) to (15);
		\draw (19.center) to (16);
		\draw (20.center) to (17);
		\draw [style=thickArr, bend right=90, looseness=1.75] (7.center) to (29);
		\draw [style=thickArr, in=90, out=-120] (29) to (30.center);
		\draw [style=thickArr, in=90, out=-60] (29) to (31.center);
		\draw [style=thickArr, in=90, out=-90, looseness=1.25] (31.center) to (34);
		\draw [style=thickArr, in=90, out=-90, looseness=0.25] (30.center) to (33);
		\draw [style=downArrow] (33) to (36.center);
		\draw [style=downArrow] (34) to (35.center);
		\draw [style=downArrow, bend left=90, looseness=1.50] (35.center) to (14.center);
		\draw [style=downArrow, bend left=90, looseness=1.50] (13.center) to (12.center);
		\draw [style=downArrow, bend left=90, looseness=1.50] (36.center) to (10.center);
		\draw (78.center) to (77);
		\draw [style=downArrow] (79.center) to (75.center);
		\draw [style=downArrow] (76.center) to (80.center);
		\draw [style=downArrow, bend right=90, looseness=0.75] (9.center) to (79.center);
		\draw (87.center) to (86);
		\draw [style=downArrow, bend right=90, looseness=0.75] (8.center) to (88.center);
		\draw [style=downArrow] (88.center) to (84.center);
		\draw [style=downArrow] (85.center) to (89.center);
		\draw [style=thickArr] (92.center) to (7.center);
		\draw [style=thickArr, in=-90, out=90, looseness=0.75] (93.center) to (92.center);
		\draw [style=thickArr] (94.center) to (93.center);
		\draw (95.center) to (96);
	\end{pgfonlayer}
\end{tikzpicture}}
\endpgfgraphicnamed}\] %jg struts in boxes
		
		\item
			
			Repeating this procedure for $\mathbf{D_r}$ gives the following diagram, where the box labelled ``[that]'' denotes where this type came from, but it is \emph{not} the same as the type of ``that''.
			\[{%
\beginpgfgraphicnamed{leftDiagram}
\begin{tikzpicture}
	\begin{pgfonlayer}{nodelayer}
		\node [style=none] (0) at (-1.25, 1) {};
		\node [style=none] (1) at (-0.25, 1) {};
		\node [style=none] (2) at (-1.25, -2.25) {};
		\node [style=none] (3) at (-0.25, -1) {};
		\node [style=none] (4) at (-0.25, 0.5) {};
		\node [style=new style 1] (5) at (-1.25, 0.5) {};
		\node [style=none] (6) at (3.25, 1) {};
		\node [style=none] (7) at (2.5, 1) {};
		\node [style=none] (8) at (3.25, -1) {};
		\node [style=none] (9) at (2.5, -0.25) {};
		\node [style=none] (10) at (2.5, 0.5) {};
		\node [style=new style 1] (11) at (3.25, 0.5) {};
		\node [style=none] (12) at (-1.75, 0.25) {$NP$};
		\node [style=none] (13) at (0.25, 0.25) {$N$};
		\node [style=none] (14) at (2.25, 0.25) {$N$};
		\node [style=none] (15) at (3.75, 0.25) {$N$};
		\node [style=none] (16) at (1.25, 1) {};
		\node [style=none] (17) at (1.25, -0.25) {};
		\node [style=none] (18) at (1, 0.25) {$N$};
		\node [style=new style 0] (19) at (-0.75, 1.5) {\strut The};
		\node [style=new style 0] (20) at (1.25, 1.5) {\strut papers};
		\node [style=new style 0] (21) at (3, 1.5) {\strut [that]};
	\end{pgfonlayer}
	\begin{pgfonlayer}{edgelayer}
		\draw (4.center) to (5);
		\draw [style=downArrow] (3.center) to (1.center);
		\draw [style=downArrow] (0.center) to (2.center);
		\draw (10.center) to (11);
		\draw [style=downArrow] (9.center) to (7.center);
		\draw [style=downArrow] (6.center) to (8.center);
		\draw [style=downArrow] (16.center) to (17.center);
		\draw [style=downArrow, bend right=90, looseness=1.25] (17.center) to (9.center);
		\draw [style=downArrow, bend left=90] (8.center) to (3.center);
	\end{pgfonlayer}
\end{tikzpicture}}
\endpgfgraphicnamed}\] %jg struts in boxes
			
		\item

			To weave together these two diagrams, we first juxtapose the two graphs, and ``close'' both the protruding $\semantics{S} \Leftarrow \semantics{!NP}$ clasp from (1), and the $\semantics{N} \Rightarrow \semantics{N}$-[that]-clasp from (2).

			\[\scalebox{0.6}{{%
\beginpgfgraphicnamed{pGapEx1.1}
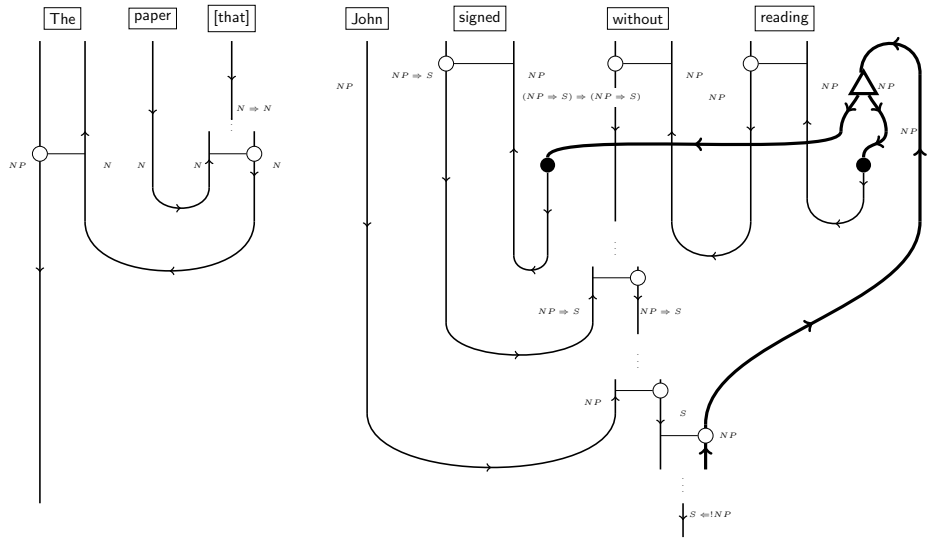
\begin{tikzpicture}
	\begin{pgfonlayer}{nodelayer}
		\node [style=none] (0) at (0.5, 1) {};
		\node [style=none] (1) at (2.25, 1) {};
		\node [style=none] (2) at (3.75, 1) {};
		\node [style=none] (3) at (6, 1) {};
		\node [style=none] (4) at (7.25, 1) {};
		\node [style=none] (5) at (9, 1) {};
		\node [style=none] (6) at (10.25, 1) {};
		\node [style=none] (7) at (12.75, 0.25) {};
		\node [style=none] (8) at (0.5, -7.25) {};
		\node [style=none] (9) at (2.25, -5.25) {};
		\node [style=none] (10) at (3.75, -3.75) {};
		\node [style=none] (11) at (6, -3) {};
		\node [style=none] (12) at (7.25, -3) {};
		\node [style=none] (13) at (9, -3) {};
		\node [style=none] (14) at (10.25, -2.5) {};
		\node [style=new style 1] (15) at (9, 0.5) {};
		\node [style=new style 1] (16) at (6, 0.5) {};
		\node [style=new style 1] (17) at (2.25, 0.5) {};
		\node [style=none] (18) at (10.25, 0.5) {};
		\node [style=none] (19) at (7.25, 0.5) {};
		\node [style=none] (20) at (3.75, 0.5) {};
		\node [style=none] (21) at (12, 0) {$NP$};
		\node [style=none] (22) at (10.75, 0) {$NP$};
		\node [style=none] (23) at (0, 0) {$NP$};
		\node [style=none] (24) at (4.25, 0.25) {$NP$};
		\node [style=none] (25) at (8.25, -0.25) {$NP$};
		\node [style=none] (26) at (7.75, 0.25) {$NP$};
		\node [style=lab] (27) at (5.25, -0.25) {$(NP\Rightarrow S)\Rightarrow (NP\Rightarrow S)$};
		\node [style=none] (28) at (1.5, 0.25) {$NP\Rightarrow S$};
		\node [style=new style 4] (29) at (11.5, 0) {};
		\node [style=none] (30) at (11, -1) {};
		\node [style=none] (31) at (12, -1) {};
		\node [style=none] (32) at (12.5, -1) {$NP$};
		\node [style=counit] (33) at (4.5, -1.75) {};
		\node [style=counit] (34) at (11.5, -1.75) {};
		\node [style=none] (35) at (11.5, -2.5) {};
		\node [style=none] (36) at (4.5, -3.75) {};
		\node [style=none] (74) at (6, -3.5) {$\VDOTS$};
		\node [style=none] (75) at (5.5, -4) {};
		\node [style=none] (76) at (6.5, -4) {};
		\node [style=new style 1] (77) at (6.5, -4.25) {};
		\node [style=none] (78) at (5.5, -4.25) {};
		\node [style=none] (79) at (5.5, -5.25) {};
		\node [style=none] (80) at (6.5, -5.5) {};
		\node [style=none] (81) at (7, -5) {\tiny $NP \Rightarrow S$};
		\node [style=none] (82) at (4.75, -5) {\tiny $NP \Rightarrow S$};
		\node [style=none] (83) at (6.5, -6) {$\VDOTS$};
		\node [style=none] (84) at (6, -6.5) {};
		\node [style=none] (85) at (7, -6.5) {};
		\node [style=new style 1] (86) at (7, -6.75) {};
		\node [style=none] (87) at (6, -6.75) {};
		\node [style=none] (88) at (6, -7.25) {};
		\node [style=none] (89) at (7, -8.5) {};
		\node [style=none] (90) at (5.5, -7) {\tiny $NP$};
		\node [style=none] (91) at (7.5, -7.25) {\tiny $S$};
		\node [style=none] (92) at (12.75, -3) {};
		\node [style=none] (93) at (8, -7.5) {};
		\node [style=none] (94) at (8, -8.5) {};
		\node [style=none] (95) at (7, -7.75) {};
		\node [style=new style 1] (96) at (8, -7.75) {};
		\node [style=none] (97) at (-6.75, 1) {};
		\node [style=none] (98) at (-5.75, 1) {};
		\node [style=none] (99) at (-6.75, -9.25) {};
		\node [style=none] (100) at (-5.75, -3) {};
		\node [style=none] (101) at (-5.75, -1.5) {};
		\node [style=new style 1] (102) at (-6.75, -1.5) {};
		\node [style=none] (103) at (-2, -1) {};
		\node [style=none] (104) at (-3, -1) {};
		\node [style=none] (105) at (-2, -3) {};
		\node [style=none] (106) at (-3, -2.25) {};
		\node [style=none] (107) at (-3, -1.5) {};
		\node [style=new style 1] (108) at (-2, -1.5) {};
		\node [style=none] (109) at (-7.25, -1.75) {$NP$};
		\node [style=none] (110) at (-5.25, -1.75) {$N$};
		\node [style=none] (111) at (-3.25, -1.75) {$N$};
		\node [style=none] (112) at (-1.5, -1.75) {$N$};
		\node [style=none] (113) at (-4.25, 1) {};
		\node [style=none] (114) at (-4.25, -2.25) {};
		\node [style=none] (115) at (-4.5, -1.75) {$N$};
		\node [style=none] (116) at (-2.5, -0.5) {};
		\node [style=none] (117) at (-2.5, -0.75) {$\VDOTS$};
		\node [style=none] (118) at (-2.5, 1) {};
		\node [style=none] (119) at (-2, -0.5) {$N \Rightarrow N$};
		\node [style=none] (120) at (7.5, -9.25) {};
		\node [style=none] (121) at (7.5, -8.75) {$\VDOTS$};
		\node [style=none] (122) at (7.5, -10) {};
		\node [style=none] (123) at (8.10, -9.5) {$S \Leftarrow !NP$};
		\node [style=none] (124) at (8.5, -7.75) {$NP$};
		\node [style=new style 0] (125) at (-6.25, 1.5) {\strut The};
		\node [style=new style 0] (126) at (-4.25, 1.5) {\strut papers};
		\node [style=new style 0] (127) at (-2.5, 1.5) {\strut [that]};
		\node [style=new style 0] (128) at (0.5, 1.5) {\strut John};
		\node [style=new style 0] (129) at (3, 1.5) {\strut signed};
		\node [style=new style 0] (130) at (6.5, 1.5) {\strut without};
		\node [style=new style 0] (131) at (9.75, 1.5) {\strut reading};
	\end{pgfonlayer}
	\begin{pgfonlayer}{edgelayer}
		\draw [style=downArrow] (14.center) to (6.center);
		\draw [style=downArrow] (5.center) to (13.center);
		\draw [style=downArrow] (12.center) to (4.center);
		\draw [style=downArrow] (3.center) to (11.center);
		\draw [style=downArrow] (10.center) to (2.center);
		\draw [style=downArrow] (1.center) to (9.center);
		\draw [style=downArrow] (0.center) to (8.center);
		\draw (18.center) to (15);
		\draw (19.center) to (16);
		\draw (20.center) to (17);
		\draw [style=thickArr, bend right=90, looseness=1.75] (7.center) to (29);
		\draw [style=thickArr, in=90, out=-120] (29) to (30.center);
		\draw [style=thickArr, in=90, out=-60] (29) to (31.center);
		\draw [style=thickArr, in=90, out=-90, looseness=1.25] (31.center) to (34);
		\draw [style=thickArr, in=90, out=-90, looseness=0.25] (30.center) to (33);
		\draw [style=downArrow] (33) to (36.center);
		\draw [style=downArrow] (34) to (35.center);
		\draw [style=downArrow, bend left=90, looseness=1.50] (35.center) to (14.center);
		\draw [style=downArrow, bend left=90, looseness=1.50] (13.center) to (12.center);
		\draw [style=downArrow, bend left=90, looseness=1.50] (36.center) to (10.center);
		\draw (78.center) to (77);
		\draw [style=downArrow] (79.center) to (75.center);
		\draw [style=downArrow] (76.center) to (80.center);
		\draw [style=downArrow, bend right=90, looseness=0.75] (9.center) to (79.center);
		\draw (87.center) to (86);
		\draw [style=downArrow, bend right=90, looseness=0.75] (8.center) to (88.center);
		\draw [style=downArrow] (88.center) to (84.center);
		\draw [style=downArrow] (85.center) to (89.center);
		\draw [style=thickArr] (92.center) to (7.center);
		\draw [style=thickArr, in=-90, out=90, looseness=0.75] (93.center) to (92.center);
		\draw [style=thickArr] (94.center) to (93.center);
		\draw (95.center) to (96);
		\draw (101.center) to (102);
		\draw [style=downArrow] (100.center) to (98.center);
		\draw [style=downArrow] (97.center) to (99.center);
		\draw (107.center) to (108);
		\draw [style=downArrow] (106.center) to (104.center);
		\draw [style=downArrow] (103.center) to (105.center);
		\draw [style=downArrow] (113.center) to (114.center);
		\draw [style=downArrow, bend right=90, looseness=1.25] (114.center) to (106.center);
		\draw [style=downArrow, bend left=90] (105.center) to (100.center);
		\draw [style=downArrow] (118.center) to (117.center);
		\draw [style=downArrow] (120.center) to (122.center);
	\end{pgfonlayer}
\end{tikzpicture}}
\endpgfgraphicnamed}}\] %jg struts in boxes
			
			Finally, we may bend the $\semantics{S} \Leftarrow \semantics{!NP}$ string up and clasp it to the $\semantics{N} \Rightarrow \semantics{N}$ string- This provides the complete diagrammatic interpretation of the $\blstar$-derivation which shows that ``The papers that John signed without reading'' is a noun phrase.
			\[\scalebox{0.6}{{%
\beginpgfgraphicnamed{pGapEx1.2}
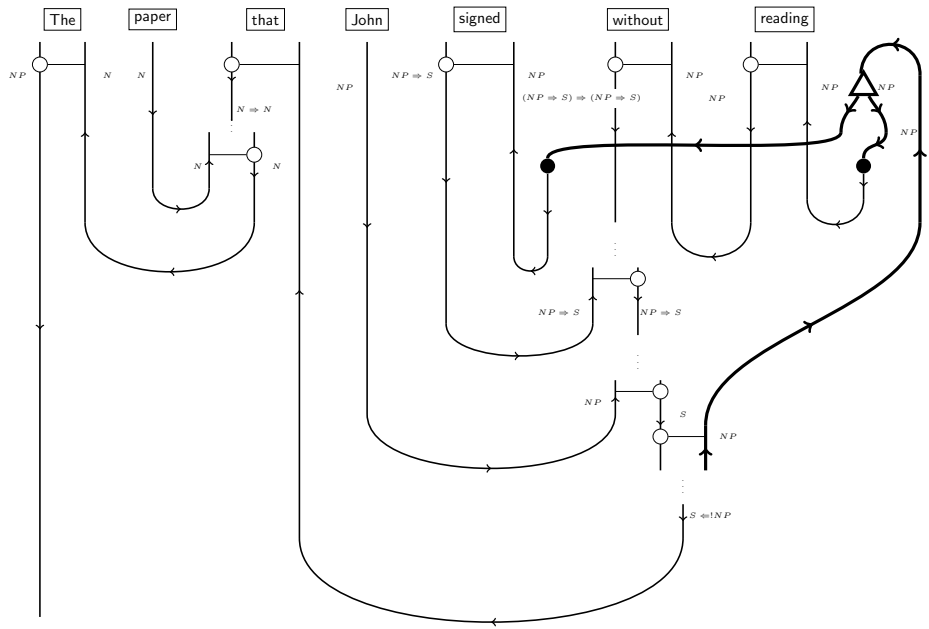
\begin{tikzpicture}
	\begin{pgfonlayer}{nodelayer}
		\node [style=none] (0) at (0.5, 1) {};
		\node [style=none] (1) at (2.25, 1) {};
		\node [style=none] (2) at (3.75, 1) {};
		\node [style=none] (3) at (6, 1) {};
		\node [style=none] (4) at (7.25, 1) {};
		\node [style=none] (5) at (9, 1) {};
		\node [style=none] (6) at (10.25, 1) {};
		\node [style=none] (7) at (12.75, 0.25) {};
		\node [style=none] (8) at (0.5, -7.25) {};
		\node [style=none] (9) at (2.25, -5.25) {};
		\node [style=none] (10) at (3.75, -3.75) {};
		\node [style=none] (11) at (6, -3) {};
		\node [style=none] (12) at (7.25, -3) {};
		\node [style=none] (13) at (9, -3) {};
		\node [style=none] (14) at (10.25, -2.5) {};
		\node [style=new style 1] (15) at (9, 0.5) {};
		\node [style=new style 1] (16) at (6, 0.5) {};
		\node [style=new style 1] (17) at (2.25, 0.5) {};
		\node [style=none] (18) at (10.25, 0.5) {};
		\node [style=none] (19) at (7.25, 0.5) {};
		\node [style=none] (20) at (3.75, 0.5) {};
		\node [style=none] (21) at (12, 0) {$NP$};
		\node [style=none] (22) at (10.75, 0) {$NP$};
		\node [style=none] (23) at (0, 0) {$NP$};
		\node [style=none] (24) at (4.25, 0.25) {$NP$};
		\node [style=none] (25) at (8.25, -0.25) {$NP$};
		\node [style=none] (26) at (7.75, 0.25) {$NP$};
		\node [style=lab] (27) at (5.25, -0.25) {$(NP\Rightarrow S)\Rightarrow (NP\Rightarrow S)$};
		\node [style=none] (28) at (1.5, 0.25) {$NP\Rightarrow S$};
		\node [style=new style 4] (29) at (11.5, 0) {};
		\node [style=none] (30) at (11, -1) {};
		\node [style=none] (31) at (12, -1) {};
		\node [style=none] (32) at (12.5, -1) {$NP$};
		\node [style=counit] (33) at (4.5, -1.75) {};
		\node [style=counit] (34) at (11.5, -1.75) {};
		\node [style=none] (35) at (11.5, -2.5) {};
		\node [style=none] (36) at (4.5, -3.75) {};
		\node [style=none] (74) at (6, -3.5) {$\VDOTS$};
		\node [style=none] (75) at (5.5, -4) {};
		\node [style=none] (76) at (6.5, -4) {};
		\node [style=new style 1] (77) at (6.5, -4.25) {};
		\node [style=none] (78) at (5.5, -4.25) {};
		\node [style=none] (79) at (5.5, -5.25) {};
		\node [style=none] (80) at (6.5, -5.5) {};
		\node [style=none] (81) at (7, -5) {\tiny $NP \Rightarrow S$};
		\node [style=none] (82) at (4.75, -5) {\tiny $NP \Rightarrow S$};
		\node [style=none] (83) at (6.5, -6) {$\VDOTS$};
		\node [style=none] (84) at (6, -6.5) {};
		\node [style=none] (85) at (7, -6.5) {};
		\node [style=new style 1] (86) at (7, -6.75) {};
		\node [style=none] (87) at (6, -6.75) {};
		\node [style=none] (88) at (6, -7.25) {};
		\node [style=none] (89) at (7, -8.5) {};
		\node [style=none] (90) at (5.5, -7) {\tiny $NP$};
		\node [style=none] (91) at (7.5, -7.25) {\tiny $S$};
		\node [style=none] (92) at (12.75, -3) {};
		\node [style=none] (93) at (8, -7.5) {};
		\node [style=none] (94) at (8, -8.5) {};
		\node [style=none] (95) at (8, -7.75) {};
		\node [style=new style 1] (96) at (7, -7.75) {};
		\node [style=none] (97) at (-6.75, 1) {};
		\node [style=none] (98) at (-5.75, 1) {};
		\node [style=none] (99) at (-6.75, -11.75) {};
		\node [style=none] (100) at (-5.75, -3) {};
		\node [style=none] (101) at (-5.75, 0.5) {};
		\node [style=new style 1] (102) at (-6.75, 0.5) {};
		\node [style=none] (103) at (-2, -1) {};
		\node [style=none] (104) at (-3, -1) {};
		\node [style=none] (105) at (-2, -3) {};
		\node [style=none] (106) at (-3, -2.25) {};
		\node [style=none] (107) at (-3, -1.5) {};
		\node [style=new style 1] (108) at (-2, -1.5) {};
		\node [style=none] (109) at (-7.25, 0.25) {$NP$};
		\node [style=none] (110) at (-5.25, 0.25) {$N$};
		\node [style=none] (111) at (-3.25, -1.75) {$N$};
		\node [style=none] (112) at (-1.5, -1.75) {$N$};
		\node [style=none] (113) at (-4.25, 1) {};
		\node [style=none] (114) at (-4.25, -2.25) {};
		\node [style=none] (115) at (-4.5, 0.25) {$N$};
		\node [style=none] (116) at (-2.5, -0.5) {};
		\node [style=none] (117) at (-2.5, -0.75) {$\VDOTS$};
		\node [style=none] (118) at (-2.5, 1) {};
		\node [style=none] (119) at (-2, -0.5) {$N \Rightarrow N$};
		\node [style=none] (120) at (7.5, -9.25) {};
		\node [style=none] (121) at (7.5, -8.75) {$\VDOTS$};
		\node [style=none] (122) at (7.5, -10) {};
		\node [style=none] (123) at (8.1, -9.5) {$S \Leftarrow !NP$};
		\node [style=none] (124) at (8.5, -7.75) {$NP$};
		\node [style=none] (125) at (-1, -10) {};
		\node [style=none] (126) at (-1, 1) {};
		\node [style=new style 1] (127) at (-2.5, 0.5) {};
		\node [style=none] (128) at (-1, 0.5) {};
		\node [style=new style 0] (129) at (-6.25, 1.5) {\strut The};
		\node [style=new style 0] (130) at (-4.25, 1.5) {\strut papers};
		\node [style=new style 0] (131) at (-1.75, 1.5) {\strut that};
		\node [style=new style 0] (132) at (0.5, 1.5) {\strut John};
		\node [style=new style 0] (133) at (3, 1.5) {\strut signed};
		\node [style=new style 0] (134) at (6.5, 1.5) {\strut without};
		\node [style=new style 0] (135) at (9.75, 1.5) {\strut reading};
	\end{pgfonlayer}
	\begin{pgfonlayer}{edgelayer}
		\draw [style=downArrow] (14.center) to (6.center);
		\draw [style=downArrow] (5.center) to (13.center);
		\draw [style=downArrow] (12.center) to (4.center);
		\draw [style=downArrow] (3.center) to (11.center);
		\draw [style=downArrow] (10.center) to (2.center);
		\draw [style=downArrow] (1.center) to (9.center);
		\draw [style=downArrow] (0.center) to (8.center);
		\draw (18.center) to (15);
		\draw (19.center) to (16);
		\draw (20.center) to (17);
		\draw [style=thickArr, bend right=90, looseness=1.75] (7.center) to (29);
		\draw [style=thickArr, in=90, out=-120] (29) to (30.center);
		\draw [style=thickArr, in=90, out=-60] (29) to (31.center);
		\draw [style=thickArr, in=90, out=-90, looseness=1.25] (31.center) to (34);
		\draw [style=thickArr, in=90, out=-90, looseness=0.25] (30.center) to (33);
		\draw [style=downArrow] (33) to (36.center);
		\draw [style=downArrow] (34) to (35.center);
		\draw [style=downArrow, bend left=90, looseness=1.50] (35.center) to (14.center);
		\draw [style=downArrow, bend left=90, looseness=1.50] (13.center) to (12.center);
		\draw [style=downArrow, bend left=90, looseness=1.50] (36.center) to (10.center);
		\draw (78.center) to (77);
		\draw [style=downArrow] (79.center) to (75.center);
		\draw [style=downArrow] (76.center) to (80.center);
		\draw [style=downArrow, bend right=90, looseness=0.75] (9.center) to (79.center);
		\draw (87.center) to (86);
		\draw [style=downArrow, bend right=90, looseness=0.75] (8.center) to (88.center);
		\draw [style=downArrow] (88.center) to (84.center);
		\draw [style=downArrow] (85.center) to (89.center);
		\draw [style=thickArr] (92.center) to (7.center);
		\draw [style=thickArr, in=-90, out=90, looseness=0.75] (93.center) to (92.center);
		\draw [style=thickArr] (94.center) to (93.center);
		\draw (95.center) to (96);
		\draw (101.center) to (102);
		\draw [style=downArrow] (100.center) to (98.center);
		\draw [style=downArrow] (97.center) to (99.center);
		\draw (107.center) to (108);
		\draw [style=downArrow] (106.center) to (104.center);
		\draw [style=downArrow] (103.center) to (105.center);
		\draw [style=downArrow] (113.center) to (114.center);
		\draw [style=downArrow, bend right=90, looseness=1.25] (114.center) to (106.center);
		\draw [style=downArrow, bend left=90] (105.center) to (100.center);
		\draw [style=downArrow] (118.center) to (117.center);
		\draw [style=downArrow] (120.center) to (122.center);
		\draw [style=downArrow, bend left=90, looseness=0.75] (122.center) to (125.center);
		\draw [style=downArrow] (125.center) to (126.center);
		\draw (128.center) to (127);
	\end{pgfonlayer}
\end{tikzpicture}}
\endpgfgraphicnamed}}\] %jg struts in boxes
	\end{enumerate}
\normalsize

The VSS interpretation of this derivation/diagram is a linear map of the form
\begin{eqnarray*}
  &i:& (\semantics{NP}\Leftarrow \semantics{N}) \otimes \semantics{N} \otimes ((\semantics{N} \Rightarrow \semantics{N})\Leftarrow (\semantics{S} \Leftarrow \semantics{!NP})) \otimes \\ %jg4 an extra line break, so as not to overflow
  &&  \semantics{NP} \otimes ((\semantics{NP} \Rightarrow \semantics{S})\Leftarrow \semantics{NP}) \otimes \\
 && 
 (((\semantics{NP} \Rightarrow \semantics{S}) \Rightarrow (\semantics{NP} \Rightarrow \semantics{S}))\Leftarrow \semantics{NP}) \otimes
 (\semantics{NP} \Leftarrow \semantics{NP}) 
 \longrightarrow
 \semantics{NP}
\end{eqnarray*}
defined on elements as follows, denoting the bracketed subscripts using Sweedler notation:
%\scriptsize
%\begin{align*}
%	The( - ) \otimes \ov{paper} \otimes \sum_{that}(C^{that}(m_j^*\otimes m_{j'})\otimes(s_{k_0}\otimes \bar{n}_{i_0}^*)^*) \otimes \ov{John} \otimes signed(-,-) \otimes \sum_{without}(C^{w/o}(n_i^* \otimes s_k)^* \otimes (n_{i'}^*\otimes s_{k'}) \otimes n_{i''}^*) \otimes reading (-)
%	\\ \mapsto
%	\sum_{that ,\, without} the(m_j^*(\ov{paper})m_{j'}(s_{k_0}\otimes \bar{n}_{i_0}^*)^*(n_{i'}^*(\ov{John}s_{k'})(n_i^*\otimes s_k)^*(signed(-,-_{(1)}))n_{i''}^*(reading(-_{(2)}))))
%\end{align*}
%\normalsize
%Or without referring to bases we have the same map with nicer notation, defined as:
\begin{align*}
	i: \textit{The}(-) \otimes \ov{\textit{paper}} \otimes \textit{that}(-,-) \otimes \ov{\textit{John}} \otimes \textit{signed}(-,-) \otimes \textit{without}( -,-,-) \otimes \textit{reading}(-) \\
	\longmapsto \quad \textit{the}(\textit{that}(\ov{\textit{paper}}, \textit{without}(\ov{\textit{John}}, \textit{signed}(-,-_{(1)}), \textit{reading}(-_{(2)})))) . %jg4 full stop
\end{align*}

The above is computed in the following steps: First, we calculate the vector space interpretations of $\mathbf{D_l}$ and then find its transpose to get the interpretation of the line below $\mathbf{D_l}$ in the whole derivation. Then, we calculate the vector space interpretation of $\mathbf{D_r}$. Finally, we combine the interpretations of $\mathbf{D_r}$ and the transpose of $\mathbf{D_l}$ with the $\bs L$ rule to get the full interpretation. The details are as follows:
	
	\begin{enumerate}[(1)]
		\item
		
			We read off the interpretation of $\mathbf{D_l}$ from the derivation rule by rule:
			We start with an element of the domain: $\ov{\textit{John}} \otimes \textit{signed}(-,-) \otimes \textit{without}( -,-,-) \otimes \textit{reading}(-) \otimes \bar{n}$, where $\bar{n}$ is a vector in $\semantics{!NP}$.
			Following the tree, we first copy $\bar{n}$, which in Sweedler notation is given as:
			
			\[\ov{\textit{John}} \otimes \textit{signed}(-,-) \otimes \textit{without}( -,-,-) \otimes \textit{reading}(-) \otimes \bar{n}_{(1)} \otimes \bar{n}_{(2)} . %jg4 full stop
\]
			
			Then we permute $\bar{n}_{(1)}$ in front of $\textit{signed}(-,-)$, again following the tree, giving us
			\[\ov{\textit{John}} \otimes \textit{signed}(-,-) \otimes \bar{n}_{(1)} \otimes \textit{without}( -,-,-) \otimes \textit{reading}(-) \otimes \bar{n}_{(2)} . %jg4 full stop
\]
			Next, the $!L$ rule is applied twice, denoted by dropping the bars:
			\[\ov{\textit{John}} \otimes \textit{signed}(-,-) \otimes n_{(1)} \otimes \textit{without}( -,-,-) \otimes \textit{reading}(-) \otimes n_{(2)} . %jg4 full stop
\]
			
			Now we simply evaluate, again according to the tree, as follows:
			\[\begin{array}{cl}
			& \quad \ov{\textit{John}} \otimes \textit{signed}(-,-) \otimes n_{(1)} \otimes \textit{without}( -,-,-) \otimes \textit{reading}( n_{(2)}) \\
			\stackrel{\id \otimes \ev}{\longmapsto} &\quad \ov{\textit{John}} \otimes \textit{signed}(-,-) \otimes n_{(1)} \otimes \textit{without}( -,-,\textit{reading}( n_{(2)}) \\
			\stackrel{\id \otimes \ev \otimes \id}{\longmapsto} &\quad \ov{\textit{John}} \otimes \textit{signed}(-, n_{(1)}) \otimes \textit{without}( -,-,\textit{reading}( n_{(2)}) \\
			\stackrel{\id \otimes \ev}{\longmapsto} &\quad \ov{\textit{John}} \otimes \textit{without}( -,\textit{signed}(-, n_{(1)}),\textit{reading}( n_{(2)}) \\
			\stackrel{\ev}{\longmapsto} &\quad \textit{without}( \ov{\textit{John}},\textit{signed}(-, n_{(1)}),\textit{reading}( n_{(2)}) ) . %jg4 full stop
			\end{array}\]
			
		 To curry the above map, we essentially `forget' the $\bar{n}$, and the later components $n_{(1)}, n_{(2)}$. Formally, the curried version of the interpretation above is given by the map:
		\begin{align*} %jg linebreak
		\ov{\textit{John}} \otimes \textit{signed}(-,-) \otimes \textit{without}( -,-,-) \otimes \textit{reading}(-) 
                \\ \mapsto 
                \textit{without}( \ov{\textit{John}},\textit{signed}(-, -_{(1)}),\textit{reading}( -_{(2)}) )
		\end{align*}
		
		\item 
		
			We apply the same method to calculate the interpretation of $\mathbf{D_r}$, taking $(m_j)_j$ as a basis of $\semantics{N}$, giving us the arbitrary element $\sum_j C_j m_j^\bot \otimes m_j \in \semantics{N}\Rightarrow\semantics{N}$, yielding the following linear map:
			\begin{align*}
				\textit{the}(-) \otimes \ov{\textit{paper}} \otimes \sum_j C_j m_j^\bot \otimes m_j 
				\stackrel{\id \otimes \ev}{\longmapsto} \textit{the}(-) \otimes \sum_j C_j m_j^\bot (\ov{\textit{paper}}) m_j 
				\\ \stackrel{\ev}{\longmapsto} \textit{the} \left( \sum_j C_j m_j^\bot (\ov{\textit{paper}}) m_j \right) . %jg4 full stop
			\end{align*}
		
		\item
			
			We combine (2) and (3) using the $(/ L)$ rule. To make the application of the rule clear, we mark where the rule is applied using the same symbols as the rule presentation in Table \ref{Lrules}:
			\[\overbrace{\textit{the}(-) \otimes \ov{\textit{paper}}}^{\Delta_1} \otimes \overbrace{\textit{that}(-,-)}^{B/ A} \otimes \overbrace{\ov{\textit{John}} \otimes \textit{signed}(-,-) \otimes \textit{without}( -,-,-) \otimes \textit{reading}(-)}^{\Gamma} .\]
			
			The interpretation of the full derivation is then the following map:
			\begin{align*}
			i: \textit{the}(-) \otimes \ov{\textit{paper}} \otimes \textit{that}(-,-) \otimes \ov{\textit{John}} \otimes \textit{signed}(-,-) \otimes \textit{without}( -,-,-) \otimes \textit{reading}(-) \\
			 \longmapsto \quad \textit{the}(\textit{that}(\ov{\textit{paper}}, \textit{without}(\ov{\textit{John}}, \textit{signed}(-,-_{(1)}), \textit{reading}(-_{(2)})))) . %jg4 full stop
			\end{align*}
	\end{enumerate}

%\begin{figure}[!h]
%\[\scalebox{0.6}{\tikzfig{pGapEx}}\]
%\caption{Diagrammatic Semantics of ``the paper John signed without reading''}
%\label{pGapDiagram}
%\end{figure}
%

\section{Experimental Validation}
The reader might have rightly been wondering which one of these interpretations, the Cogebra or the Cofree-inspired coalgebra model, produces the correct semantic representation. We provide an answer by performing an experimental evaluation of these copying maps. In a nutshell, this is done by implementing the resulting vector representations on large corpora of data and experiment with a task to provide insights. We chose an extension of a disambiguation task often used in the DisCoCat. For implementation, we use different state of the art neural word vector embeddings for nouns and follow a well known procedure to turn them into verb matrices. We instantiate our copying maps on these vectors and matrices and create phrase vectors and use these in the disambiguation task. 

\subsection{The Parasitic Gap Phrase Disambiguation Task}
 The disambiguation task was that originally proposed in \cite{GrefenSadrEMNLP}, but we work with the advanced data set of \cite{KartSadrCoNLL}, which contains verbs deemed as \emph{genuinely ambiguous} by \cite{PickeringFrisson}, as those verbs whose meanings are not related to each other. We extended this latter with a second verb and a preposition that provided enough data to turn the dataset from a set of pairs of transitive sentences to a set of pairs of parsitic gap phrases. As an example, consider the verb {\bf file}, with meanings {\bf register} and {\bf smooth}. The original dataset has sentences: %jg4 colon
\def\PRIME{$^\prime$}
\begin{quote}
S: accounts that the local government filed \\
 S1: accounts that the local government registered\\
 S2: accounts that the local government smoothed\\
 S\PRIME: nails that the young woman filed\\
 S\PRIME1: nails that the young woman registered\\
 S\PRIME2: nails that the young woman smoothed
 \end{quote}

We extend these to parasitic gap phrases: %jg4 colon
\begin{quote}
P: accounts that the local government filed after inspecting \\ %jg deleted stray single quote
 P1: accounts that the local government registered after inspecting \\
 P2: accounts that the local government smoothed after inspecting\\
 P\PRIME: nails that the young woman filed after cutting\\
 P\PRIME1: nails that the young woman registered after cutting\\
 P\PRIME2: nails that the young woman smoothed after cutting
 \end{quote}
 
The extension process was as follows. For each entry of \cite{KartSadrCoNLL}, we needed an extra verb and a preposition to turn it into a parasitic gap phrase. The verb was chosen from a list of verbs that had most frequently colocated with the verb of the original sentence. The preposition was chosen either from examples of phrases with these verbs in the context of the first verb, or decided based on the meaning of the second verb. We annotated each entry with a binary label, where 0 indicated a bad disambiguation pair and 1 a good one. For instance, in the above example, pairs that got a 1 label were (P, P1) and (P\PRIME, P\PRIME2), %jg P\PRIME2 was P"2 - a typo?
whereas pairs that got a 0 label where (P, P2) and (P\PRIME, P\PRIME1). 
The full dataset is online \cite{datasets}. %jg don't use footnote for reference

 We then followed the same procedure as in \cite{KartSadrCoNLL} to disambiguate the phrases with the ambiguous verb: (1) build vectors for phrases P, P1, and P2, and also P\PRIME, P\PRIME1, and P\PRIME2, (2) check whether vector of P is closer to vector of P1 or vector of P2 and whether P\PRIME{} is close to P\PRIME2 or P\PRIME1. If yes, then we have two correct outputs, (3) compute a mean average precision (MAP), by counting in how many of the pairs vector of the phrase with the ambiguous verb in it is closer to the vector of the phrase with its appropriate meaning.

 \subsection{Word Vectors and Verb Matrices}

We work with the parasitic gap phrases that have the general form:
\begin{quote}
 ``A’s the B C’ed Prep D’ing''
 \end{quote}
 where C and D are verbs and their vector representations are multilinear maps. C is a bilinear map that takes A and B as input and D is a linear map that takes A as input. We represent the preposition Prep by the trilinear map $\textit{Prep}$. %jg \textit not \mathit
The vector representation of the parasitic gap phrase with a proper copying operator is as follows:
 \[
 \textit{Prep}({C}(\ov{B},\ov{A}), D(\ov{A}))
 \]
 for $C$ and $D$ multilinear maps and $\ov{A}$ and $\ov{B}$, vectors, denoted as follows:
\[
\ov{A} = \sum_i C^A_i n_i \qquad \ov{B} = \sum_i C^B_i n_i . %jg4 full stop
\]
The multilinear maps that implement the verbs are built using a method referred to as \emph{Relational} in DisCoCat \cite{GrefenSadrEMNLP}. Given a verb $V$, this map is built following the formula given below:
\[
V = \sum_i \ov{s}_i \otimes \ov{o}_i
\]
where $\ov{s}_i$'s are the vector representations of the subjects of the verb $V$ across the corpus and $\ov{o}_i$'s are the vector representations of its objects. The algorithm that implements this formula works as follows:
\begin{tabbing}
1: go over a part of speech tagged corpus sentence by sentence\\
2: for each sentence whose main verb is $V$, extracts its nominal subject and object\\
3: loop\\
\qquad -- set $sum$ to be 0\\
\qquad -- retrieve vector representations of the subject and object\\
\qquad -- take their Kronecker tensor \\
\qquad -- add the Kronecker tensors to $sum$
\end{tabbing}
The implementation details were as follows:
\begin{enumerate}
\item Each Kronecker tensor is recursively added to the Kronecker tensors obtained from other sentences of the corpus that have $V$ as their main verb. 
\item In cases where subjects/objects are noun phrases, the main noun of the phrases is extracted and used as the subject/object. This is what is referred to as \emph{nominal} in the pseudo code. 
\item For the vector representations of the nominal subjects and objects of sentences, we experimented with three neural embedding architectures: BERT\footnote{Note that in general BERT gives %jg3 suggest deleting "you" [agreed]
WordPiece embeddings should a word be able to be split into tokens, however BERT composes such WordPiece embeddings internally via the WordPiece Tokenisation Algorithm \cite{tokenization} to provide our desired word-embedding.} \cite{BERT}, FastText (FT) \cite{FastText}, and Word2Vec CBOW (W2V) \cite{Word2Vec}. For BERT, the architecture itself provides phrase embeddings, read from an internal hidden layer of the network. For FT, W2V and GloVe, as customary, we added the word embeddings to obtain phrase embeddings. The $\textit{Prep}$ was taken to be addition in all cases.
\item For BERT, the experimental results are obtained by using ``BERT-Base, Uncased: 12-layer, 768-hidden, 12-heads, 110M parameters'' pre-trained model\cite{base}. %jg don't use footnote for reference
We used bert-as-service \cite{bert-as-service}\footnote{Should the reader wish to reproduce this experiment, we note that it is easier to use the following transformers library \cite{transformers} with the following two lines of code to load the pretrained BERT model:\\
\texttt{from transformers import TFBertTokenizer} \\ \texttt{tf\_tokenizer = TFBertTokenizer.from\_pretrained("bert-base-uncased")}}.
 for encoding. For W2V and FT, we experimented with 300 dimensional vectors, trained with a minimum word frequency of 50 and a window of 5 over 15 iterations, on the ukWackypedia corpus.

\item For each verb, we collected sentences from ukWaC corpus \cite{ukwac} using the NoSketch web interface \cite{nosketch}. %jg don't use footnote for reference
We filtered the sentences having the input verb as the main verb. For each phrases of the dataset, we also built phrase vectors using the neural architectures. 

\item For the actual implementations, we used the \texttt{NumPy} library \cite{numpy} of programming language \emph{Python~3}. %jg don't use footnote for reference
NumPy is the fundamental package for scientific computing in Python. It is a Python library that provides a multidimensional array object, various derived objects (such as masked arrays and matrices), and an assortment of routines for fast operations on arrays, including mathematical, logical, shape manipulation, sorting, selecting, I/O, discrete Fourier transforms, basic linear algebra, basic statistical operations, random simulation and much more. The Python %jg capitalize Python
scripts were run using Google's \emph{Colab} tool, which enables access to more GPUs and RAM. 
The notebooks are available on a Google Drive \cite{notebooks}, along with the code for BERT embeddings and the tensors we built from them. The W2V and FT embeddings and their associated tensors \cite{compdisteval-ellipsis} were developed for the experiments of previous work \cite{wijnholds-sadrzadeh-2019-evaluating}. %jg don't use footnote for reference

\end{enumerate}

On a historical note, the Relational method was originally introduced in \cite{GrefenSadrEMNLP} to provide an experimental validation for the DisCoCat model, and later studied further in \cite{Grefenstette2015}. This method was used in experimental validations of applications of DisCoCat, e.g. in \cite{KartSadrCoNLL,kartsaklis-sadrzadeh-2013-prior}, its comparison with neural word embeddings \cite{milajevs-etal-2014-evaluating} and extensions of DisCoCat, e.g. to entailment \cite{Sadrzadeh2018} and to larger fragments of language containing relative clauses \cite{Sadretal2013Frob}.

\subsection{Copying Maps and Phrase Vectors}

The different types of copying maps developed in this paper provide us with the following options for the vector representation of each parasitic gap phrase.
\begin{eqnarray*}
\mbox{\bf Cogebra copying} &&
\begin{array}[t]{@{}l@{}} %jg linebreak
(a)\, \textit{Prep} \left( C(\ov{B}, \ov{A}), D(\sum_i n_i) \right), \\
(b)\, \textit{Prep} \left( C(\ov{B}, \sum_i n_i), D(\ov{A})\right)
\end{array}\\
\mbox{\bf Cofree-inspired copying} && \textit{Prep} \left(C(\ov{B}, \ov{A}) + D(\vec{1}), C(\ov{B},\vec{1}) + D(\ov{A})\right)
%\mbox{\bf full copying} && \textit{Prep} \left( C(\ov{B}, \ov{A}), D(A) \right)
\end{eqnarray*}
When the Relational multilinear map of each verb is applied to the vectors of the subject $A$ and object $B$ of each parasitic gap phrase, the above expressions get simplified. 
%We follow the \emph{copy object} simplification method of \cite{KartSadrCoNLL}. 
In this case, the copying maps reduced to the following vector representations:
\begin{eqnarray*}
\mbox{\bf Cogebra copying} && 
\begin{array}[t]{@{}l@{}} %jg linebreak
(a)\, \textit{Prep} \left(\ov{A} \odot ({C} \times \ov{B}), {D} \times \sum_i n_i \right), \\
(b)\, \textit{Prep} \left((\sum_i n_i) \odot ({C} \times \ov{B}), {D} \times \ov{A})\right)
\end{array}\\
\mbox{\bf Cofree-inspired copying} &&\textit{Prep} \left((\ov{A} \odot ({C} \times \ov{B})) + ({D} \times \vec{1}), (\vec{1} \odot ({C} \times \vec{B})) + ({D} \times \vec{A})\right) . %jg4 full stop
%\mbox{\bf full copying} && \textit{Prep} \left(\ov{A} \odot ({C} \times \ov{B}), {D} \times \ov{A} \right)
\end{eqnarray*}

\noindent
For comparison, we also implemented a model where a \emph{Full} copying operation $\Delta(\ov{v}) = \ov{v} \otimes \ov{v}$ was used, resulting in a third option $\textit{Prep} \left( C(\ov{B}, \ov{A}), D(A) \right)$, with the \emph{copy-object} model 
\[
\textit{Prep} \left(\ov{A} \odot ({C} \times \ov{B}), {D} \times \ov{A} \right) . %jg4 full stop
\]
Note that this copying is non-linear and thus cannot be an instance of our $\fdVect$ categorical semantics; we are only including it to study how the other copying models will do in relation to it. 

\subsection{Results}

\begin{table}[!t]
\centering
\begin{tabular}{|c||c||c|c||c|c|}
\hline
 Model & MAP & Model & MAP & Model & MAP \\
\hline \hline
 BERT & 0. 65 & FT(+) & 0.55 & W2V (+) & 0.46 \\
 \hline
Full & 0.48 & Full & 0.57 & Full & 0.54 \\
 Cofree-inspired & 0.47 & Cofree-inspired & 0.56 & Cofree-inspired & 0.54 \\
 Cogebra (a) & 0.46 & Cogebra (a) & 0.56 & Cogebra (a) & 0.46 \\
 Cogebra (b) & 0.42 & Cogebra (b) & 0.37 & Cogebra (b) & 0.39\\
\hline
\end{tabular}
\centering
\caption{Parasitic Gap Phrase Disambiguation Results} %jg caption after table, like Tables 1,2
\label{tab:exampleres}
\end{table}

The results of experimenting with these models are presented in Table~\ref{tab:exampleres}. Uniformly, in all the neural architectures, the Full model provided a better disambiguation than other linear copying models. This better performance was closely followed by the Cofree-inspired model: in BERT, the Full model obtained an MAP of 0.48, and the Cofree-inspired model an MAP of 0.47; in FT, we have 0.57 for Full and 0.56 for Cofree-inspired; and in W2V we have 0.54 for both models. Also uniformly, in all of the neural architectures, the Cogebra (a) did better than the Cogebra (b). It is not surprising that the Full copying did better than other two copyings, since this is the model that provides two identical copies of the head noun $A$. This kind of copying can only be obtained via the application of a non-linear $\Delta$. The fact that our linear Cofree-inspired copying closely followed the Full model, shows that in the absence of Full copying, we can always use the Cofree-inspired as a reliable approximation. It was also not surprising that the Cofree-inspired model did better than either of the Cogebra models, as this model uses the sum of the two possibilities, each encoded in one of the Cogebra (a) or (b). That Cogebra (a) performed better than Cogebra (b), shows that it is more important to have a full copy of the object for the main verb rather than the secondary verb of a parasitic gap phrase. Using this, we can say that verb $C$ that got a full copy of its object $A$, played a more important role in disambiguation, than verb $D$, which only got a vector of 1's as a copy of $A$. Again, this is natural, as the secondary verb only provides subsidiary information. 

The most effective disambiguation of the new dataset was obtained via the BERT phrase vectors, followed by the {Full} model. BERT is a contextual neural network architecture that provides different meanings for words in different contexts, using a large set of tuned parameters on large corpora of data. There is evidence that BERT's phrase vectors do encode some grammatical information in them. So it is not surprising that these embeddings provided the best disambiguation result. In the other neural embeddings: W2V and FT, however, the Full and its Cofree-inspired approximation provided better results. Recall that in these models, phrase embeddings are obtained by just adding the word embeddings, and addition forgets the grammatical structure. That the categorical models, which are type-driven and work along the grammatical structure, outperformed these models is a very promising result. 

We note that MAP's of all models were quite high in comparison with the results of the original dataset, i.e. \cite{GrefenSadrEMNLP}. The original dataset consists of Subject-Verb-Object (SVO) sentences. The fact that our results are better means that turning the SVO sentences into parasitic gap phrases, which are longer and provide more context for disambiguation, has helped disambiguate the verbs better. Finally, although our goal has not been to outperform the performance of holistic neural phrase embeddings that do not perform copying or any other explicit grammatical operations, in two out of three of these (W2V and FT), %jg2 missing close paren
our best model had a better accuracy. 

\section{Conclusions and Further Work}\label{sec:conclusion}

We have provided sound categorical and vector space semantics for the Lambek calculus with relevant modality, and have introduced candidate diagrammatic semantics. 
We provided three different vector space interpretations for the relevant modality and experimented with them in a disambiguation task. In order to do so, we extended the dataset of \cite{KartSadrCoNLL} to parasitic gap phrases with a main ambiguous verb. We implemented the models using three different neural network architectures. One of our interpretations performed very similar to, and in one case the same as, a full but non-linear copying model. The best categorical phrase models performed better than the additive neural phrase embeddings. The state of the art neural phrase embedding (BERT), however, provided the overall best disambiguation result. This is not surprising since these models encode both contextual and structural phrase information. The results of the categorical models can be improved by building better quality bilinear maps: a direction we are pursuing at the moment. 

Proving coherence of the diagrammatic semantics using proof nets of Modal Lambek Calculus \cite{moortgat1996multimodal}, developed for clasp-string diagrams in \cite{wijnholds2017coherent} constitutes work in progress. Proving coherence would allow us to do all our derivations diagrammatically, making the sequent calculus labour superfluous. However, we suspect there are better notations for the diagrammatic semantics perhaps more closely related to the proof nets of linear logic. Another path to explore is that of differential categories. The structure of a differential category as laid out in \cite{St2006} seems an appropriate setting for our work, yet we do not make full use of the actual differential structure in this paper, just the coalgebra modality. Perhaps there is more structure available with a useful linguistic use.

Another avenue to explore is to alter the underlying syntax i.e. $\blstar$. There appears to be a way to achieve a model of contraction which in practice is exactly the full copying but whose underlying theory yields it  as a linear map, %jg3 I couldn't parse that [Have amended]
namely a projection from a bounded tensor algebra. This would be done by bounding the $!$-functor in a style similar to that of Bounded Linear Logic \cite{Girard1992} or Soft Linear Logic \cite{Lafont2004}.

%jg4 remove ISSN, ISBN
%jg4 typeset URLs in "note" field
%jg4 DOIs should display complete URL, and omit "DOI"
\bibliographystyle{plainnat} %jg use natbib style
\bibliography{references} %jg fixed MANY references

\end{document}